%% file: main_arxiv.tex
\documentclass[11pt,letterpaper]{article}

\newif\ifsdm \sdmfalse

\usepackage[margin=1in]{geometry}
\usepackage[utf8]{inputenc}
\usepackage[T1]{fontenc}

\usepackage{amsmath, amssymb, amsfonts, amsthm}
\usepackage{mathtools}
\usepackage{amsopn}

\usepackage{graphicx}
\usepackage{subcaption}
\usepackage{epstopdf}
\usepackage{booktabs}
\usepackage{multirow}
\usepackage{float}
\usepackage{xcolor}

\usepackage[shortlabels]{enumitem}
\usepackage{algorithm}
\usepackage{algorithmic}

\usepackage[numbers,round]{natbib}

\usepackage{authblk}

\usepackage{url}
\usepackage{hyperref}
\hypersetup{colorlinks,citecolor=blue,linkcolor=blue,urlcolor=blue,breaklinks=true}

\theoremstyle{plain}
\newtheorem{theorem}{Theorem}[section]
\newtheorem{lemma}[theorem]{Lemma}

\theoremstyle{definition}

\theoremstyle{remark}

\begin{document}

\title{\textbf{Sculpting the Latent Space: \\ Disentanglement with WAE-MMD \\ and Programmable Priors}}

\author[1]{Quentin Fruytier\thanks{Work performed during an internship at InterDigital Communications, AI Lab.}}
\author[2]{Akshay Malhotra}
\author[2]{Shahab Hamidi-Rad}
\author[2]{Aditya Sant}
\author[1]{Aryan Mokhtari}
\author[1]{Sujay Sanghavi}
\affil[1]{The University of Texas at Austin}
\affil[2]{InterDigital Communications, AI Lab}

\date{}
\maketitle

\begin{abstract}
Learning disentangled representations, where semantic features are independently captured in a latent space, is dominated by Variational Auto Encoders (VAE) and Adversarial Training. In this paper, we provide direct quantitative and visual evidence that these methods consistently fail to enforce the target factorized representation on the aggregate posterior, leading to entangled representations. We quantify this failure using the Latent Predictability Score (LPS), an unsupervised metric of mutual predictability that empirically lower bounds the level of entanglement in a latent space. To address these limitations, we leverage the non-parametric Maximum Mean Discrepancy (MMD) to directly sculpt the aggregate posterior enabling programmable priors. We demonstrate (1) state-of-the-art unsupervised statistical independence, with a 70\% reduction in detected entanglement compared to leading baselines; (2)~the flexibility to optionally inject semi-supervised information for explicit semantic alignment; and (3) precise shaping of the aggregate posterior distribution (validated through quantization-aware training) without sacrificing reconstruction quality.
\end{abstract}

\input{Sections/Introduction.tex}
\input{Sections/Method.tex}
\input{Sections/ExpnResults.tex}

\input{Sections/Conclusion.tex}

\bibliographystyle{plainnat}
\bibliography{references}

\clearpage
\thispagestyle{empty}
\null
\clearpage

\appendix
\input{Sections/Appendix}

\end{document}

%% file: Sections/Introduction.tex
\section{Introduction}
\label{sec:Introduction}

A particularly promising goal of modern machine learning is to learn \textit{disentangled} representations, where individual latent units are mutually independent and semantically interpretable \citep{higgins2017beta}. Achieving this is considered a critical step towards more robust and generalizable models, ultimately enabling capabilities like compositional generalization, fairness, and causal reasoning \citep{scholkopf2021toward}. However, disentangled representations are not unique, and guaranteed alignment to specific semantic factors is theoretically impossible without some level of supervision \cite{HYVARINEN1999429, khemakhem2020variational, locatello2019challenging, locatello2020weakly}.

As such, the dominant paradigm for unsupervised disentanglement has been limited to explicitly or implicitly encouraging the latent space to adopt a representation where each latent unit is statistically independent. Seminal works utilize the Variational Autoencoder (VAE) \citep{higgins2017beta, kim2018disentangling, chen2018isolating, burgess2018understanding, mathieu2019disentangling, locatello2020weakly, khemakhem2020variational} or Adversarial Training \citep{chen2016infogan, yeats2022nashae} to constrain the posterior distribution toward this goal. However, this paradigm has been scrutinized for failing to achieve true disentanglement \citep{locatello2019challenging}, as the recovery of interpretable factors is frequently an incidental byproduct of random initialization rather than a guaranteed outcome of the objective, highlighting the fundamental unidentifiability of the unsupervised setting \citep{HYVARINEN1999429, khemakhem2020variational}. Consequently, the need for flexible methods that can incorporate semi-supervised data for semantic alignment is becoming increasingly critical \citep{locatello2020weakly}.

In this work, we decouple the broad objective of disentanglement into learning features that are both \textbf{(1) mutually independent} and \textbf{(2) semantically meaningful}. To realize this decoupling and address the shortcomings of prior work, we propose a Wasserstein Autoencoder (WAE-MMD) variant \cite{tolstikhin2017wasserstein} that sculpts the latent space's aggregate posterior distribution using the non-parametric Maximum Mean Discrepancy (MMD) \citep{gretton2012kernel}. We also develop the Latent Predictability Score (LPS), an unsupervised empirical lower bound on latent entanglement, as a rejection test for condition (1). Our primary contributions are summarized as follows:

\begin{figure}[t!]
    \centering
    \includegraphics[width=\ifsdm\linewidth\else0.8\linewidth\fi]{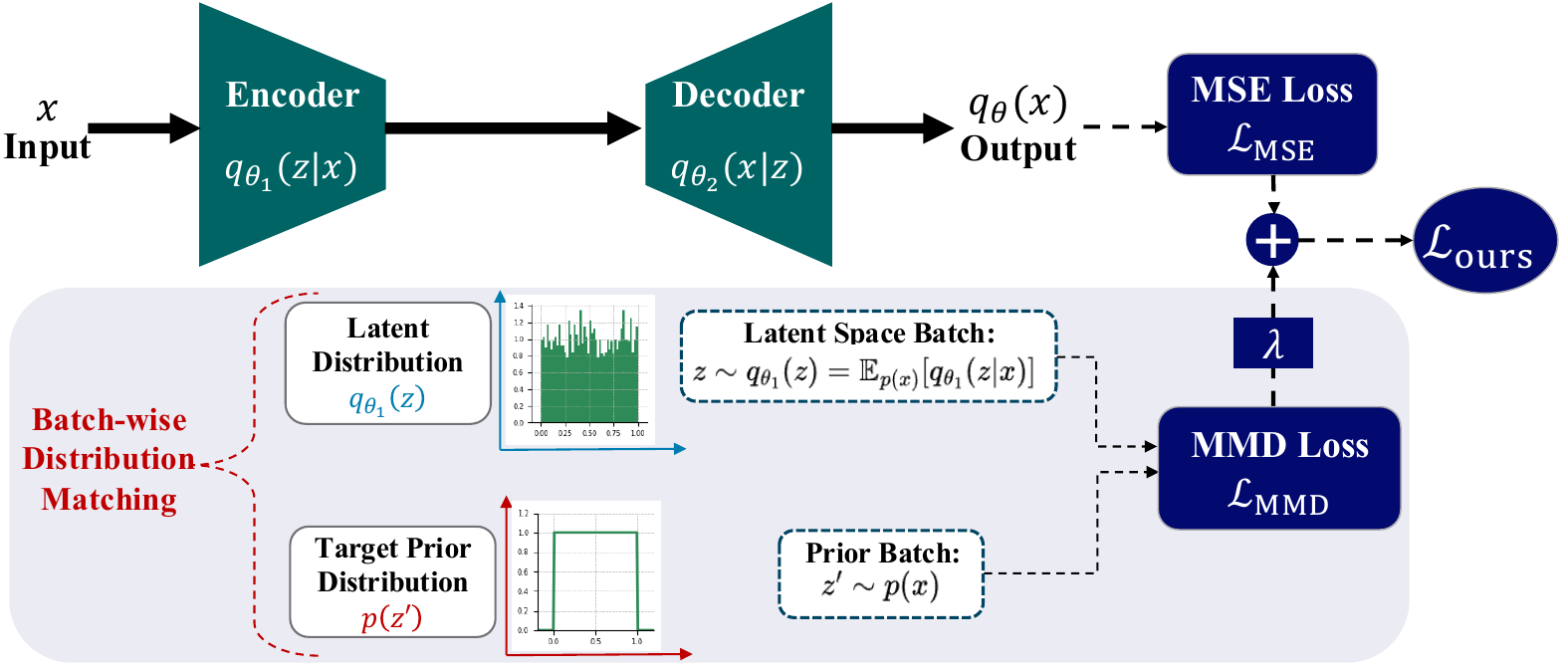}
    \caption{Visualizing the Sculpting Framework.}
    \label{fig:architecture}
\end{figure}

\begin{figure*}[t!]
    \centering
    \begin{subfigure}[b]{0.19\textwidth}
        \centering
        \includegraphics[width=\linewidth]{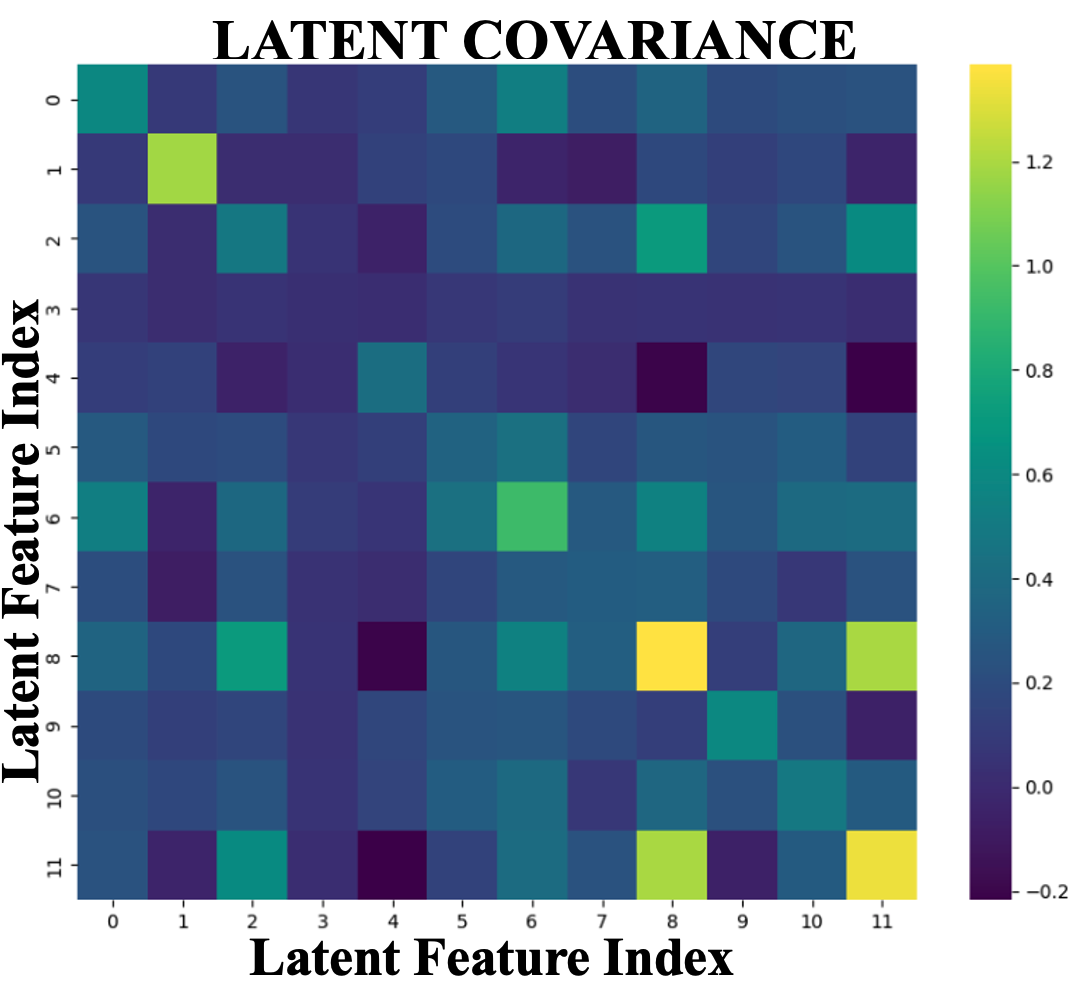}\\
        \vspace{2mm}
        \includegraphics[width=\linewidth]{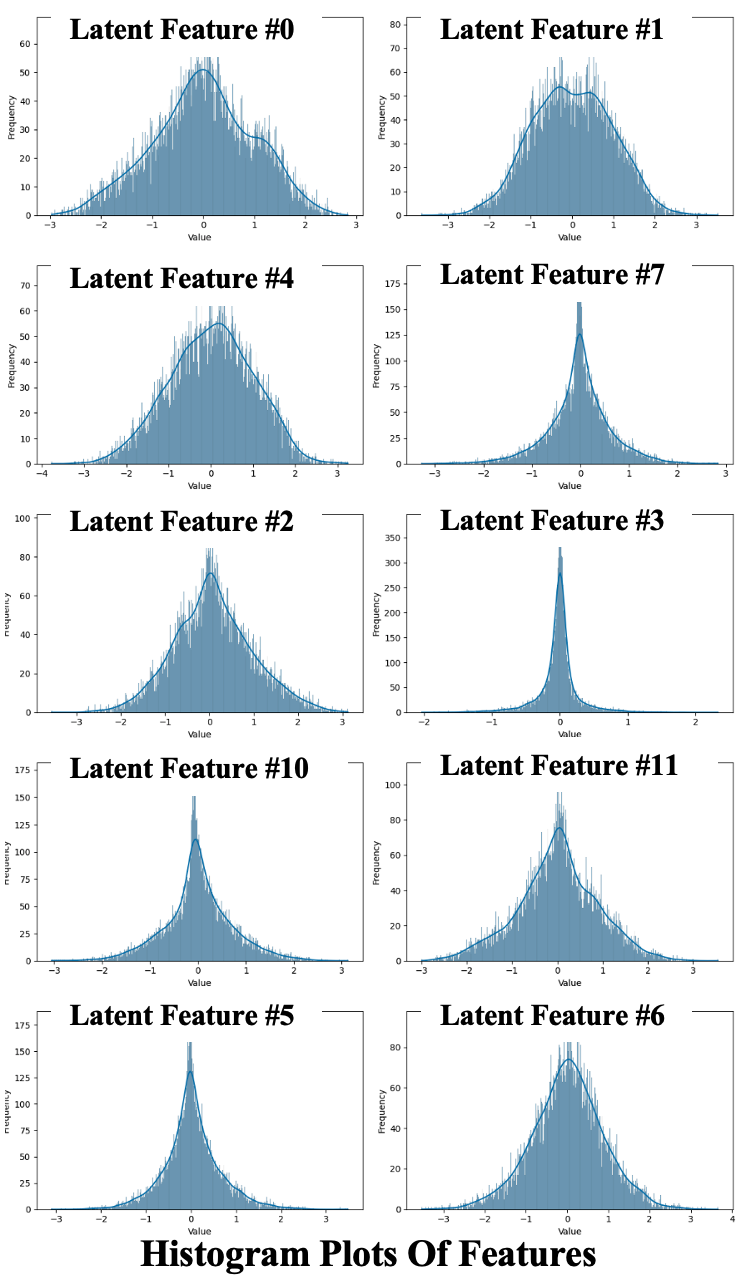}
        \caption{Standard VAE}
        \label{fig:col_vae1}
    \end{subfigure}
    \begin{subfigure}[b]{0.19\textwidth}
        \centering
        \includegraphics[width=\linewidth]{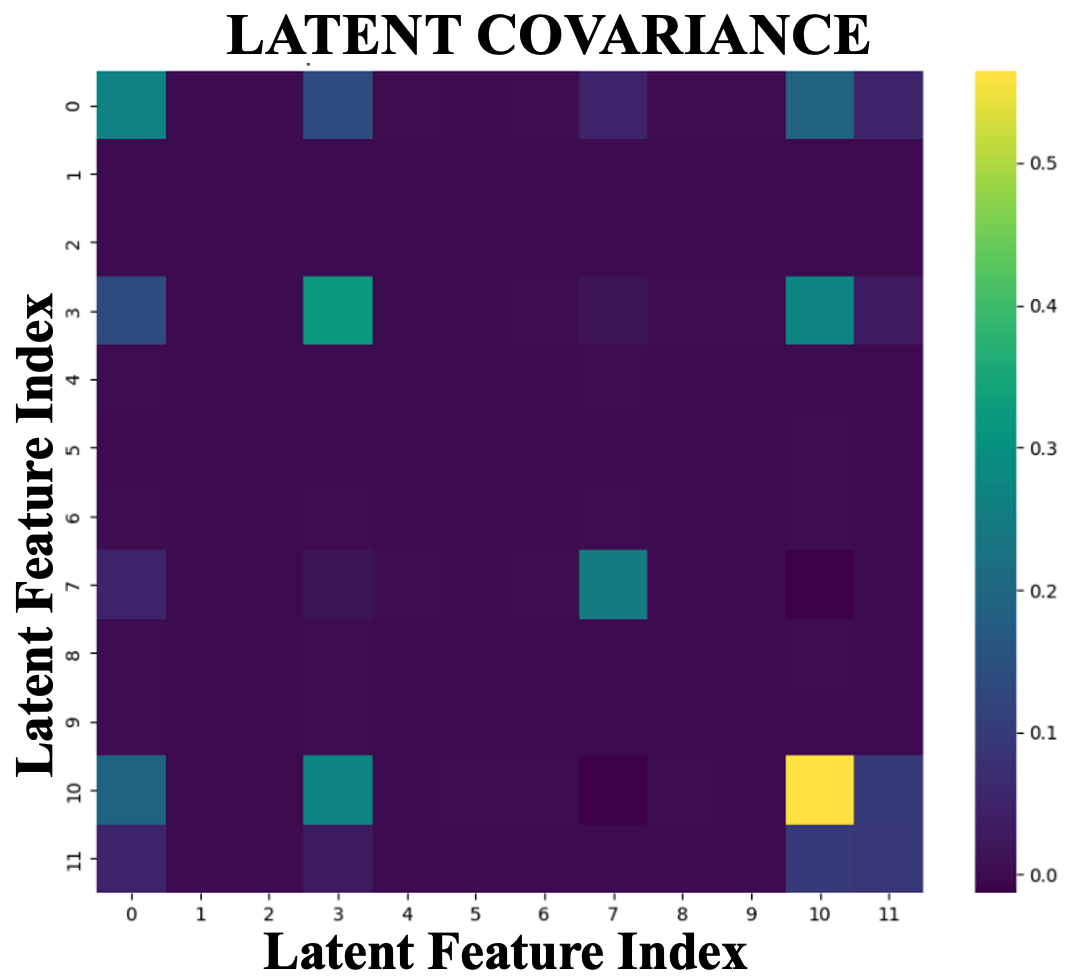}\\
        \vspace{2mm}
        \includegraphics[width=\linewidth]{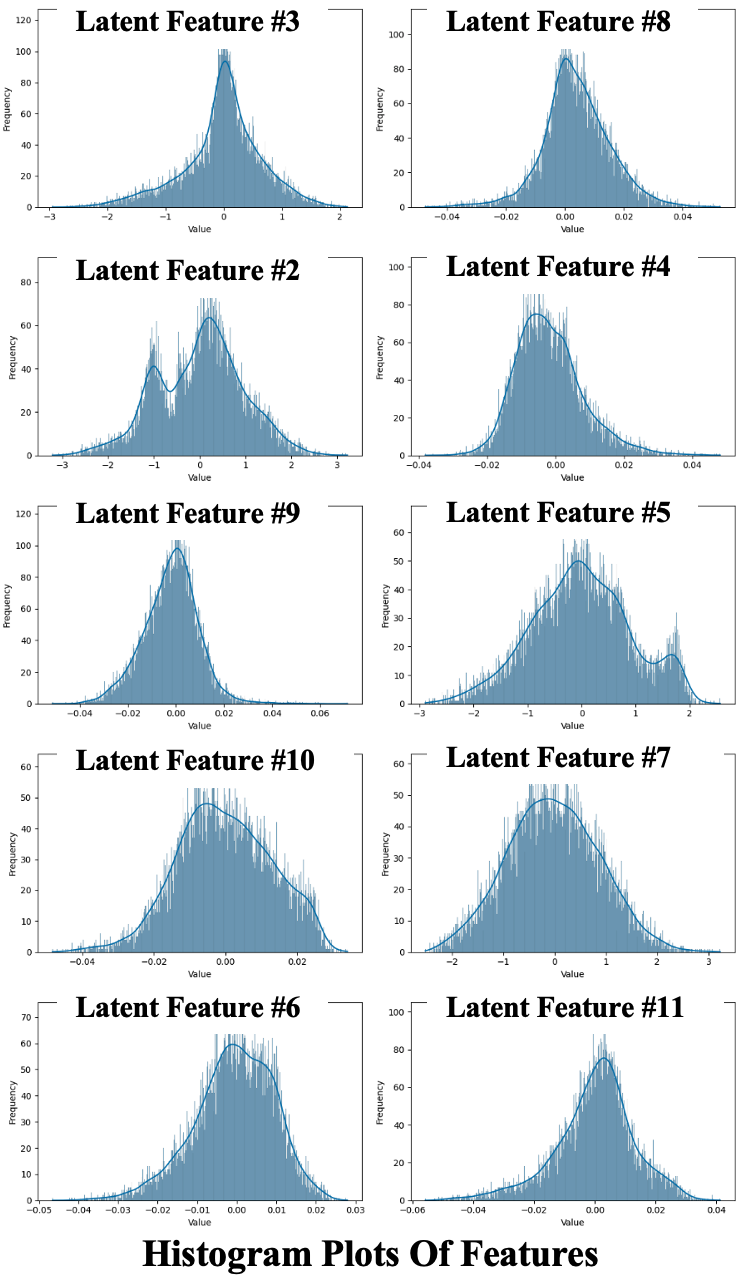}
        \caption{$\beta$-VAE ($\beta=4$)}
        \label{fig:col_vae4}
    \end{subfigure}
    \begin{subfigure}[b]{0.2\textwidth}
        \centering
        \includegraphics[width=0.95\linewidth]{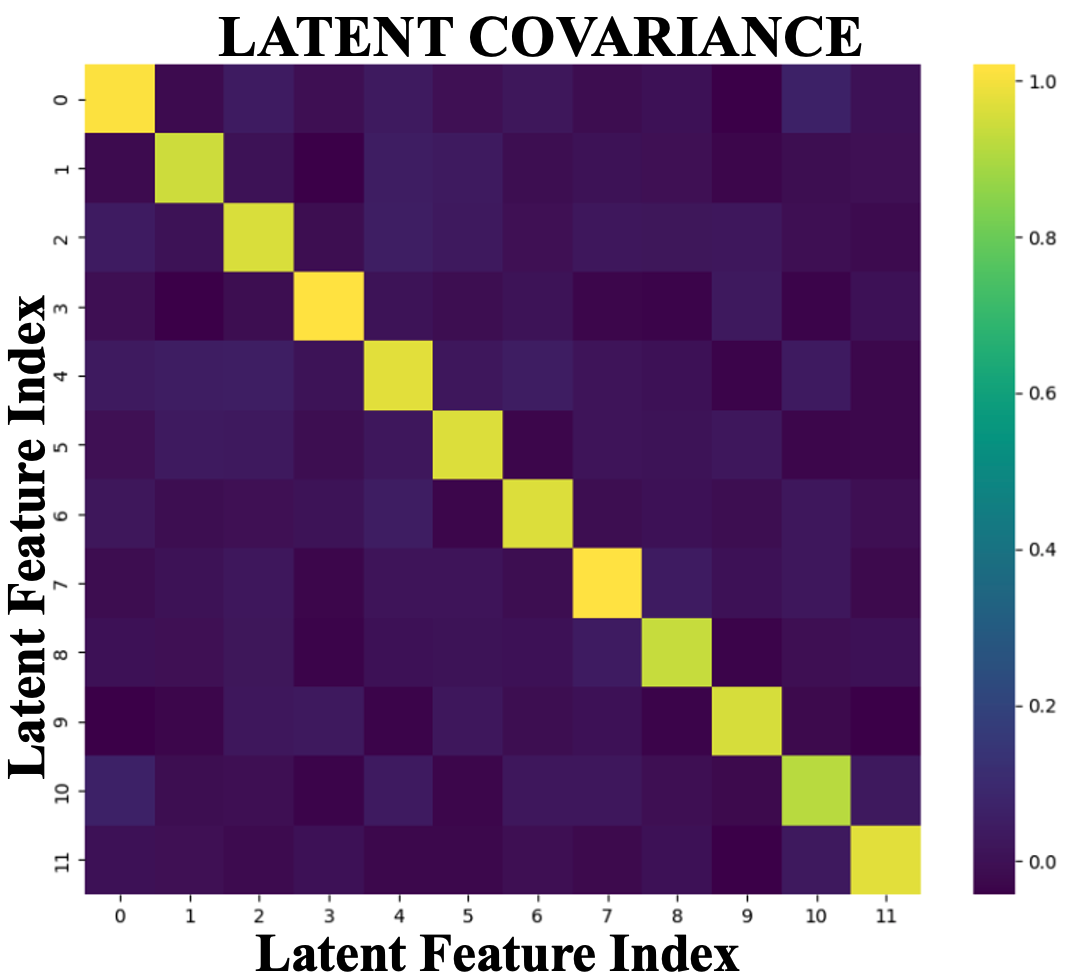}\\
        \vspace{2mm}
        \includegraphics[width=0.95\linewidth]{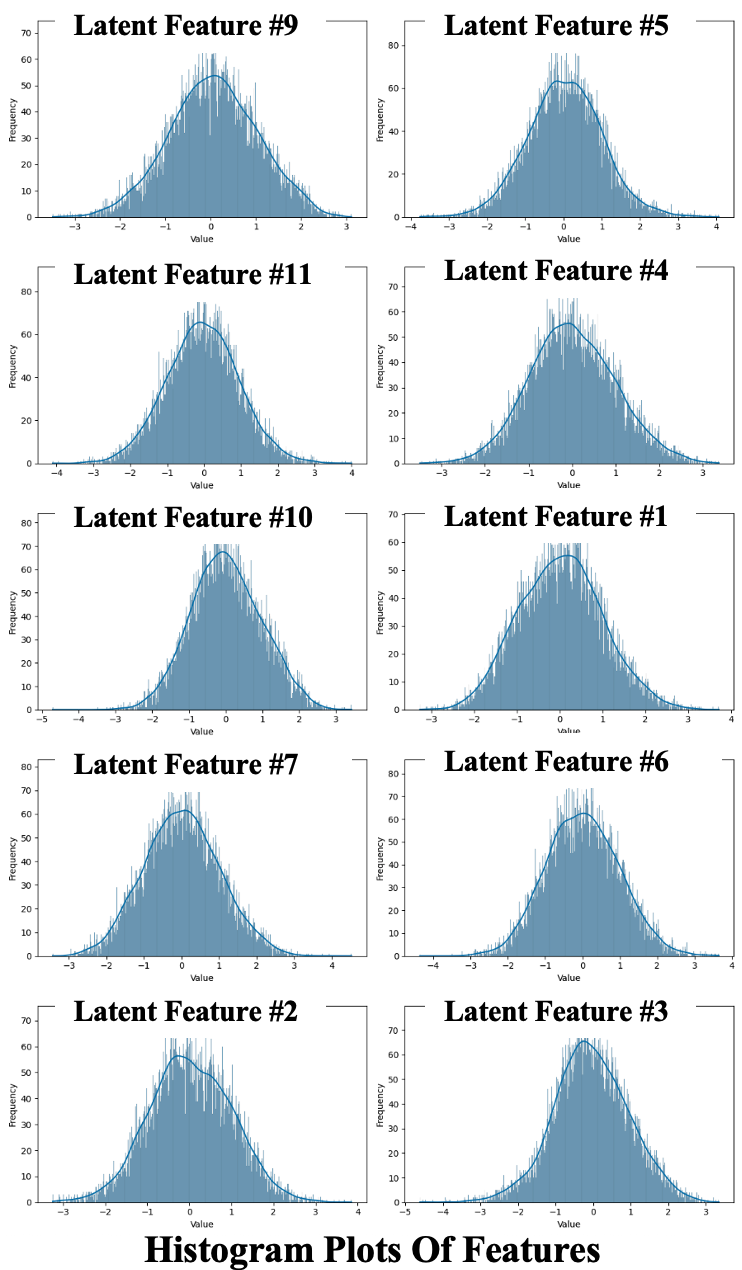}
        \caption{\textbf{SWAE-MMD}}
        \label{fig:col_mmd}
    \end{subfigure}
    \begin{subfigure}[b]{0.19\textwidth}
        \centering
        \includegraphics[width=\linewidth]{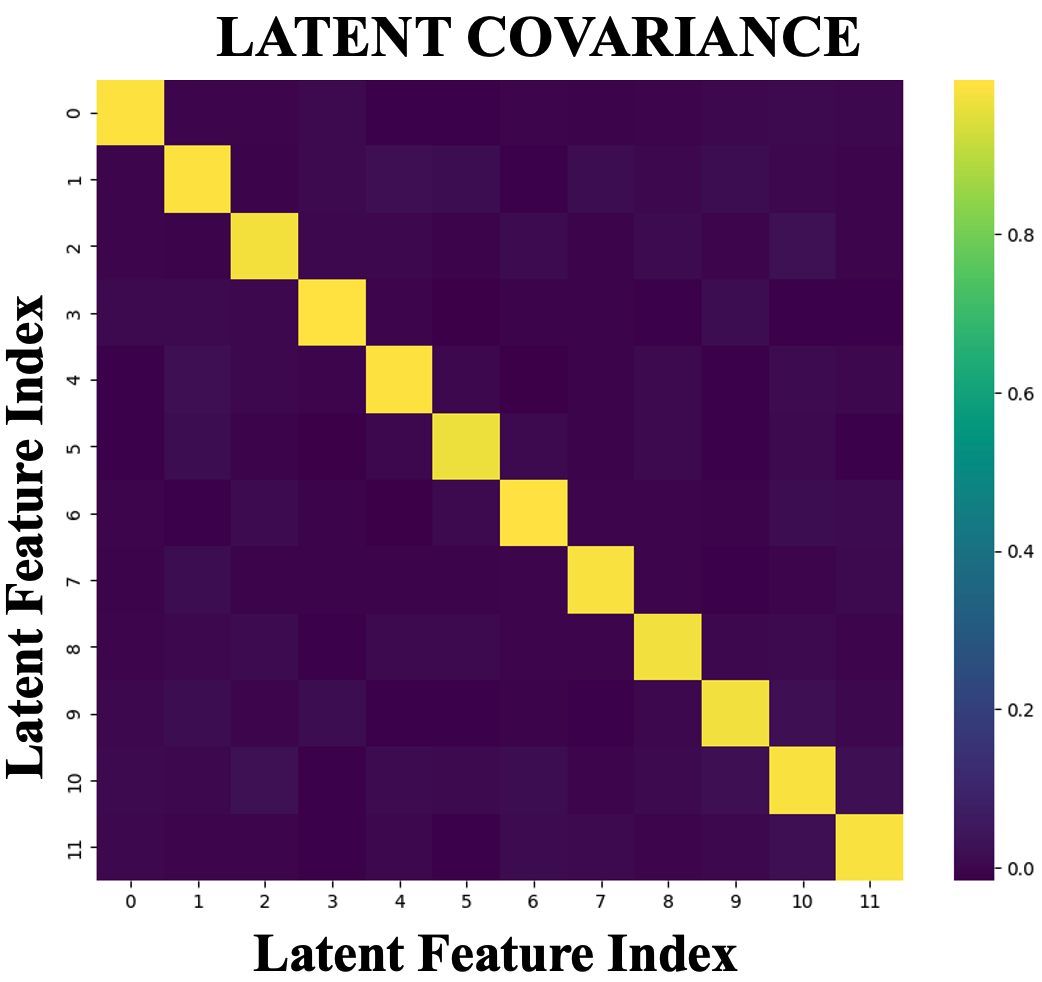}\\
        \vspace{2mm}
        \includegraphics[width=\linewidth]{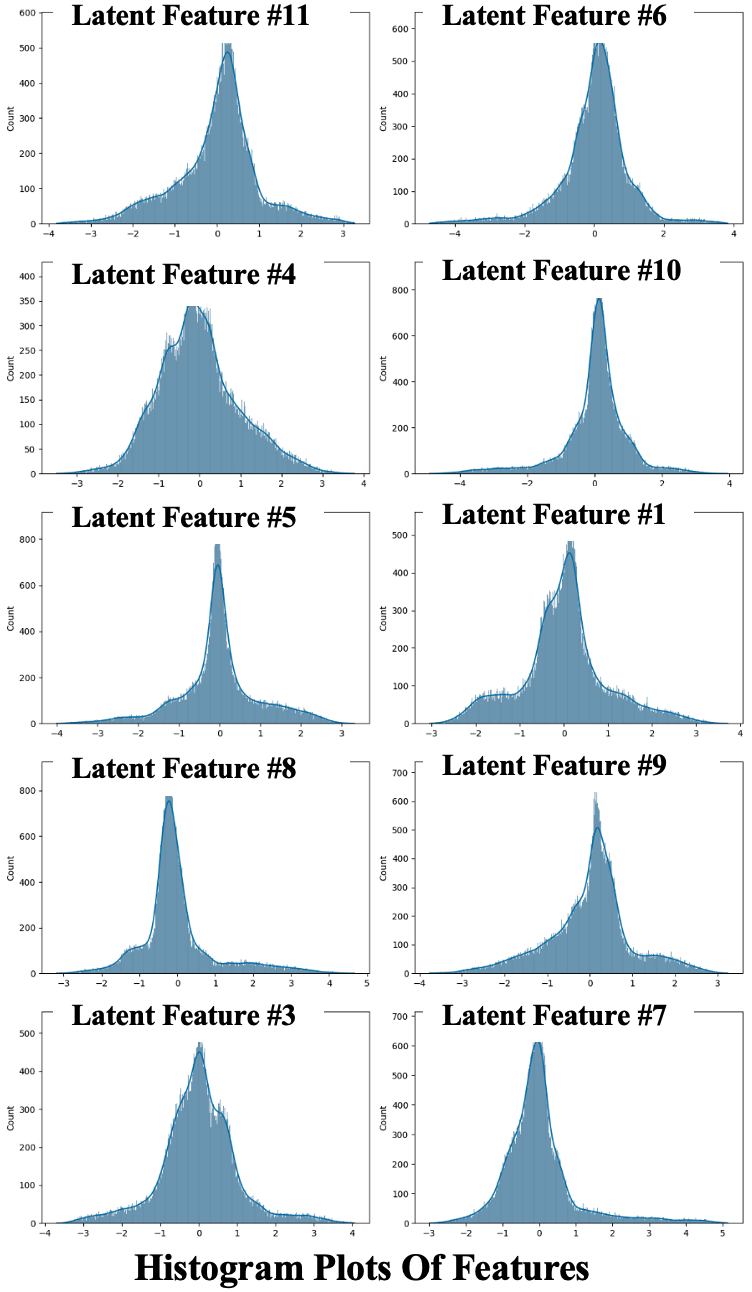}
        \caption{AE-KLD}
        \label{fig:col_kld_gauss}
    \end{subfigure}
    \begin{subfigure}[b]{0.19\textwidth}
        \centering
        \includegraphics[width=\linewidth]{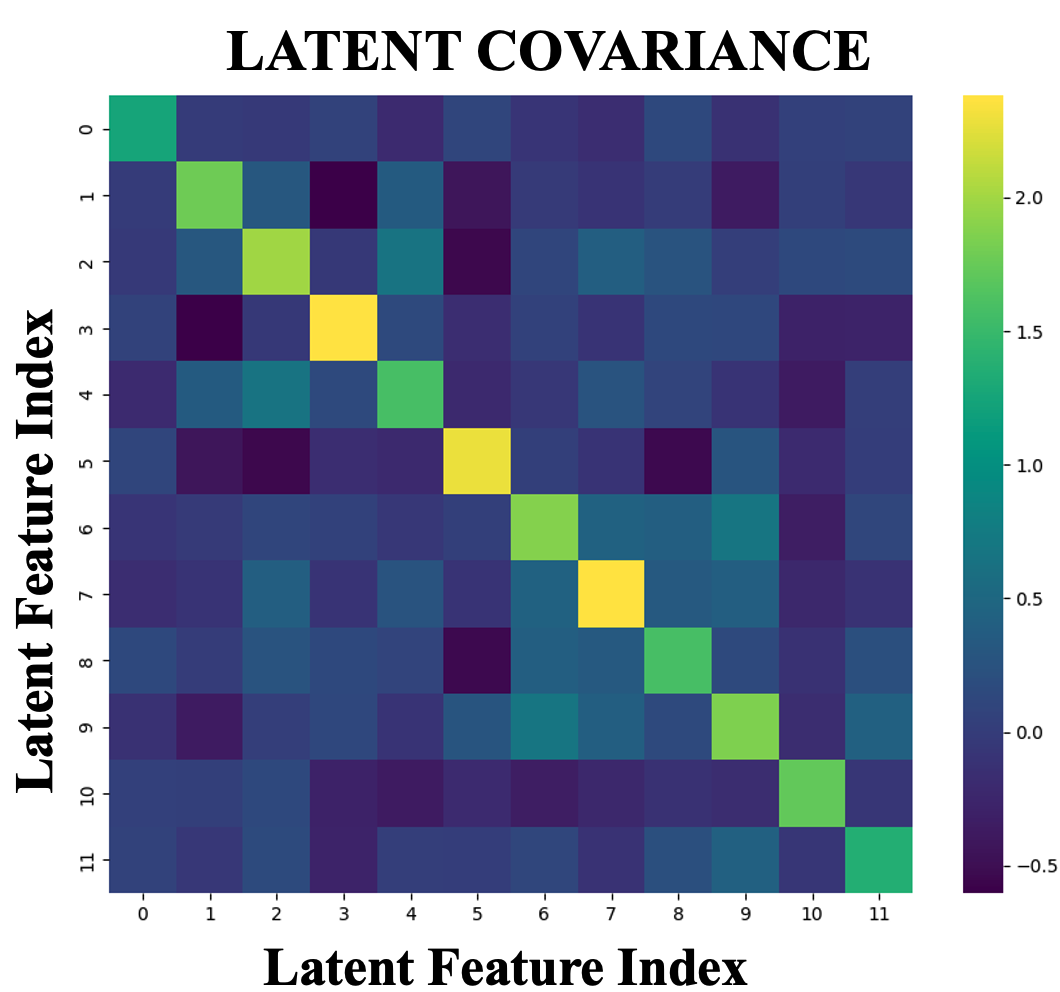}\\
        \vspace{2mm}
        \includegraphics[width=\linewidth]{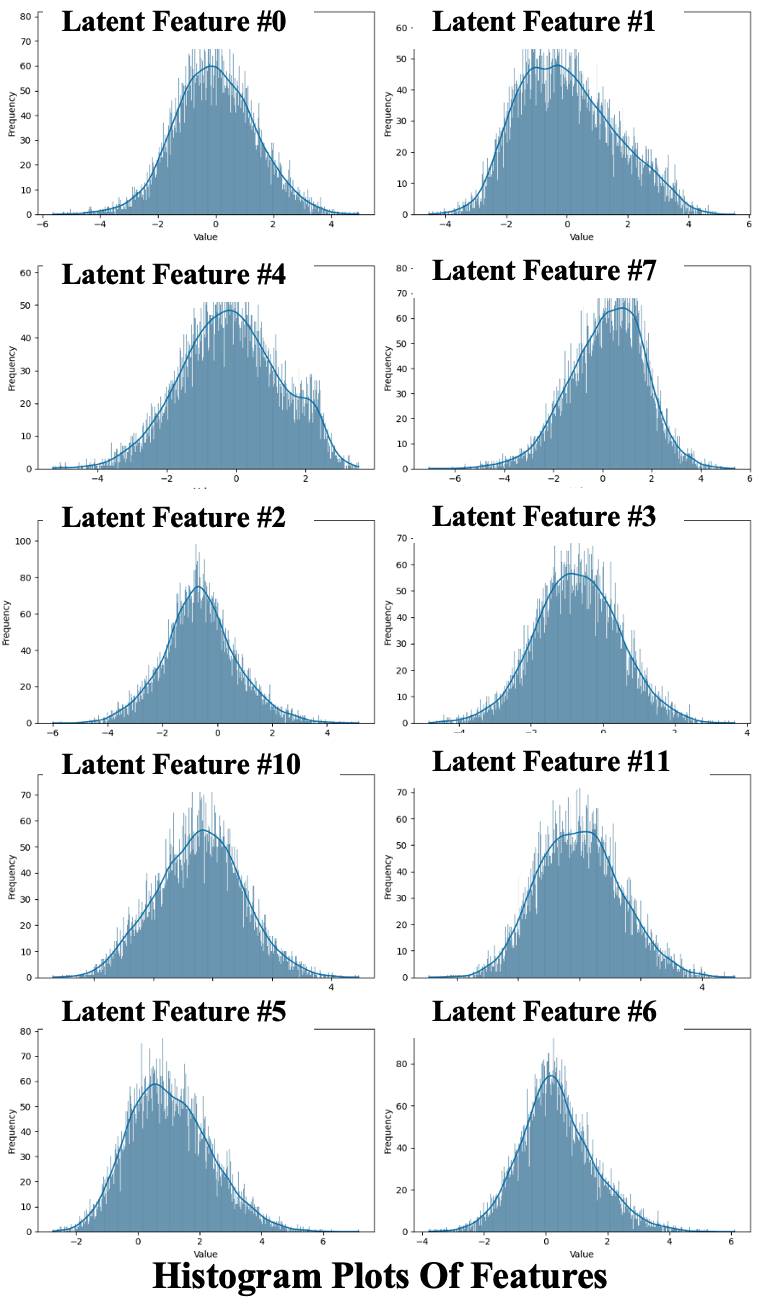}
        \caption{Standard AE}
        \label{fig:col_ae}
    \end{subfigure}

    \caption{
        \textbf{Visual comparison of learned latent distributions on MNIST for key baselines.} Each column corresponds to a different model, showing the covariance matrix between the latent space features over the dataset (top row) and the histogram plots for the marginal distribution of the latent space features over the whole dataset (bottom row). The VAE-based models (a, b) and the standard AE (e) fail to enforce either independence or the target Gaussian geometry. While a batch-wise KLD regularizer (d) achieves a diagonal covariance, its parametric nature prevents it from correctly shaping the marginal distributions leading to co-dependence (LPS equal to 0.97). Only the MMD-regularized model (c) successfully enforces both properties, producing a representation that is truly independent and matches the target prior.
    }
    \label{fig:full_visual_comparison}
\end{figure*}

\begin{itemize}
    \item \textbf{An Unsupervised Empirical Lower Bound For Entanglement.} We introduce the Latent Predictability Score (LPS), closely related to the adversary-based training objective of the NashAE \cite{yeats2022nashae}, as a stable, unsupervised empirical lower bound on the mutual dependence of a latent space. We frame it as a necessary ``rejection test": a high LPS score (high predictability) allows us to reliably reject the hypothesis that a representation is disentangled.
    \item \textbf{A Critical Failure Analysis of Popular Baselines.} We first provide quantitative evidence in Table~\ref{tab:main_results_revised_v2} that the dominant VAE-based framework (including $\beta$-VAE, FactorVAE, and $\beta$-TCVAE) fails to satisfy (1), with LPS reporting significant entanglement. This is further validated visually in Fig~\ref{fig:full_visual_comparison}, where we observe VAE methods consistently fail to enforce the target factorized prior. We attribute this failure primarily to the limitations of the parametric KL-divergence regularizer. Second, LPS reports significant entanglement for the NashAE, even when its training adversary indicates mutual independence. This discrepancy exposes the optimization instability inherent to adversarial frameworks, where the inner-loop adversary fails to converge sufficiently to act as a reliable proxy for the true latent dependence.
    \item \textbf{A Unified Framework for Independence, Alignment, and Distribution Shaping.} We propose the Programmable Priors, an MMD-based sculpting regularizer inspired by the Wasserstein Autoencoder that simultaneously addresses three critical desiderata: \textbf{Statistical Independence~(1).} It explicitly enforces a factorized prior, achieving a 70\% reduction in latent space entanglement compared to the next best baseline, FactorVAE \cite{kim2018disentangling}; \textbf{Semantic Alignment~(2).} It provides a semi-supervised mechanism to mitigate the NICA unidentifiability problem by aligning latents to \textit{pseudo-labels}; \textbf{Distribution Shaping~(3).} It enables the explicit sculpting of complex aggregate marginal priors. We demonstrate this through Quantization-Aware Training (QAT) for data compression, where the effectiveness of the latent quantizer hinges on the aggregate marginal posterior matching the quantizer's expected distribution.
    We further demonstrate that our framework achieves all three goals with minimal trade-offs in reconstruction quality.
\end{itemize}
To facilitate reproducibility and encourage further research, our complete codebase, including all experiments and our novel metric, is publicly available \href{https://github.com/Quentin-Fruytier/Sculpting-Latent-Spaces-With-MMD}{here}.

\section{Background And Prior Work}
\label{sec:background}

This section provides the necessary background on the generative models, theoretical challenges, and statistical tools that form the context for our work. We begin by formalizing the autoencoder and variational autoencoder paradigms before discussing the core challenges of disentanglement and introducing the Maximum Mean Discrepancy as a powerful non-parametric regularizer.

\textbf{Latent Variable Models and the Variational Autoencoder.} In its simplest form, a latent variable model is defined by a joint distribution $p(x,z)$ of observed data $x$ and latent variables $z$. In deep learning, this relationship is parameterized by a neural network, often subdivided into an encoder $q_{\theta_1}(z|x)$ and a decoder $q_{\theta_2}(x|z)$. A natural training objective is to maximize the log-likelihood, $\mathbb{E}_{p(x)}[\log q_{\theta}(x)]$. Training these two components end-to-end with this objective is the basis of the classical Autoencoder (AE) paradigm.

Prior work has largely focused on frameworks that provide the ability to specify and constrain the latent space's posterior distribution $q_{\theta_1}(z|x)$. Re-writing the log-likelihood as $\mathbb{E}_{p(x)}[\log \mathbb{E}_{q_{\theta_1}(z|x)} [q_{\theta_2}(x|z)]]$ provides the most obvious objective to maximize, but direct optimization in this form is generally intractable. As a solution, the Variational Autoencoder (VAE) framework provides a popular and effective solution that consists in maximizing the Evidence Lower Bound (ELBO) of the log-likelihood:
\begin{equation}
\label{eqn:ELBO_loss}
    \mathcal{L}_{\text{ELBO}}(x) := \underbrace{\mathbb{E}_{q_{\theta_1}(z|x)}[\log q_{\theta_2}(x|z)]}_{\text{Reconstruction Loss}} - \underbrace{KL(q_{\theta_1}(z|x)||p(z))}_{\text{Regularization Loss}}
\end{equation}
In a standard VAE, the encoder outputs the parameters of a diagonal Gaussian, a mean vector $\mu(x)$ and a variance vector $\sigma^2(x)$. A latent code is then sampled via the reparameterization trick: $z = \mu(x) + \sigma(x) \cdot \epsilon$, where $\epsilon \sim \mathcal{N}(0,I)$. The regularization term becomes the per-sample KL-divergence between the approximate posterior and the prior, which is typically a standard normal distribution: $KL(N(\mu(x),\sigma(x))||N(0,I))$.

\textbf{The Challenge of Unsupervised Disentanglement.} In the context of unsupervised disentanglement, the overarching goal is to learn a representation where the latent vector $z$ is factorized, meaning its components are statistically independent. The 'Regularization Loss' of the VAE is thought to encourage this by forcing the posterior distribution towards a $0$ mean and identity covariance Gaussian. In fact, seminal works like $\beta$-VAE \citep{higgins2017beta}, FactorVAE \citep{kim2018disentangling}, and $\beta$-TCVAE \citep{chen2018isolating} all build on this principle through distinct tweaks to the $KL$ term.  However, Figure~\ref{fig:full_visual_comparison} (a-b) demonstrates that this mechanism is unreliable and fail to enforce the desired factorized distribution on the aggregate posterior: If $q_{\theta_1}(z|x) \sim N(0,I)$ as is the objective of the VAE framework, then so should the aggregate $q_{\theta_1}(z) = \mathbb{E}_p(x)[q_{\theta_1}(z|x)] \sim N(0,I)$. The non-identity covariance matrix demonstrates substantial entanglement while the non bell-curved marginals prove the learned representations cannot possibly be Gaussian. 

Other popular alternative paradigms exist, such as using adversarial training \citep{chen2016infogan, yeats2022nashae} or imposing direct geometric constraints like the learning of orthogonal group action features \citep{cha2023orthogonality}. Notably, NashAE \cite{yeats2022nashae} employs an adversary trained in an inner loop to estimate the model’s latent entanglement (using a method similar to our LPS) and subsequently penalizes the objective function.

In the end, unsupervised learning of latent feature representations that are both (1) mutually independent, and (2) semantically meaningful is complicated by deep theoretical and practical challenges. The unidentifiability of NICA \citep{HYVARINEN1999429, khemakhem2020variational}, which posits that independent and interpretable representations are not unique, thus implying that some level of supervision is necessary for (2) \citep{locatello2019challenging, locatello2020weakly}.

\textbf{The Maximum Mean Discrepancy.} An alternative to the KL-divergence for estimating the distance between two distributions is the Maximum Mean Discrepancy (MMD) \citep{gretton2012kernel}. The core intuition of MMD is to represent distributions as single mean embeddings in a high-dimensional Reproducing Kernel Hilbert Space (RKHS). The MMD between two distributions, $P$ and $Q$, is then simply the RKHS norm of the difference between their mean embeddings, $\mu_P$ and $\mu_Q$:
\begin{equation}
    \text{MMD}(P, Q) = \|\mu_P - \mu_Q\|_{\mathcal{H}}
\end{equation}
If the mean embeddings are identical, the MMD is zero, and the distributions are identical. This is operationalized via a kernel function, $k(\cdot, \cdot)$, such as a Gaussian RBF kernel.

In practice, finite set of samples $Z = \{z_1, \dots, z_m\} \sim P$ and $Z' = \{z'_1, \dots, z_n\} \sim Q$ are used to estimate the mean embeddings. The MMD is then the distance between these empirical estimates:
\begin{equation}
    \text{MMD}(P, Q) \approx \left\| \frac{1}{m}\sum_{i=1}^{m} \phi(z_i) - \frac{1}{n}\sum_{j=1}^{n} \phi(z'_j) \right\|_{\mathcal{H}}
\end{equation}
where $\phi$ is the feature map into the RKHS associated with the kernel $k$. Using the kernel trick, $\langle \phi(a), \phi(b) \rangle_{\mathcal{H}} = k(a, b)$, we can expand this into a computable form. The standard biased empirical estimator for the squared MMD is given by the sum of three terms:
\ifsdm
\begin{multline}
    \label{eq:mmd_estimator}
\text{MMD}^{2}(P, Q) = \frac{1}{m(m-1)}\sum_{i\neq j} k(z_i, z_j) \\
- \frac{2}{mn}\sum_{i,j} k(z_i, z'_j) + \frac{1}{n(n-1)}\sum_{i\neq j} k(z'_i, z'_j).
\end{multline}
\else
\begin{equation}
    \label{eq:mmd_estimator}
\text{MMD}^{2}(P, Q) = \frac{1}{m(m-1)}\sum_{i\neq j} k(z_i, z_j) - \frac{2}{mn}\sum_{i,j} k(z_i, z'_j) + \frac{1}{n(n-1)}\sum_{i\neq j} k(z'_i, z'_j).
\end{equation}
\fi

In this paper, we utilize the popular biased version of this estimator where the condition $i \neq j$ is not enforced in the summation. The principle of using an MMD penalty as a more stable substitute for other divergence function in Neural Networks is a well-established concept, primarily explored for generative modeling, domain adaptation, and representational knowledge distillation \citep{tolstikhin2017wasserstein, li2017mmd, long2017deep, huang2017like, zhao2019infovae}. In the section that follows, we use the MMD regularizer as a precise instrument to inject a wide range of explicit, user-defined inductive biases.

%% file: Sections/Method.tex
\section{The MMD Regularizer for Sculpting the Latent Space}
\label{sec:method}

Our core technical contribution is the formulation of a deterministic MMD-based sculpting mechanism, Sculpting Wasserstein Auto Encoder - MMD (SWAE-MMD), to impose a specific dependence structure on the aggregate posterior whilst injecting semantic alignment to address the unidentifiability of unsupervised disentanglement. Formally, we desire to shape the aggregate posterior, $z\sim q_{\theta_1}(z) = \mathbb{E}_{p(x)}[q_{\theta_1}(z|x)]$, to match a prior distribution $p(z)$ while imposing posterior structure programmed as $q_{\theta_1}(z|x) = p(u|x)$, where $u$ is the semantic feature set to align to. To achieve this goal, we utilize an element-wise auxiliary network (for example, a single hidden-layer MLP) to parametrize the relationship between the posterior structure of $z$ and the semantic features as $z|x \sim f_\phi(u)|x$. We then compare the two joint distributions $(q_{\theta_1}(z, f_\phi(u)))$ and $(p(z, f_\phi(u)))$ with the MMD. With this in mind, our framework proposes to maximize the following lower bound on the log-likelihood:

\ifsdm
\begin{multline}
    \label{eqn:MMDFramework}
    \mathcal{L}_{\text{ours}} := \underbrace{\mathbb{E}_{p(x)}[\log q_{\theta}(x)]}_{\textrm{Log-likelihood}} \\
    - \lambda \cdot \underbrace{\text{MMD}^2\Big( q_{\theta_1}(z, f_\phi(u)), p(z,f_\phi(u)) \Big)}_{\textrm{MMD Regularization}},
\end{multline}
\else
\begin{equation}
    \label{eqn:MMDFramework}
    \mathcal{L}_{\text{ours}} := \underbrace{\mathbb{E}_{p(x)}[\log q_{\theta}(x)]}_{\textrm{Log-likelihood}} - \lambda \cdot \underbrace{\text{MMD}^2\Big( q_{\theta_1}(z, f_\phi(u)), p(z,f_\phi(u)) \Big)}_{\textrm{MMD Regularization}}.
\end{equation}
\fi

where $\lambda$ is a hyperparameter that controls the strength of the regularization. We learn parameters $\theta, \phi$ to maximize this objective. By minimizing this joint MMD, the encoder and auxiliary networks are forced to learn a mapping where specific latent dimensions are factorized and locked to specific semantic factors. 

This mechanism effectively unifies three distinct representation engineering goals into a single framework: (1) \textbf{Statistical Independence} via a factorized prior $p(z)$; (2) \textbf{Semantic Alignment} via the auxiliary mapping $f_\phi(u)$; and (3) \textbf{Distribution Shaping} via the specific geometry of the target prior. 

Our work draws inspiration from the WAE-MMD \citep{tolstikhin2017wasserstein} which proposed to use MMD for generative modeling. Instead, we shift the focus from generative modeling to representation learning and extend their framework to allow for explicit sculpting of the latent space. Indeed, if there are no desired semantic features to align to and the aggregate posterior target is simply chosen as $p(z)~N(0,I)$, then the objective simplifies to that of the standard WAE-MMD:
\begin{equation}
\label{equation: WAE-MMD}
    \mathcal{L}_{\text{WAE}} := \mathbb{E}_{p(x)}[\log q_{\theta}(x)] - \lambda \cdot \text{MMD}^2(q_{\theta_1}(z),p(z)).
\end{equation}

\textbf{The Necessity of MMD over Analytic Divergences.} 
A natural question is why we employ the sample-based MMD rather than a more computationally efficient analytic alternative. A common approach for shaping the aggregate posterior is to apply the $KL$ divergence to the batch statistics of the encoded features. While this method is computationally faster than the quadratic-time MMD estimator, its reliance on parametric assumptions constitutes a critical flaw. 

As demonstrated visually in Figure~\ref{fig:full_visual_comparison}(d), the batch-wise KL regularizer (replacing the MMD loss by the KL divergence in \eqref{equation: WAE-MMD}) successfully enforces a diagonal covariance matrix (matching the first two moments) but fails to constrain the higher-order moments required to shape the marginals into Gaussians. This failure leads to substantial entanglement; we report an LPS score of $0.97$ for the KL-based method, indicating that the features remain almost completely co-dependent despite the regularization. Consequently, while the non-parametric MMD is computationally heavier due to its sample-based nature, it is strictly necessary for this task. Unlike the KL divergence, the MMD allows us to ``program'' arbitrary, non-analytic priors (ranging from simple Uniform distributions to the complex, entangled geometries demonstrated in our latent copying experiment in Appendix \ref{app:KnowledgeTransfer}) and ensures the model matches the entire distribution geometry, not just the summary statistics.

%% file: Sections/ExpnResults.tex
\section{Experiments and Results}
\label{sec:experiments}
\begin{figure*}[t!]
    \centering
    \begin{subfigure}[b]{0.19\textwidth}
        \centering
        \includegraphics[width=\textwidth]{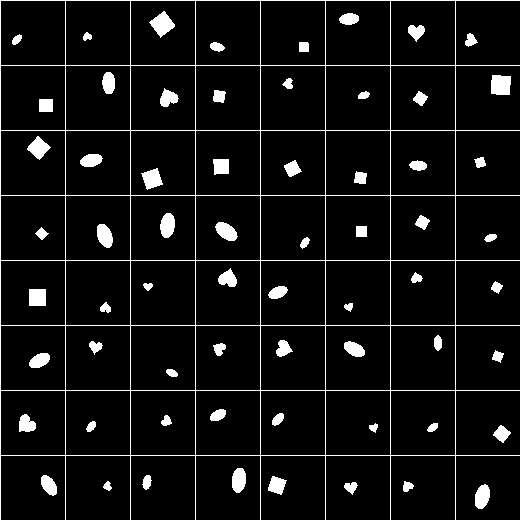}
        \caption{dSprites}
        \label{fig:ex_dsprites}
    \end{subfigure}
    \hfill 
    \begin{subfigure}[b]{0.19\textwidth}
        \centering
        \includegraphics[width=\textwidth]{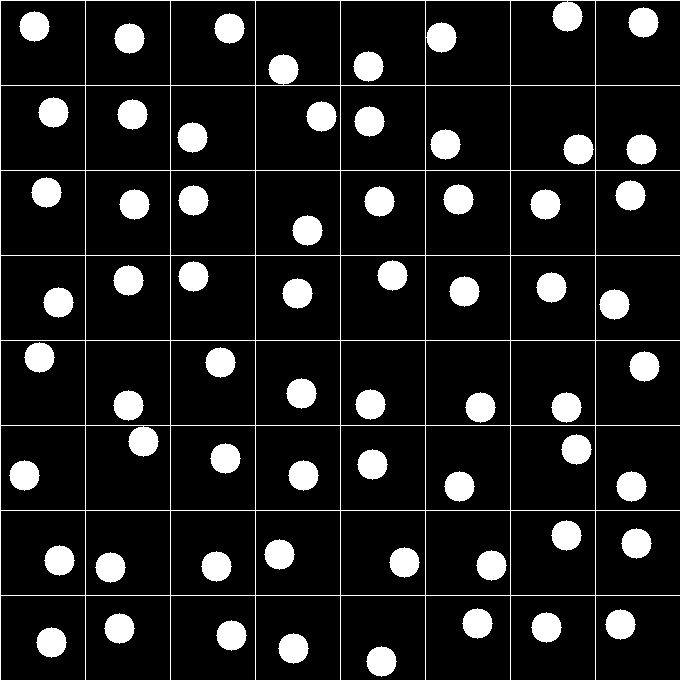}
        \caption{XY}
        \label{fig:ex_xy}
    \end{subfigure}
    \hfill 
    \begin{subfigure}[b]{0.19\textwidth}
        \centering
        \includegraphics[width=\textwidth]{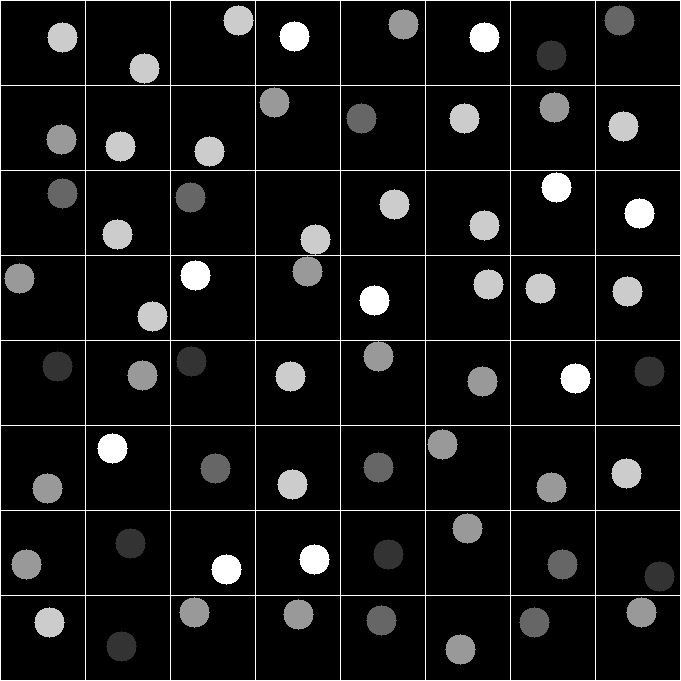}
        \caption{XYC}
        \label{fig:ex_xyc}
    \end{subfigure}
    \hfill 
    \begin{subfigure}[b]{0.19\textwidth}
        \centering
        \includegraphics[width=\textwidth]{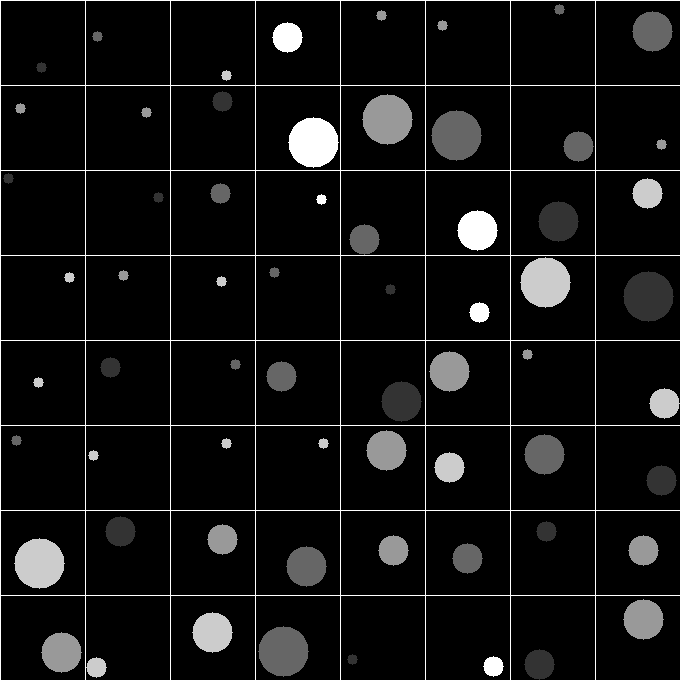}
        \caption{XYCS}
        \label{fig:ex_xycs}
    \end{subfigure}
    \hfill 
    \begin{subfigure}[b]{0.19\textwidth}
        \centering
        \includegraphics[width=\textwidth]{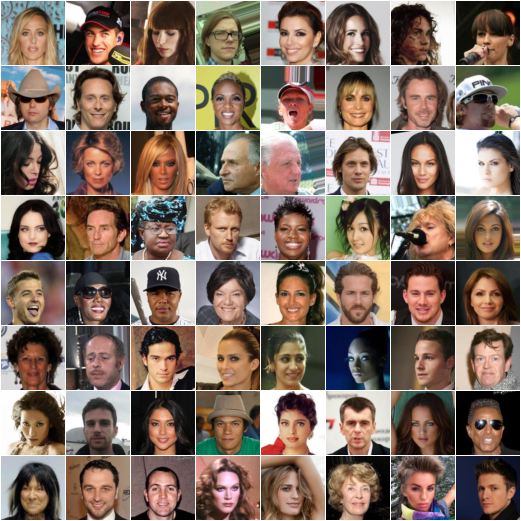}
        \caption{CelebA}
        \label{fig:ex_celebA}
    \end{subfigure}
    
    \caption{Example images (represented in an $8\times8$ grid) from the synthetic datasets used in our experiments. From left to right: (a) dSprites, with factors like shape, scale, and orientation; (b-d) The XY family of datasets, with factors for position (XY), color (C), and shape (S); (c) The CelebA dataset consisting of faces with 40 different semantic factors.}
    \label{fig:dataset_examples}
\end{figure*}

\begin{table*}[t!]
\centering
\caption{
    Main quantitative results comparing our unsupervised MMD-based Autoencoder against baselines (the latent space dimension is specified next to the dataset name). Each cell presents two key metrics: \textbf{LPS-lgbm ($R^2$)} | \textit{reconstruction} \textbf{NMSE (dB)} where lower values are better ($\downarrow$). The best performance for each metric on each dataset is highlighted in \textbf{bold}. Averaged over all $8$ datasets, AE-MMD obtains a 70\% reduction in detected latent space entanglement compared to Factor VAE, the next best baseline, all while maintaining a better image reconstruction quality.
}
\label{tab:main_results_revised_v2}
\resizebox{\textwidth}{!}{%
\begin{tabular}{@{}l*{16}{c}@{}}
\toprule
& \multicolumn{2}{c}{ImageNet ($d=64$)} & \multicolumn{2}{c}{CIFAR10 ($d=64$)} & \multicolumn{2}{c}{MNIST ($d=12$)} & \multicolumn{2}{c}{dsprites ($d=5$)} & \multicolumn{2}{c}{XY ($d=2$)} & \multicolumn{2}{c}{XYC ($d=3$)} & \multicolumn{2}{c}{XYCS ($d=4$)} & \multicolumn{2}{c}{celebA ($d=32$)} \\
\cmidrule(lr){2-3} \cmidrule(lr){4-5} \cmidrule(lr){6-7} \cmidrule(lr){8-9} \cmidrule(lr){10-11} \cmidrule(lr){12-13} \cmidrule(lr){14-15} \cmidrule(lr){16-17}
\textbf{Model} & \textbf{LPS} & \textbf{NMSE (dB)} & \textbf{LPS} & \textbf{NMSE (dB)} & \textbf{LPS} & \textbf{NMSE (dB)} & \textbf{LPS} & \textbf{NMSE (dB)} & \textbf{LPS} & \textbf{NMSE (dB)} & \textbf{LPS} & \textbf{NMSE (dB)} & \textbf{LPS} & \textbf{NMSE (dB)}& \textbf{LPS} & \textbf{NMSE (dB)} \\
\midrule
\textbf{SWAE-MMD (ours)} & \textbf{0.03} & \textbf{-11.70} & \textbf{0.04} & \textbf{-15.61} & \textbf{0.22} & \textbf{-11.92} & \textbf{0.04} & \textbf{-17.89} & \underline{-0.05} & \underline{-15.22} & \textbf{-0.01} & \textbf{-16.31} & \textbf{0.11} & \textbf{-16.76} & \underline{0.08} & -14.37 \\
DGAE & 0.77 & -10.3 & 0.69 & \underline{-15.53} & 0.63 & \underline{-11.48} & 0.56 & \underline{-14.61} & 0.25 & -6.74 & 0.50 & -10.94 & 0.59 & -15.02 & 0.48 & \textbf{-14.40} \\
NASHAE & 0.56 & \underline{-11.32} & \underline{0.17} & -15.02 & \textbf{0.22} & -11.32 & 0.10 & -13.48 & \textbf{-0.07} & -13.08 & 0.14 & -14.93 & 0.37 & -13.15 & 0.36 & -13.67 \\
FactorVAE ($\gamma$$=$10.0) & \underline{0.44} & -10.75 & 0.63 & -12.49 & \underline{0.27} & -8.62 & \underline{0.08} & -14.56 & -0.04 & -\textbf{15.36} & \underline{0.00} & \underline{-15.92} & \underline{0.17} & \underline{-15.52} & \textbf{0.07} & \underline{-14.39} \\
$\beta$-TCVAE ($\beta$$=$4.0) & 0.79 & -9.80 & 0.89 & -10.68 & 0.84 & -5.97 & 0.22 & -8.83 & -0.04 & -14.68 & \textbf{-0.01} & -13.72 & \underline{0.17} & -12.40 & 0.59 & -13.10 \\
$\beta$-VAE ($\beta$$=$4) & 0.73 & -9.87 & 0.85 & -10.49 & 0.90 & -4.61 & 0.52 & -8.20 & \underline{-0.05} & -14.46 & \underline{0.00} & -13.33 & 0.28 & -10.98 & 0.53 & -13.30 \\
\midrule
Standard AE & 0.46 & -11.77 & 0.51 & -15.83 & 0.54 & -12.09 & 0.19 & -18.74 & -0.03 & -16.90 & 0.04 & -17.81 & 0.26 & -18.42 & 0.44 & -14.55 \\
Standard VAE & 0.47 & -10.61 & 0.63 & -12.50 & 0.25 & -8.63 & 0.10 & -14.66 & -0.05 & -15.72 & 0.02 & -16.04 & 0.18 & -15.53 & 0.07 & -14.43 \\
\bottomrule
\end{tabular}%
}
\end{table*}
In this section, we motivate and introduce a new unsupervised evaluation metric -- the Latent Predictability Score (LPS) -- to empirically lower bound the entanglement of a latent space. Following this, we present a comprehensive analysis of our Programmable Prior Framework through three key experimental lenses: (1) unsupervised enforcing of mutual independence in the latent space, (2) semi-supervised alignment with semantic factors, and (3) unsupervised aggregate posterior distribution shaping demonstrated through compatibility with downstream quantization tasks.

\subsection{A Novel Unsupervised Metric for Latent Independence}
\label{sec:exp_metric}

\noindent\textbf{Motivation.} A significant limitation of popular disentanglement metrics like the mutual information gap (MIG) \citep{chen2018isolating} or the Disentanglement Completeness Information score (DCI) \citep{eastwood2018framework} is their reliance on access to semantically meaningful factors for evaluation. Such labels are unavailable for most real-world datasets, creating a critical gap for a metric that can directly quantify the quality of a learned representation in a truly unsupervised manner. Moreover, several works at the interface of NICA and disentanglement have highlighted that \textbf{semantically meaningful and disentangled representations are not unique} which considerably degrades the reliability of any supervised metric \cite{locatello2019challenging, locatello2020weakly, khemakhem2020variational}. To address these challenges, we propose to use an unsupervised metric that empirically lower bounds the level of mutual dependence of the learned latent features over the data, directly serving as a rejection test for the satisfiability of step (1) for obtaining disentangled representations.

\noindent\textbf{Introducing the Latent Predictability Score (LPS).} The core intuition is simple and closely related to the training objective of the NashAE \citep{yeats2022nashae}: if a $d$ dimensional latent space with latent features $\{z_1, \dots, z_d\}$ is truly independent, then any given feature $z_i$ should not be predictable from the others, $z_{-i}$. The LPS procedure quantifies this intuition as follows:
\begin{enumerate}
    \item A representative set of latent vectors is obtained by encoding a large sample of data.
    \item For each latent dimension $i$ from $1$ to $d$:
    \begin{enumerate}
        \item The feature $z_i$ is treated as a target label.
        \item A regression model $f_\psi$ is trained to predict $z_i$ using the remaining $d-1$ dimensions, $z_{-i} = \{z_j\}_{j \neq i}$.
        \item The quality of this prediction is evaluated on a withheld test set using the coefficient of determination ($R^2_i$) score.
    \end{enumerate}
    \item The final LPS is the average $R^2$ score across all $d$ dimensions: $\text{LPS} = \frac{1}{d}\sum_{i=1}^d R^2_i$.
\end{enumerate}

\noindent\textbf{Discussion and Formal Analysis.} The final score is a direct measure of latent dependency. A higher score implies that the features are entangled and predictable from one another, while a score approaching zero indicates a high degree of mutual independence. Therefore, a \textbf{lower LPS score is better}. Crucially, we formalize LPS not just as a heuristic, but as a rigorous lower bound on the true mutual dependence of the latent space.

\begin{lemma}[LPS Lower Bound Property]
Let the entanglement, $\eta^2_i$, be the true fraction of variance of latent dimension $z_i$ that is theoretically explainable with the remaining dimensions $z_{-i}$. Then, the coefficient of determination for dimension $i$, denoted $R^2_i$, satisfies:
\begin{equation}
    R^2_i \leq \eta^2_i.
\end{equation}
Consequently, the LPS estimates a lower bound of the true average mutual dependence of the representation.
\end{lemma}

\textit{Proof.} 
Let $\sigma^2_{opt}$ denote the minimal variance estimator of $z_i|z_{-i}$, given as $\mathbb{E}[z_i|z_{-i}]$. It's irreducible error is given as
\begin{equation*}
    \sigma^2_{opt} = \mathbb{E}_{z_{-i}}[\text{Var}(z_i | z_{-i})] = \mathbb{E}_z[(z_i - \mathbb{E}[z_i|z_{-i}])^2.
\end{equation*}
The entanglement of latent feature $z_i$ with respect to $z_{-i}$, the true fraction of the variance of latent dimension $z_i$ that is theoretically explainable by the remaining dimensions $z_{-i}$, is given by
\begin{equation*}
    \eta_i^2 = 1 - \frac{\mathbb{E}_{z_{-i}}[\text{Var}(z_i | z_{-i})]}{\text{Var}(z_i)} = 1 - \frac{\mathbb{E}_z[(z_i - \mathbb{E}[z_i|z_{-i}])^2]}{\mathbb{E}[(z_i - \mathbb{E}[z_i])^2]}. 
\end{equation*}
The LPS score finds a sub-optimal regression model $f_\psi$ with empirical error $\text{MSE}_{\text{LPS}} = \mathbb{E}[(z_i - f_\psi(z_{-i}))^2]$. Since no model can outperform the minimum variance estimator, $\text{MSE}_{\text{LPS}} \geq \sigma^2_{opt}$. Substituting this inequality into the definition of the coefficient of determination yields 
\begin{equation}
    R^2_i = 1 - \frac{\text{MSE}_{\text{LPS}}}{\text{Var}(z_i)} \leq 1 - \frac{\sigma^2_{opt}}{\text{Var}(z_i)} = \eta^2_i.
\end{equation} \hfill $\square$

This lemma establishes LPS as a necessary "rejection test": if a model achieves an LPS of $0.5$, we can mathematically confirm that \textit{at least} 50\% of the latent information is entangled, regardless of how the ground truth factors are defined.

\noindent\textbf{Variants.} To ensure the predictability measure is robust, we employ two powerful and distinct regression models for this task. We therefore report two versions of our metric: \textbf{LPS-lgbm}, which uses the highly effective LightGBM gradient boosting framework \citep{ke2017lightgbm}, and \textbf{LPS-mlp}, which uses a $2$ hidden layers Multi-Layer Perceptron. Details and empirical analysis of this metric compared to classic supervised metrics can be found in Appendix~\ref{app:Metric Analysis}. 

\subsection{Experimental Setup}
\label{sec:exp_setup}

\noindent\textbf{Datasets.} We evaluate our method across a diverse suite of benchmarks to test its scalability and effectiveness. We use synthetic datasets with known semantic factors: \textbf{dSprites} \citep{dsprites17}, \textbf{CelebA} \citep{liu2015faceattributes}, and the simple generative environments \textbf{XY}, \textbf{XYC}, and \textbf{XYCS} introduced in \citep{cha2023orthogonality}, which consist of circles with varying position (XY), color intensity (C), and shape (S). To test performance on complex, real-world data, we use \textbf{MNIST} \citep{deng2012mnist}, \textbf{CIFAR-10} \citep{krizhevsky2009learning}, and \textbf{TinyImageNet} \citep{le2015tiny}.

\noindent\textbf{Baselines.} We compare our MMD-based approach against $5$ established and state-of-the-art methods for unsupervised disentanglement. Our primary baselines are \textbf{$\bf\beta$-VAE} \citep{higgins2017beta}, \textbf{$\bf\beta$-TCVAE} \cite{chen2018isolating}, \textbf{FactorVAE} \cite{kim2018disentangling}, \textbf{DGAE} \citep{cha2023orthogonality}, and \textbf{NashAE} \cite{yeats2022nashae}. We also include a standard \textbf{Autoencoder (AE)} with no explicit disentanglement regularizer and standard \textbf{VAE} as control variables to establish a performance floor.

While our reported experiments used a batch size of $512$ and a regularization coefficient of $\lambda=0.3$, the MMD regularizer proved robust, yielding consistent results across a wide range of these hyperparameters. The implementation, comprehensive experimental results, and ablation studies for the MMD-based sculpting regularizer can be found in Appendix~\ref{app:implementation_details}, \ref{app:Results} and \ref{app:MMD_Robustness} respectively.

\begin{figure*}[h!]
    \centering
    \begin{subfigure}[b]{0.29\textwidth}
        \centering
        \includegraphics[width=\linewidth]{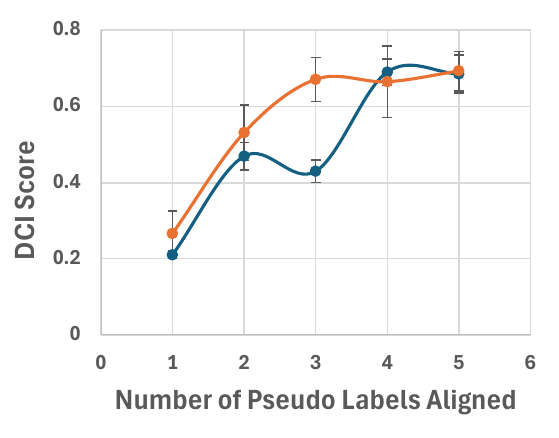}
    \end{subfigure}
    \begin{subfigure}[b]{0.29\textwidth}
        \centering
        \includegraphics[width=\linewidth]{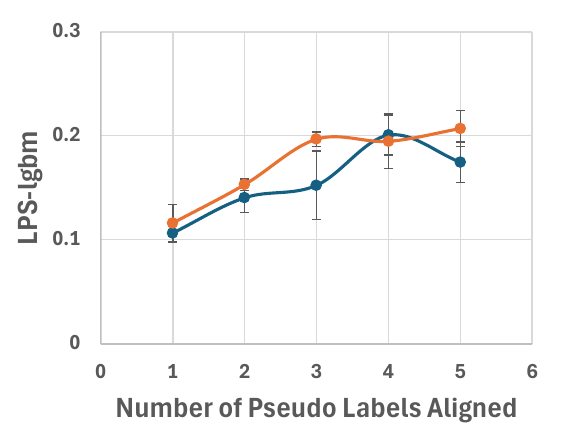}
    \end{subfigure}
    \hfill
    \begin{subfigure}[b]{0.39\textwidth}
        \centering
        \includegraphics[width=\linewidth]{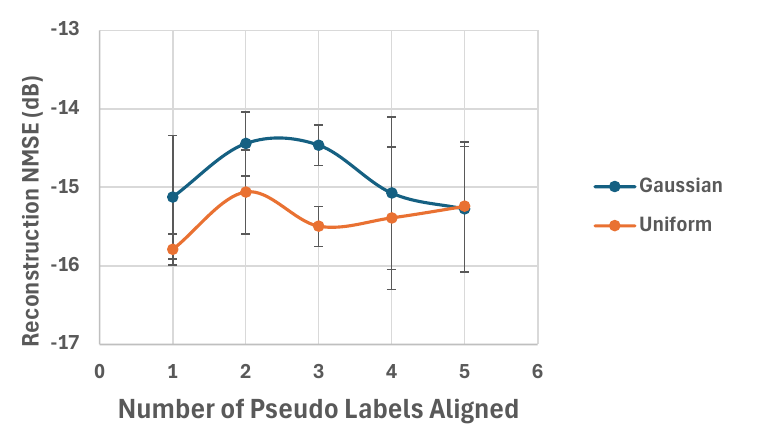}
    \end{subfigure}
    \caption{
        \textbf{Aligning the Latent Space to Semantic Features Using The Programmable Prior On dSprites.} The plots show the DCI Disentanglement Score (left), the Latent Predictability Score (LPS) using LGBM (middle), and the reconstruction NMSE in decibels (right) as the number of pseudo-labels used for alignment increases. The DCI scores were always evaluated with respect to all 5 semantic features (Shape, Scale, Orientation, X-Position, Y-Position).
    }
    \label{fig:Alignment_Results}
\end{figure*}

\subsection{State-of-the-Art Reduction in Entanglement}
\label{sec:exp_quantitative}

For unsupervised disentanglement, our quantitative analysis reveals that our MMD-based regularizer achieves state-of-the-art performance in enforcing latent statistical independence, a claim supported across $8$ datasets with several model configurations. The condensed results in Table~\ref{tab:main_results_revised_v2} demonstrate that our method consistently yields the lowest Latent Predictability Scores (LPS) with a 70\% reduction compared to the leading baseline FactorVAE, signifying a lower degree of entanglement at the same prediction level, with little to no degradation in reconstruction quality. Interestingly, we note that the NashAE performs poorly on several baselines (ImageNet, XYCS, celebA) while its trained adversary responsible for detecting co-dependence in a very similar way to how we calculate the LPS reports almost no entanglement. We attribute this critical failure to optimization instability inherent to adversarial frameworks, where the inner-loop adversary fails to converge sufficiently to act as a reliable proxy for the true latent dependence.

In Appendix~\ref{app:MMD_Robustness}, we provide a comprehensive ablation study showing the stability, scalability, and robustness of our MMD-based sculpting regularizer. As shown in Figure~\ref{fig:LPS_vs_latent_dim}, while most baseline methods learn increasingly entangled representations as the latent space grows, our MMD regularizer maintains a consistently low LPS. Moreover, Figures \ref{fig:MMDAblation_regcoef_batchsize} and \ref{fig:MMDAblation_KernelTypes} highlight the stability and robustness of the regularization as there is only a very small sensitivity in the model's LPS score and reconstruction NMSE when changing the training batch size from $64$ to $2048$, the regularization coefficient from $0.1$ to $5$, the number of bandwidth in the median heuristic from $1$ to $5$, and even across some kernel types.

Lastly, Appendix~\ref{app:programmable_priors} demonstrates the impact of changing the aggregate posterior's target prior on the alignment to semantic features of our unsupervised disentanglement method. Our experiments on dsprites and XY(C/S) synthetic datasets demonstrate that the subsequent alignment of the representations can more than double with this simple change highlighting the fact that the choice of prior is, on its own, a powerful inductive bias for learning representations.

\subsection{Alignment to Interpretable Features with the Programmable Prior}
\label{sec:qualitative_analysis}
While achieving (1) mutual independence is the primary objective of unsupervised methods, the ultimate goal of disentanglement is to (2) align these independent features with interpretable, real-world concepts. Our framework addresses this by allowing us to inject inductive bias via Programmable Priors.

\begin{figure}[h!]
    \centering
    \begin{subfigure}[b]{0.49\textwidth}
        \begin{subfigure}[b]{0.45\textwidth}
            \centering
            \includegraphics[width=\linewidth]{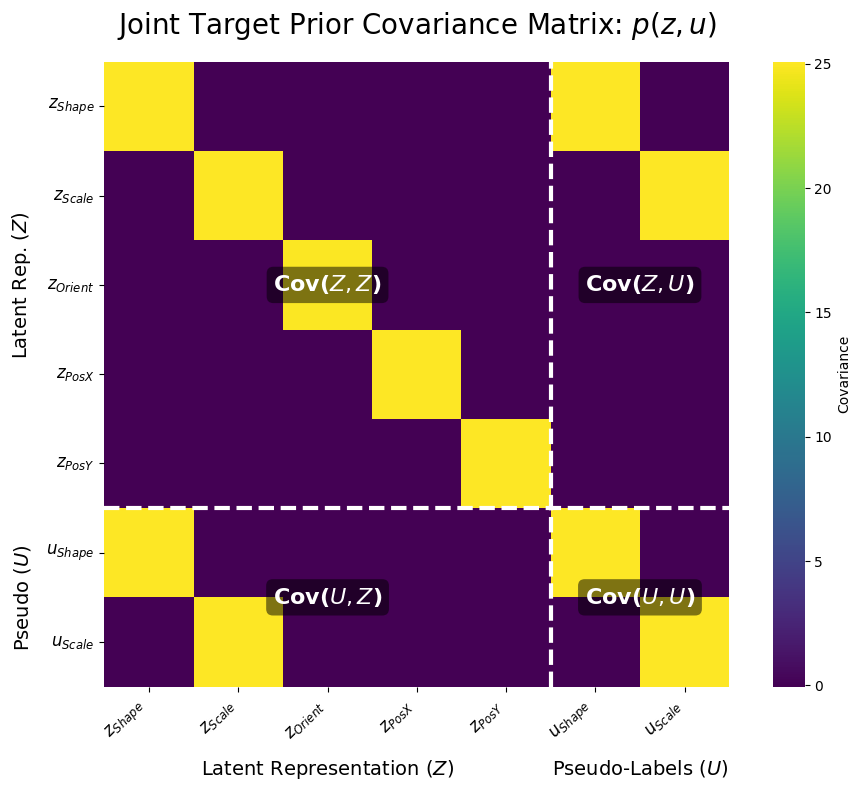}
        \end{subfigure}
        \begin{subfigure}[b]{0.45\textwidth}
            \centering
            \includegraphics[width=\linewidth]{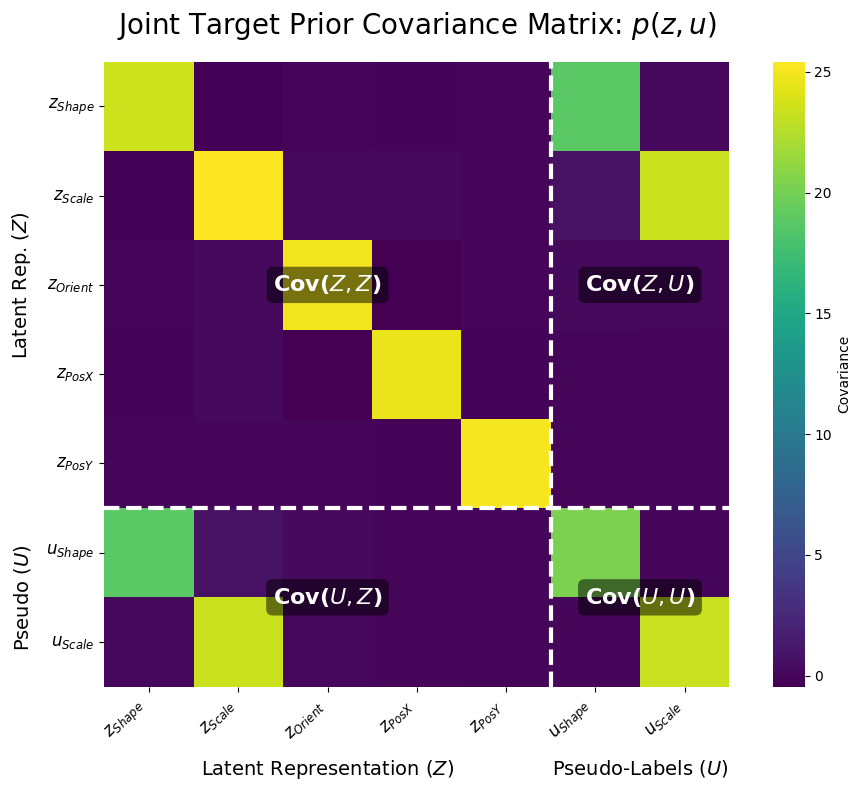}
        \end{subfigure}
        \caption{Aligning $2$ pseudo-labels}
    \end{subfigure}
    \hfill
    \begin{subfigure}[b]{0.49\textwidth}
        \begin{subfigure}[b]{0.45\textwidth}
            \centering
            \includegraphics[width=\linewidth]{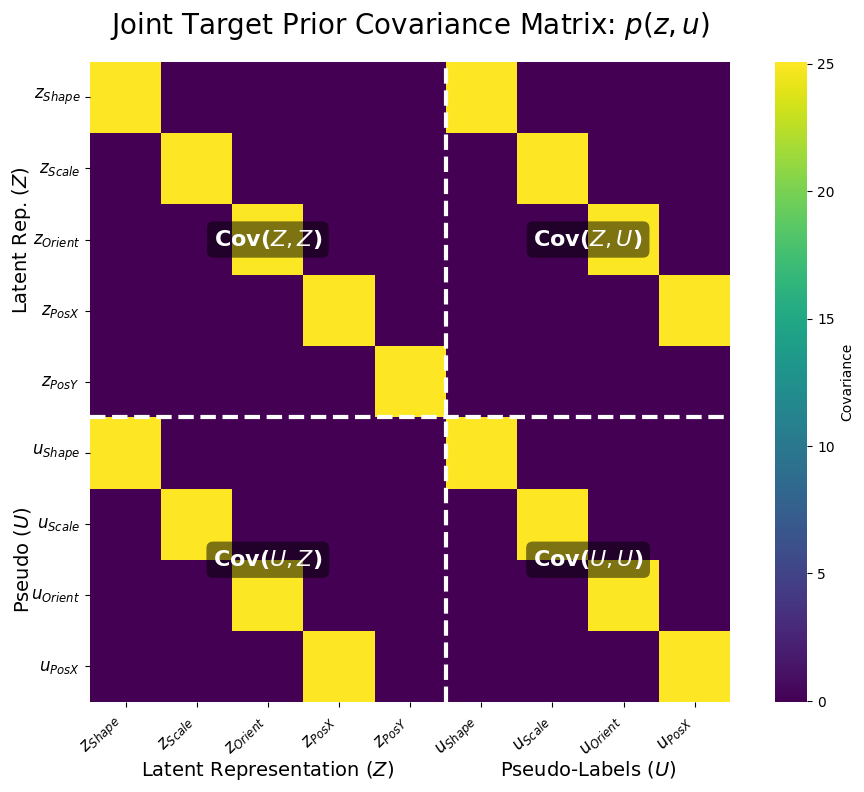}
        \end{subfigure}
        \begin{subfigure}[b]{0.45\textwidth}
            \centering
            \includegraphics[width=\linewidth]{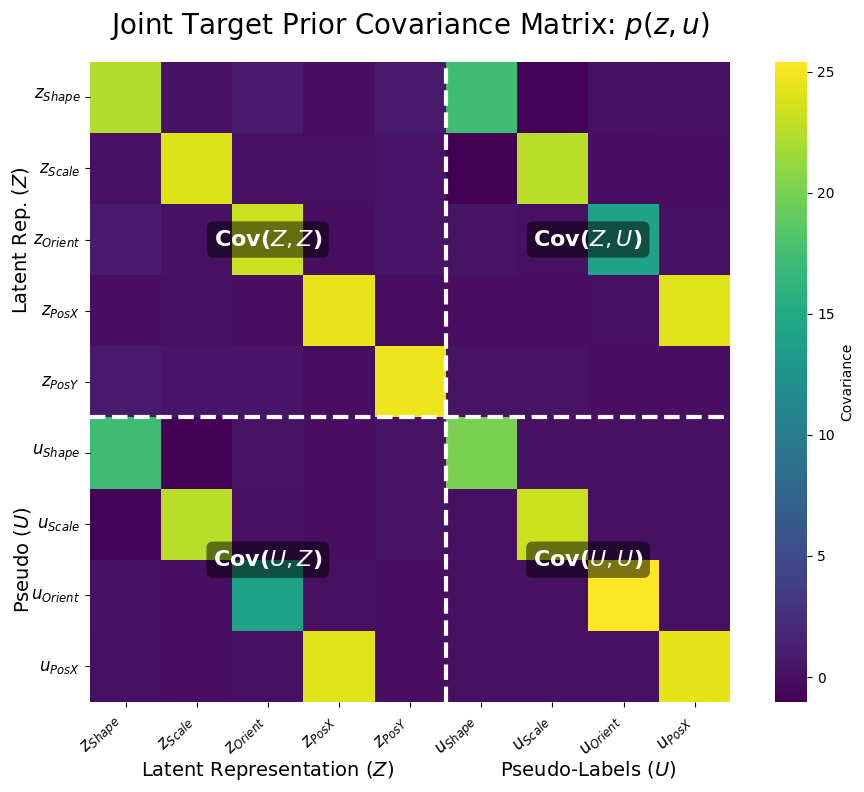}
        \end{subfigure}
        \caption{Aligning $4$ pseudo-labels}
    \end{subfigure}
    \caption{
        \textbf{Visualization of Alignment Mechanism in Programmable Prior Framework.} The figure shows the covariance matrix of  the joint distribution $p(z,f(u))$ for the target (left) and the learned joint distribution (right) with $2$ and $4$ pseudo-labels respectively on the dsprites dataset.
    }
    \label{fig:AlignmentPriorsCov}
\end{figure}

\begin{figure*}[h!]
    \centering
    \begin{subfigure}[b]{0.435\textwidth}
        \begin{subfigure}[b]{0.49\textwidth}
            \centering
            \includegraphics[width=\linewidth]{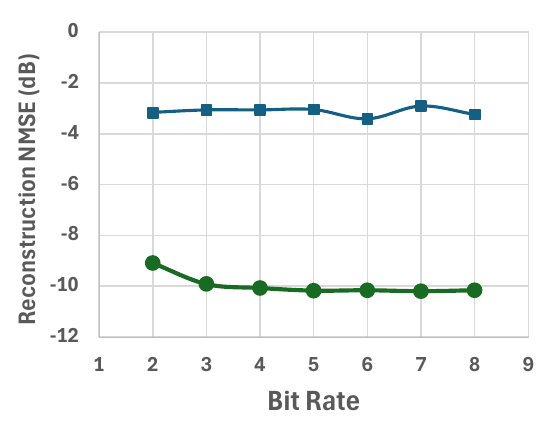}
        \end{subfigure}
        \begin{subfigure}[b]{0.49\textwidth}
            \centering
            \includegraphics[width=\linewidth]{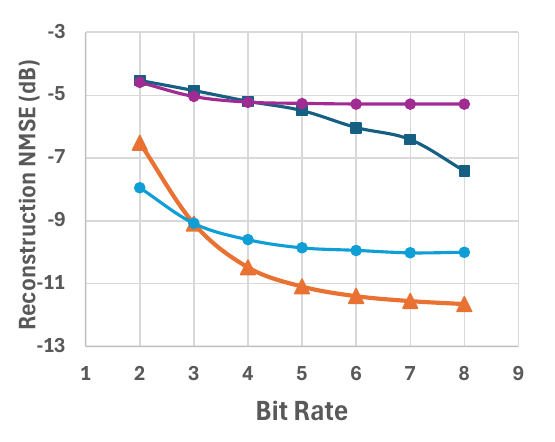}
        \end{subfigure}
        \caption{MNIST}
    \end{subfigure}
    \hfill
    \begin{subfigure}[b]{0.54\textwidth}
        \begin{subfigure}[b]{0.405\textwidth}
            \centering
            \includegraphics[width=\linewidth]{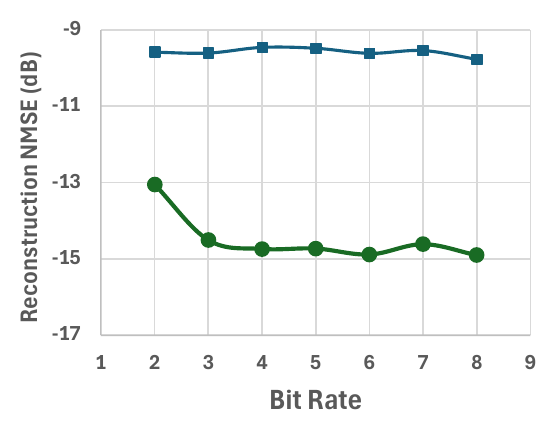}
        \end{subfigure}
        \begin{subfigure}[b]{0.58\textwidth}
            \centering
            \includegraphics[width=\linewidth]{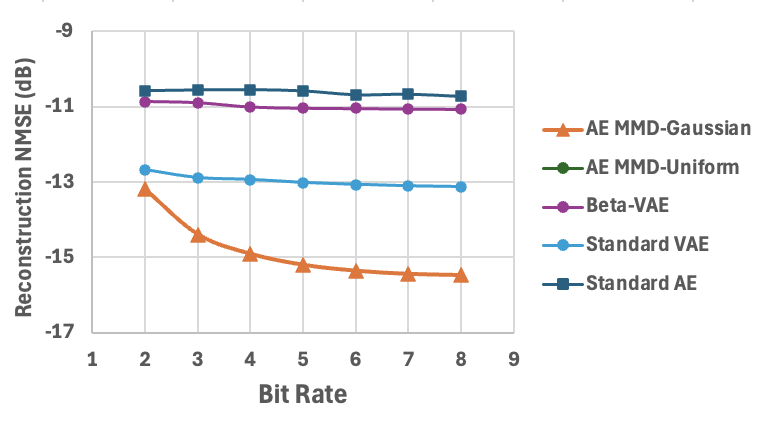}
        \end{subfigure}
        \caption{CIFAR10}
    \end{subfigure}
    \caption{
        \textbf{Compression with the Programmable Prior.} We compare the reconstruction NMSE (dB) of our MMD-regularized models against baselines with a uniform (left) or a Gaussian (right) quantizer.
    }
    \label{fig:QuantizationResults}
\end{figure*}

Formally, we construct a target prior where $z$ is sampled from a base distribution $p(z)$ (e.g., Gaussian or Uniform), and the semantic constraint is enforced by defining $f_\phi(u) = z_{0:k}$, where $k$ is the number of pseudo-labels to align. This joint pair $(z, f_\phi(u))$ acts both as a ``glue" that aligns our model's latents $z$ with semantic features $f_\phi(u)$, and as the target prior that ensures the latent space is both mutually independent and correctly shaped (see visualization in Fig~\ref{fig:AlignmentPriorsCov}).

We evaluate this mechanism on the dSprites dataset by progressively increasing the number of semantic factors (up to 5) for which we provide pseudo-labels. Figure~\ref{fig:Alignment_Results} demonstrates the effectiveness of this approach. The DCI score rises sharply as we align more factors, reaching a peak when $3$ out of $5$ features are aligned for the uniform prior, and 4 out of 5 for the Gaussian prior. In comparison, state-of-the-art baselines such as  Ada-GVAE and Ada-ML-VAE \cite{locatello2020weakly} report DCI scores on dsprites lower than $0.5$ without being able to additionally constrain the aggregate posterior distribution. This confirms that our framework can successfully force the model to adopt a representation where specific latent dimensions correspond one-to-one with semantic factors, effectively addressing the NICA unidentifiability problem. Interestingly, as the model aligns with more semantic features, the LPS score increases (indicating \textit{lower} statistical independence). This confirms a crucial insight from the literature: the ``ground truth" factors in classic benchmarks like dSprites are often not statistically independent (e.g., size and position constraints). By forcing the model to align with these entangled semantic concepts, we necessarily sacrifice pure statistical independence. This empirically validates our claim that (1) independence and (2) alignment are distinct and sometimes conflicting goals specifically on the benchmark datasets. Throughout this process, the reconstruction error remains low and stable. This indicates that the model is able to find a representation that satisfies the alignment constraints without compromising its ability to compress and reconstruct the data.

\subsection{Quantizer Compatibility With the Programmable Prior}
\label{sec:Quantization_main_body}

A critical advantage of our Programmable Prior Framework is its ability to sculpt the aggregate posterior into distributions inherently compatible with downstream tasks like compression. 
To demonstrate this, we trained our model with a quantization block in loop to quantize the encoder output. We utilize two commonly used quantizers: a uniform quantizer which considers a uniform distribution in the latent space ($\mathcal{U}[0, 1]$) and a non-uniform quantizer, which considers a Gaussian distribution ($N(0,1)$). With our MMD-based framework, we constrain the latent space to have a latent distribution matched to the quantizer distribution. The downstream compression performance is measured by quantizing the learned latent codes to various bit-rates ($2$ to $8$ bits per dimension) and computing the reconstruction NMSE in dB.

As shown in Figure~\ref{fig:QuantizationResults}, our MMD-based models (blue lines) consistently outperforms the baseline by a wide margin ($\approx 2-7$ dB). This confirms our framework allows to explicitly program the prior to match the quantizer's expected distribution (whether Gaussian or Uniform), achieving high-fidelity compression ``out of the box", without the need for complex post-training vector quantization or entropy coding schemes.

%% file: Sections/Conclusion.tex
\section{Conclusion}
\label{sec:conclusion}
In this work, we introduced a novel unsupervised metric to lower bound the entanglement of learned representations. We subsequently challenged the foundational assumptions of the VAE and Adversarial Training based paradigms for unsupervised disentanglement, demonstrating that the analytical KL-divergence and adversaries are flawed mechanisms for enforcing statistical independence, the first step towards a disentangled representation. As a solution, we introduced the Programmable Priors, a unified method built on the Maximum Mean Discrepancy that solves three distinct representation engineering goals: mutual independence, semantic alignment, and aggregate posterior distribution shaping.
Our experiments confirm that this approach achieves state-of-the-art statistical independence on complex datasets without sacrificing reconstruction quality. Crucially, we showed that while independence is necessary, it is not sufficient for interpretability. By leveraging our framework's ability to enforce a programmable joint distribution, we demonstrated a novel semi-supervised mechanism to solve the alignment problem, effectively bridging the gap between statistical disentanglement and human-interpretable features. Ultimately, this work establishes the prior not as a fixed assumption, but as a precise engineering tool for sculpting latent spaces.

%% file: Sections/Appendix.tex
\appendix
\section*{Appendix}

\subsubsection*{LLM Use Acknowledgment}
We acknowledge the use of Google's Gemini Pro 2.5 in the preparation of this manuscript. The model was utilized to aid in polishing grammar and clarity, assist with generating formatting for tables and figures, and serve as an assistant for the retrieval of prior work. The authors meticulously reviewed and edited all generated content and take full responsibility for the accuracy and originality of this work.

\vspace{1em} 

\newlength{\appendixtocindent}
\setlength{\appendixtocindent}{1.5em}

\subsection*{Table Of Contents}
\noindent 

\noindent\textbf{\ref{app:implementation_details} \quad Experiment And Implementation Details \dotfill \pageref{app:implementation_details}}
\vspace{0.5em} \\ 
\noindent\hspace*{\appendixtocindent} \ref{sec:exp_design} \quad Experimental Design \dotfill \pageref{sec:exp_design} \\
\noindent\hspace*{\appendixtocindent} \ref{app:Evaluation_Protocol_and_Metrics} \quad Evaluation Protocol and Metrics \dotfill \pageref{app:Evaluation_Protocol_and_Metrics} \\
\noindent\hspace*{\appendixtocindent} \ref{app:MMD_Regularizer_Implementation} \quad MMD Regularizer Implementation \dotfill \pageref{app:MMD_Regularizer_Implementation} \\

\vspace{1em} 

\noindent\textbf{\ref{app:Results} \quad Comprehensive  Quantitative Experimental Results \dotfill \pageref{app:Results}}
\vspace{1em} \\

\textbf{\ref{app:Metric Analysis} \quad LPS Metric Analysis \dotfill \pageref{app:Metric Analysis}}
\vspace{0.5em} \\ 
\noindent\hspace*{\appendixtocindent} \ref{app:paradox} \quad The Alignment-Independence Paradox \dotfill \pageref{app:paradox} \\
\noindent\hspace*{\appendixtocindent} \ref{app:instability} \quad Metric Instability Across Runs \dotfill \pageref{app:instability} \\
\noindent\hspace*{\appendixtocindent} \ref{app:lps_stability} \quad LPS as a Stable, Intrinsic Measure \dotfill \pageref{app:lps_stability}

\vspace{1em}
\noindent\textbf{\ref{app:MMD_Robustness} \quad Scalibility, Stability, and Robustness of The MMD-Based Sculpting Regularizer \dotfill \pageref{app:MMD_Robustness}}

\vspace{1em}
\noindent\textbf{\ref{app:programmable_priors} \quad The Programmable Priors As a Powerful Inductive Bias: Extended Results \dotfill \pageref{app:programmable_priors}}
\vspace{0.5em} \\
\noindent\hspace*{\appendixtocindent} \ref{app:dSprites_hybrid} \quad dSprites: Engineering a Mixed-Distribution Prior \dotfill \pageref{app:dSprites_hybrid} \\
\noindent\hspace*{\appendixtocindent} \ref{app:XY_hybrid} \quad XY: Engineering a Uniform Prior \dotfill \pageref{app:XY_hybrid} \\
\noindent\hspace*{\appendixtocindent} \ref{app:XYC_hybrid} \quad XYC: Engineering a GMM-Heavy Prior \dotfill \pageref{app:XYC_hybrid} \\
\noindent\hspace*{\appendixtocindent} \ref{app:XYCS_hybrid} \quad XYCS: Engineering a Mixed-Component Prior \dotfill \pageref{app:XYCS_hybrid}

\vspace{1em}
\noindent\textbf{\ref{app:Additional_Experiments} \quad Additional Experiments \dotfill \pageref{app:Additional_Experiments}}
\vspace{0.5em} \\
\noindent\hspace*{\appendixtocindent} \ref{app:KnowledgeTransfer} \quad The Programmabe Prior for Representational Knowledge Transfer \dotfill \pageref{app:KnowledgeTransfer} \\

\newpage

\section{Experiment and Implementation Details}
\label{app:implementation_details}

\subsection{Experimental Design}
\label{sec:exp_design}

\subsubsection*{Datasets}
Our method was evaluated on a diverse suite of benchmarks to assess its scalability and effectiveness.
\begin{itemize}
    \item \textbf{Synthetic Datasets:} For controlled experiments with known ground-truth factors, we used \textbf{dSprites} \citep{dsprites17} and the \textbf{XY}, \textbf{XYC}, and \textbf{XYCS} environments \citep{cha2023orthogonality}. These datasets feature simple shapes with explicit factors of variation, such as position (XY), color (C), and shape (S), and CelebA \citep{liu2015faceattributes}.
    \item \textbf{Real-World Datasets:} To test performance on complex, high-dimensional data, we used the standard image classification benchmarks \textbf{MNIST} \citep{deng2012mnist}, \textbf{CIFAR-10} \citep{krizhevsky2009learning}, and \textbf{Tiny ImageNet} \citep{le2015tiny}.
\end{itemize}

\subsubsection*{Baseline Models}
We benchmarked our MMD-based approach against established methods for unsupervised disentanglement, including \textbf{$\beta$-VAE} \citep{higgins2017beta}, \textbf{$\beta$-TCVAE} \citep{chen2018isolating}, \textbf{FactorVAE} \citep{kim2018disentangling}, \textbf{DGAE} \citep{cha2023orthogonality}, and \textbf{NashAE} \cite{yeats2022nashae}. We also included a standard \textbf{Autoencoder (AE)} and a standard \textbf{VAE} as controls to establish a performance baseline without explicit disentanglement pressures.

\subsubsection*{Training Details}
All models were trained using the Adam optimizer with a learning rate of $5 \times 10^{-4}$ and a batch size of 512. Our MMD-based regularizer used a regularization coefficient of $\lambda = 0.3$. We additionally found that our framework was not very sensitive to variation in this hyperparameter as values between $0.1$ and $2$ resulted in comparable performance. The models are fully convolutional, featuring symmetric encoder-decoder architectures. Specific depths and latent dimensions for each dataset are detailed in Table~\ref{tab:hyperparams}. Finally, for the alignment mechanism, we employ a lightweight element-wise MLP for each pseudo-label with a single hidden layer of $64$ units, structured as follows: $\text{Linear}(1, 64) \to \text{BatchNorm} \to \text{ReLU} \to \text{Linear}(64, 1)$.

\begin{table*}[tp]
\centering
\caption{Dataset-specific hyperparameters. The "Number of Layers" refers to the depth of the encoder and decoder individually (e.g., "6" indicates 6 layers for the encoder and 6 for the decoder).}
\label{tab:hyperparams}
\begin{tabular}{@{}lccc@{}}
\toprule
\textbf{Dataset} & \textbf{Number of Layers} & \textbf{Latent Dimension ($d$)} & \textbf{Training Epochs} \\
\midrule
XY, XYC, XYCS & 6, 6, 6 & 2, 3, 4 & 500, 250, 125 \\
dSprites      & 6       & 5         & 50              \\
CelebA      & 6       & 32           & 75              \\
MNIST         & 4       & 12        & 250             \\
CIFAR-10      & 4       & 64        & 250             \\
Tiny ImageNet & 6       & 64        & 75              \\
\bottomrule
\end{tabular}
\end{table*}

\subsection{Evaluation Protocol and Metrics}
\label{app:Evaluation_Protocol_and_Metrics}
To ensure robust and reproducible results, all reported scores are the mean and standard deviation over $5$,$10$, or $15$ independent runs with different random seeds. 

\subsection*{Metric Descriptions}
\noindent\textbf{NMSE (dB / linear).} The Normalized Mean Squared Error measures reconstruction quality, comparing the output of the decoder to the original input image. We report it on both a linear scale and a decibel (dB) scale, which is logarithmic. For the dB scale, more negative values indicate better reconstruction.

\noindent\textbf{LPS (LGBM / MLP).} Our proposed Latent Predictability Score measures the mutual independence of the learned latent features. It is based on the average coefficient of determination ($R^2$) from a regression task that attempts to predict each latent dimension from all others. A score near zero indicates a high degree of independence. Therefore, a \textbf{lower LPS score is better}. We report two variants using LightGBM and a Multi-Layer Perceptron ($2$ hidden layers of sizes $64$ and $32$ respectively) as the regressors. The regression models are trained on up to $100000$ latent samples for up to $200$ epochs for the MLP and $100$ iterations for LightGBM with a $80/20$ train/test split.

\noindent\textbf{Covariance Ratio.} This metric quantifies the degree of \textit{linear} feature independence by measuring how diagonal the latent covariance matrix is. It is calculated as the ratio of the sum of the diagonal elements to the sum of the absolute values of the off-diagonal elements. A \textbf{higher ratio indicates a more diagonal covariance matrix} and thus greater \textbf{linear} statistical independence.

\noindent\textbf{MIG (Mutual Information Gap).} This metric measures the degree to which each latent variable is informative about one, and only one, ground-truth factor. It is calculated as the un-normalized difference between the highest and second-highest mutual information over all factors \cite{chen2018isolating}. A \textbf{higher MIG score is better}.

\noindent\textbf{DCI (Disentanglement, Completeness, Informativeness) \citep{eastwood2018framework}.} This framework probes the latent space with a Random Forest regressor and reports three scores:
\begin{itemize}
    \item \textbf{Disentanglement:} Measures if each latent dimension captures at most one ground-truth factor. A higher score is better.
    \item \textbf{Completeness:} Measures if each ground-truth factor is captured by a single latent dimension. A higher score is better.
    \item \textbf{Informativeness:} Measures the overall accuracy in predicting ground-truth factors from the latent representation. A higher score is better.
\end{itemize}

\noindent\textbf{SAP (Separated Attribute Predictability).} This metric measures the extent to which each latent dimension is predictive of only a single ground-truth factor. For each factor, it computes the difference in prediction error between a model trained on a single latent dimension and a model trained on all other dimensions. A \textbf{higher SAP score is better}.

\subsection{MMD Regularizer Implementation}
\label{app:MMD_Regularizer_Implementation}
Our MMD regularizer is designed to be robust and adaptive. The following provides specific implementation details.

\textbf{Kernel Configuration}
We employ a mixture of Gaussian Radial Basis Function (RBF) kernels for all MMD calculations. The RBF kernel is defined as $k(z_i, z_j) = \exp(-\|z_i - z_j\|^2 / (2\sigma^2))$, where $\sigma$ is the bandwidth. Using a sum of kernels with different bandwidths (generated from a base bandwidth with multipliers $[0.5, 1.0, 2.0]$) allows the MMD to capture distributional discrepancies across multiple scales, leading to a more robust distance metric.

\textbf{Adaptive Bandwidth Selection}
Instead of fixing the kernel bandwidth $\sigma$, we use the \textbf{median heuristic} \citep{gretton2012kernel}. At each training step, the base bandwidth $\sigma_{\text{base}}$ is set to the median of all pairwise distances between the points in the combined set of latent samples $z \sim q_{\theta_1}(z)$ and prior samples $z' \sim p(z)$. This adaptive approach automatically scales the kernel to the current batch, eliminating the need for manual hyperparameter tuning.

\textbf{MMD Estimator}
At each training step, we draw a batch of samples from the target prior $p(z)$ of the same size as the batch of encoded latent samples from the aggregate posterior $q_{\theta_1}(z)$. The MMD loss is then computed using the standard \textbf{biased, quadratic-time estimator} of the squared MMD. Despite its $O(n^2)$ complexity, this estimator is computationally efficient for typical batch sizes and is known to be effective and robust in practice.

\newpage
\section{Comprehensive Experimental Results}
\label{app:Results}

This appendix provides the complete, unabridged quantitative results of our experiments for all models, and datasets. All reported scores are the mean and standard deviation computed over 5 independent runs with different random seeds.

\begin{table*}[tp]
\caption{Summary for TINY IMAGENET}
\label{tab:tiny_imagenet}
\centering
\resizebox{\textwidth}{!}{
\begin{tabular}{@{}l*{5}{l}@{}}
\toprule
 & \multicolumn{5}{c}{L=64} \\
\cmidrule(lr){2-6}
Baseline & Covariance Ratio & LPS\_LGBM & LPS\_MLP & NMSE\_dB & NMSE\_linear \\
\midrule
AE-MMD-Gaussian (Ours) & \textbf{54.386  $\pm$  2.220} & \textbf{0.029  $\pm$  0.002} & \textbf{0.001  $\pm$  0.004} & \textbf{-11.704  $\pm$  0.007} & \textbf{0.068  $\pm$  0.000} \\
FactorVAE ($\gamma$=10.0) & 2.338  $\pm$  0.286 & 0.437  $\pm$  0.064 & 0.483  $\pm$  0.065 & -10.754  $\pm$  0.239 & 0.084  $\pm$  0.005 \\
$\beta$-TCVAE ($\beta$=4.0) & 4.950  $\pm$  0.264 & 0.794  $\pm$  0.018 & 0.932  $\pm$  0.015 & -9.798  $\pm$  0.027 & 0.105  $\pm$  0.001 \\
$\beta$-VAE ($\beta$=4) & 4.934  $\pm$  0.104 & 0.726  $\pm$  0.018 & 0.872  $\pm$  0.035 & -9.866  $\pm$  0.009 & 0.103  $\pm$  0.000 \\
\midrule
Standard AE & 7.699  $\pm$  0.207 & 0.455  $\pm$  0.006 & 0.460  $\pm$  0.007 & -11.775  $\pm$  0.005 & 0.066  $\pm$  0.000 \\
Standard VAE & 2.409  $\pm$  0.224 & 0.470  $\pm$  0.073 & 0.518  $\pm$  0.066 & -10.608  $\pm$  0.361 & 0.087  $\pm$  0.008 \\
\bottomrule
\end{tabular}
}
\end{table*}

\begin{table*}[tp]
\caption{Summary for CIFAR10}
\label{tab:cifar10}
\centering
\resizebox{\textwidth}{!}{
\begin{tabular}{@{}l*{5}{l}@{}}
\toprule
 & \multicolumn{5}{c}{L=64} \\
\cmidrule(lr){2-6}
Baseline & Covariance Ratio & LPS\_LGBM & LPS\_MLP & NMSE\_dB & NMSE\_linear \\
\midrule
AE-MMD-Gaussian (Ours) & \textbf{65.917  $\pm$  1.788} & \textbf{0.038  $\pm$  0.004} & \textbf{0.025  $\pm$  0.008} & \textbf{-15.615  $\pm$  0.008} & \textbf{0.027  $\pm$  0.000} \\
FactorVAE ($\gamma$=10.0) & 3.347  $\pm$  0.075 & 0.628  $\pm$  0.004 & 0.788  $\pm$  0.006 & -12.491  $\pm$  0.012 & 0.056  $\pm$  0.000 \\
$\beta$-TCVAE ($\beta$=4.0) & 10.711  $\pm$  0.350 & 0.890  $\pm$  0.008 & 0.953  $\pm$  0.006 & -10.684  $\pm$  0.005 & 0.085  $\pm$  0.000 \\
$\beta$-VAE ($\beta$=4) & 11.506  $\pm$  0.238 & 0.846  $\pm$  0.008 & 0.925  $\pm$  0.006 & -10.492  $\pm$  0.006 & 0.089  $\pm$  0.000 \\
\midrule
Standard AE & 6.929  $\pm$  0.192 & 0.508  $\pm$  0.005 & 0.513  $\pm$  0.005 & -15.834  $\pm$  0.007 & 0.026  $\pm$  0.000 \\
Standard VAE & 3.405  $\pm$  0.113 & 0.632  $\pm$  0.009 & 0.793  $\pm$  0.010 & -12.503  $\pm$  0.009 & 0.056  $\pm$  0.000 \\
\bottomrule
\end{tabular}
}
\end{table*}

\begin{table*}[tp]
\caption{Summary for MNIST}
\label{tab:mnist}
\centering
\resizebox{\textwidth}{!}{
\begin{tabular}{@{}l*{5}{l}@{}}
\toprule
 & \multicolumn{5}{c}{L=12} \\
\cmidrule(lr){2-6}
Baseline & Covariance Ratio & LPS\_LGBM & LPS\_MLP & NMSE\_dB & NMSE\_linear\\
\midrule
AE-MMD-Gaussian (Ours) & 46.635  $\pm$  2.442 & 0.220  $\pm$  0.007 & 0.333  $\pm$  0.006 &  \textbf{-11.922  $\pm$  0.024} & \textbf{0.064  $\pm$  0.000}  \\
DGAE & 6.191  $\pm$  0.368 & 0.632  $\pm$  0.002 & 0.671  $\pm$  0.005 & -11.478  $\pm$  0.027 & 0.071  $\pm$  0.000 \\
NASHAE & \textbf{47.835  $\pm$  6.503} & \textbf{0.210}  $\pm$  0.014 & 0.407  $\pm$  0.023 & -11.389  $\pm$  0.133 & 0.073 $\pm$  0.002 \\
FactorVAE ($\gamma$=10.0) & 2.628  $\pm$  0.094 & 0.271  $\pm$  0.066 & \textbf{0.275  $\pm$  0.098} & -8.621  $\pm$  0.023 & 0.137  $\pm$  0.001 \\
$\beta$-TCVAE ($\beta$=4.0) & 8.071  $\pm$  2.238 & 0.838  $\pm$  0.022 & 0.893  $\pm$  0.015 & -5.967  $\pm$  0.008 & 0.253  $\pm$  0.000 \\
$\beta$-VAE ($\beta$=4) & 9.493  $\pm$  1.000 & 0.898  $\pm$  0.009 & 0.938  $\pm$  0.007 & -4.606  $\pm$  0.020 & 0.346  $\pm$  0.002 \\
\midrule
Standard AE & 8.353  $\pm$  0.486 & 0.539  $\pm$  0.007 & 0.587  $\pm$  0.008 & -12.089  $\pm$  0.020 & 0.062  $\pm$  0.000 \\
Standard VAE & 2.656  $\pm$  0.267 & 0.249  $\pm$  0.053 & 0.242  $\pm$  0.070 &  -8.630  $\pm$  0.011 & 0.137  $\pm$  0.000 \\
\bottomrule
\end{tabular}
}
\end{table*}

\begin{table*}[tp]
\caption{Summary for CELEBA (L=32)}
\label{tab:celeba}
\centering
\resizebox{\textwidth}{!}{
\begin{tabular}{@{}l*{10}{l}@{}}
\toprule
 & \multicolumn{10}{c}{L=32} \\
\cmidrule(lr){2-11}
Baseline & NMSE\_dB & NMSE\_linear & completeness & disentanglement & feature\_dependence & informativeness & lps\_lgbm & lps\_mlp & mig & sap \\
\midrule
AE-MMD-Gaussian & -14.370  $\pm$  0.012 & 0.037  $\pm$  0.000 & 0.009  $\pm$  0.000 & 0.008  $\pm$  0.000 & 61.678  $\pm$  3.671 & 0.868  $\pm$  0.001 & 0.080  $\pm$  0.003 & 0.227  $\pm$  0.009 & 0.006  $\pm$  0.002 & 0.000  $\pm$  0.000 \\
DGAE & -14.404  $\pm$  0.016 & 0.036  $\pm$  0.000 & 0.014  $\pm$  0.001 & 0.013  $\pm$  0.001 & 5.742  $\pm$  0.318 & 0.873  $\pm$  0.001 & 0.481  $\pm$  0.011 & 0.529  $\pm$  0.010 & 0.011  $\pm$  0.003 & 0.000  $\pm$  0.000 \\
FactorVAE ($\gamma$=10.0) & -14.394  $\pm$  0.150 & 0.036  $\pm$  0.001 & 0.014  $\pm$  0.001 & 0.013  $\pm$  0.001 & 62.873  $\pm$  2.077 & 0.871  $\pm$  0.001 & 0.071  $\pm$  0.007 & 0.106  $\pm$  0.018 & 0.014  $\pm$  0.003 & 0.000  $\pm$  0.000 \\
NASHAE & -13.672  $\pm$  0.340 & 0.043  $\pm$  0.003 & 0.010  $\pm$  0.001 & 0.009  $\pm$  0.001 & 18.684  $\pm$  14.394 & 0.867  $\pm$  0.001 & 0.362  $\pm$  0.185 & 0.477  $\pm$  0.181 & 0.007  $\pm$  0.001 & 0.000  $\pm$  0.000 \\
Standard AE & -14.546  $\pm$  0.012 & 0.035  $\pm$  0.000 & 0.012  $\pm$  0.001 & 0.012  $\pm$  0.001 & 7.490  $\pm$  0.169 & 0.871  $\pm$  0.000 & 0.441  $\pm$  0.004 & 0.479  $\pm$  0.005 & 0.008  $\pm$  0.001 & 0.000  $\pm$  0.000 \\
VAE & -14.426  $\pm$  0.153 & 0.036  $\pm$  0.001 & 0.015  $\pm$  0.002 & 0.014  $\pm$  0.001 & 64.075  $\pm$  3.956 & 0.871  $\pm$  0.001 & 0.073  $\pm$  0.018 & 0.100  $\pm$  0.027 & 0.017  $\pm$  0.004 & 0.000  $\pm$  0.000 \\
$\beta$-TCVAE ($\beta$=4.0) & -13.097  $\pm$  0.143 & 0.049  $\pm$  0.002 & 0.013  $\pm$  0.001 & 0.012  $\pm$  0.001 & 62.598  $\pm$  6.288 & 0.869  $\pm$  0.000 & 0.594  $\pm$  0.015 & 0.807  $\pm$  0.020 & 0.015  $\pm$  0.003 & 0.000  $\pm$  0.000 \\
$\beta$-VAE ($\beta$=4) & -13.299  $\pm$  0.019 & 0.047  $\pm$  0.000 & 0.013  $\pm$  0.001 & 0.012  $\pm$  0.001 & 74.026  $\pm$  2.330 & 0.870  $\pm$  0.000 & 0.528  $\pm$  0.021 & 0.747  $\pm$  0.030 & 0.013  $\pm$  0.002 & 0.000  $\pm$  0.000 \\
\bottomrule
\end{tabular}
}
\end{table*}

\begin{table*}[tp]
\caption{Summary for CELEBA (L=64)}
\label{tab:celeba}
\centering
\resizebox{\textwidth}{!}{
\begin{tabular}{@{}l*{10}{l}@{}}
\toprule
 & \multicolumn{10}{c}{L=64} \\
\cmidrule(lr){2-11}
Baseline & NMSE\_dB & NMSE\_linear & completeness & disentanglement & feature\_dependence & informativeness & lps\_lgbm & lps\_mlp & mig & sap \\
\midrule
AE-MMD-Gaussian & -16.087  $\pm$  0.011 & 0.025  $\pm$  0.000 & 0.011  $\pm$  0.001 & 0.012  $\pm$  0.001 & 63.822  $\pm$  2.724 & 0.866  $\pm$  0.001 & 0.097  $\pm$  0.003 & 0.309  $\pm$  0.007 & 0.004  $\pm$  0.001 & 0.000  $\pm$  0.000 \\
DGAE & -15.894  $\pm$  0.030 & 0.026  $\pm$  0.000 & 0.019  $\pm$  0.002 & 0.021  $\pm$  0.003 & 7.992  $\pm$  0.522 & 0.874  $\pm$  0.001 & 0.498  $\pm$  0.008 & 0.839  $\pm$  0.079 & 0.012  $\pm$  0.005 & 0.000  $\pm$  0.000 \\
FactorVAE ($\gamma$=10.0) & -14.885  $\pm$  0.088 & 0.032  $\pm$  0.001 & 0.018  $\pm$  0.002 & 0.020  $\pm$  0.002 & 89.942  $\pm$  7.703 & 0.871  $\pm$  0.001 & 0.566  $\pm$  0.007 & 0.829  $\pm$  0.016 & 0.013  $\pm$  0.002 & 0.000  $\pm$  0.000 \\
NASHAE & -13.557  $\pm$  1.355 & 0.046  $\pm$  0.014 & 0.011  $\pm$  0.001 & 0.013  $\pm$  0.001 & 8.334  $\pm$  3.465 & 0.866  $\pm$  0.000 & 0.759  $\pm$  0.185 & 0.930  $\pm$  0.067 & 0.008  $\pm$  0.003 & 0.000  $\pm$  0.000 \\
Standard AE & -16.251  $\pm$  0.016 & 0.024  $\pm$  0.000 & 0.015  $\pm$  0.001 & 0.017  $\pm$  0.001 & 8.984  $\pm$  0.336 & 0.870  $\pm$  0.000 & 0.516  $\pm$  0.007 & 0.566  $\pm$  0.007 & 0.007  $\pm$  0.001 & 0.000  $\pm$  0.000 \\
VAE & -14.934  $\pm$  0.027 & 0.032  $\pm$  0.000 & 0.020  $\pm$  0.002 & 0.022  $\pm$  0.003 & 89.618  $\pm$  3.090 & 0.871  $\pm$  0.001 & 0.564  $\pm$  0.010 & 0.826  $\pm$  0.007 & 0.016  $\pm$  0.005 & 0.000  $\pm$  0.000 \\
$\beta$-TCVAE ($\beta$=4.0) & -13.225  $\pm$  0.014 & 0.048  $\pm$  0.000 & 0.016  $\pm$  0.001 & 0.018  $\pm$  0.001 & 73.189  $\pm$  8.060 & 0.869  $\pm$  0.000 & 0.833  $\pm$  0.012 & 0.970  $\pm$  0.006 & 0.014  $\pm$  0.003 & 0.000  $\pm$  0.000 \\
$\beta$-VAE ($\beta$=4) & -13.275  $\pm$  0.021 & 0.047  $\pm$  0.000 & 0.015  $\pm$  0.001 & 0.016  $\pm$  0.001 & 98.531  $\pm$  9.870 & 0.869  $\pm$  0.001 & 0.815  $\pm$  0.012 & 0.959  $\pm$  0.007 & 0.011  $\pm$  0.001 & 0.000  $\pm$  0.000 \\
\bottomrule
\end{tabular}
}
\end{table*}

\begin{table*}[tp]
\caption{Summary for DSPRITES}
\label{tab:dsprites}
\centering
\resizebox{\textwidth}{!}{
\begin{tabular}{@{}l*{10}{l}@{}}
\toprule
 & \multicolumn{10}{c}{L=5} \\
\cmidrule(lr){2-11}
Baseline & Covariance Ratio & LPS\_LGBM & LPS\_MLP & MIG\_mig\_score & NMSE\_dB & NMSE\_linear & SAP\_sap\_score & completeness & disentanglement & informativeness \\
\midrule
AE-MMD-Gaussian (Ours) & 102.265  $\pm$  21.202 & 0.039  $\pm$  0.002 & 0.033  $\pm$  0.006 & 0.058  $\pm$  0.024 & -17.888  $\pm$  0.281 & 0.016  $\pm$  0.001 & 0.000  $\pm$  0.000 & 0.026  $\pm$  0.018 & 0.025  $\pm$  0.017 & 0.950  $\pm$  0.002 \\
AE-MMD-hybrid-sampler (Ours) & 263.218  $\pm$  87.577 & 0.105  $\pm$  0.008 & 0.087  $\pm$  0.012 & 0.247  $\pm$  0.030 & -16.311  $\pm$  0.261 & 0.023  $\pm$  0.001 & 0.246  $\pm$  0.027 & 0.200  $\pm$  0.038 & 0.196  $\pm$  0.038 & 0.957  $\pm$  0.001 \\
DGAE & 4.065  $\pm$  1.233 & 0.557  $\pm$  0.112 & 0.573  $\pm$  0.147 & 0.261  $\pm$  0.198 & -14.614  $\pm$  2.173 & 0.040  $\pm$  0.026 & 0.000  $\pm$  0.000 & 0.226  $\pm$  0.104 & 0.219  $\pm$  0.101 & 0.962  $\pm$  0.004 \\
FactorVAE ($\gamma$=10.0) & 8.477  $\pm$  1.004 & 0.082  $\pm$  0.010 & 0.053  $\pm$  0.017 & 0.226  $\pm$  0.079 & -14.564  $\pm$  0.110 & 0.035  $\pm$  0.001 & 0.000  $\pm$  0.000 & 0.081  $\pm$  0.033 & 0.079  $\pm$  0.032 & 0.964  $\pm$  0.004 \\
$\beta$-TCVAE ($\beta$=4.0) & 6.469  $\pm$  4.162 & 0.223  $\pm$  0.165 & 0.180  $\pm$  0.182 & 0.408  $\pm$  0.215 & -8.831  $\pm$  0.085 & 0.131  $\pm$  0.003 & 0.000  $\pm$  0.000 & 0.212  $\pm$  0.119 & 0.195  $\pm$  0.105 & 0.960  $\pm$  0.005 \\
$\beta$-VAE ($\beta$=4) & 9.614  $\pm$  5.771 & 0.525  $\pm$  0.182 & 0.523  $\pm$  0.214 & 0.529  $\pm$  0.232 & -8.200  $\pm$  0.057 & 0.151  $\pm$  0.002 & 0.000  $\pm$  0.000 & 0.236  $\pm$  0.114 & 0.217  $\pm$  0.100 & 0.956  $\pm$  0.005 \\
\midrule
Standard AE & 12.669  $\pm$  2.560 & 0.187  $\pm$  0.039 & 0.181  $\pm$  0.042 & 0.078  $\pm$  0.032 & -18.739  $\pm$  0.290 & 0.013  $\pm$  0.001 & 0.082  $\pm$  0.058 & 0.041  $\pm$  0.018 & 0.041  $\pm$  0.018 & 0.957  $\pm$  0.001 \\
Standard VAE & 5.528  $\pm$  1.291 & 0.100  $\pm$  0.016 & 0.047  $\pm$  0.023 & 0.436  $\pm$  0.365 & -14.661  $\pm$  0.225 & 0.034  $\pm$  0.002 & 0.000  $\pm$  0.000 & 0.231  $\pm$  0.194 & 0.225  $\pm$  0.189 & 0.969  $\pm$  0.011 \\
\bottomrule
\end{tabular}
}
\end{table*}

\begin{table*}[tp]
\caption{Summary for XY}
\label{tab:xy}
\centering
\resizebox{\textwidth}{!}{
\begin{tabular}{@{}l*{10}{l}@{}}
\toprule
 & \multicolumn{10}{c}{L=2} \\
\cmidrule(lr){2-11}
Baseline & Covariance Ratio & LPS\_LGBM & LPS\_MLP & MIG\_mig\_score & NMSE\_dB & NMSE\_linear & SAP\_sap\_score & completeness & disentanglement & informativeness \\
\midrule
AE-MMD-Gaussian (Ours) & 240.970  $\pm$  243.891 & -0.053  $\pm$  0.007 & -0.004  $\pm$  0.003 & 0.751  $\pm$  0.554 & -15.222  $\pm$  0.403 & 0.030  $\pm$  0.003 & 0.423  $\pm$  0.316 & 0.199  $\pm$  0.244 & 0.199  $\pm$  0.243 & 1.000  $\pm$  0.000 \\
AE-MMD-hybrid-sampler (Ours) & 380.827  $\pm$  408.352 & -0.065  $\pm$  0.020 & -0.003  $\pm$  0.005 & 1.047  $\pm$  0.414 & -15.825  $\pm$  0.154 & 0.026  $\pm$  0.001 & 0.713  $\pm$  0.139 & 0.471  $\pm$  0.230 & 0.469  $\pm$  0.233 & 1.000  $\pm$  0.000 \\
DGAE & 16.631  $\pm$  19.417 & 0.248  $\pm$  0.380 & 0.241  $\pm$  0.385 & 1.580  $\pm$  1.414 & -6.742  $\pm$  2.522 & 0.262  $\pm$  0.206 & 0.000  $\pm$  0.000 & 0.521  $\pm$  0.435 & 0.521  $\pm$  0.435 & 0.999  $\pm$  0.001 \\
NASHAE & 52.672 $\pm$ 22.762 & -0.066 $\pm$ 0.012 & 0.007 $\pm$ 0.004 & 0.124 $\pm$ 0.059 & -13.082 $\pm$ 0.849 & 0.050 $\pm$ 0.010 & 0.325 $\pm$ 0.186 & 0.134 $\pm$ 0.144 & 0.131 $\pm$ 0.142 & 0.999 $\pm$ 0.001 \\
FactorVAE ($\gamma$=10.0) & 2.022  $\pm$  1.392 & -0.040  $\pm$  0.011 & 0.015  $\pm$  0.013 & 0.693  $\pm$  0.545 & -15.364  $\pm$  0.359 & 0.029  $\pm$  0.002 & 0.537  $\pm$  0.268 & 0.337  $\pm$  0.255 & 0.336  $\pm$  0.255 & 1.000  $\pm$  0.000 \\
$\beta$-TCVAE ($\beta$=4.0) & 1.597  $\pm$  0.551 & -0.042  $\pm$  0.015 & 0.009  $\pm$  0.006 & 1.149  $\pm$  1.093 & -14.684  $\pm$  0.278 & 0.034  $\pm$  0.002 & 0.502  $\pm$  0.320 & 0.372  $\pm$  0.345 & 0.367  $\pm$  0.344 & 1.000  $\pm$  0.000 \\
$\beta$-VAE ($\beta$=4) & 2.424  $\pm$  1.371 & -0.053  $\pm$  0.013 & 0.016  $\pm$  0.015 & 0.688  $\pm$  0.643 & -14.457  $\pm$  0.137 & 0.036  $\pm$  0.001 & 0.522  $\pm$  0.323 & 0.358  $\pm$  0.339 & 0.358  $\pm$  0.339 & 1.000  $\pm$  0.000 \\
\midrule
Standard AE & 503.878  $\pm$  536.454 & -0.028  $\pm$  0.037 & 0.024  $\pm$  0.014 & 0.964  $\pm$  0.837 & -16.904  $\pm$  0.081 & 0.020  $\pm$  0.000 & 0.470  $\pm$  0.348 & 0.362  $\pm$  0.369 & 0.362  $\pm$  0.369 & 1.000  $\pm$  0.000 \\
Standard VAE & 1.284  $\pm$  0.134 & -0.055  $\pm$  0.011 & 0.015  $\pm$  0.007 & 0.195  $\pm$  0.098 & -15.720  $\pm$  0.273 & 0.027  $\pm$  0.002 & 0.309  $\pm$  0.057 & 0.103  $\pm$  0.030 & 0.102  $\pm$  0.030 & 1.000  $\pm$  0.000 \\
\bottomrule
\end{tabular}
}
\end{table*}

\begin{table*}[tp]
\caption{Summary for XYC}
\label{tab:xyc}
\centering
\resizebox{\textwidth}{!}{
\begin{tabular}{@{}l*{10}{l}@{}}
\toprule
 & \multicolumn{10}{c}{L=3} \\
\cmidrule(lr){2-11}
Baseline & Covariance Ratio & LPS\_LGBM & LPS\_MLP & MIG\_mig\_score & NMSE\_dB & NMSE\_linear & SAP\_sap\_score & completeness & disentanglement & informativeness \\
\midrule
AE-MMD-Gaussian (Ours) & 49.341  $\pm$  12.168 & -0.014  $\pm$  0.016 & 0.016  $\pm$  0.007 & 0.256  $\pm$  0.118 & -16.311  $\pm$  0.245 & 0.023  $\pm$  0.001 & 0.237  $\pm$  0.077 & 0.115  $\pm$  0.039 & 0.113  $\pm$  0.038 & 0.999  $\pm$  0.000 \\
AE-MMD-hybrid-sampler (Ours) & 324.644  $\pm$  274.142 & 0.168  $\pm$  0.030 & 0.117  $\pm$  0.035 & 0.465  $\pm$  0.336 & -15.962  $\pm$  0.205 & 0.025  $\pm$  0.001 & 0.266  $\pm$  0.240 & 0.228  $\pm$  0.224 & 0.227  $\pm$  0.224 & 1.000  $\pm$  0.000 \\
DGAE & 4.802  $\pm$  2.926 & 0.498  $\pm$  0.417 & 0.503  $\pm$  0.414 & 0.340  $\pm$  0.181 & -10.936  $\pm$  3.421 & 0.110  $\pm$  0.083 & 0.000  $\pm$  0.000 & 0.184  $\pm$  0.105 & 0.175  $\pm$  0.098 & 0.996  $\pm$  0.005 \\
NASHAE & 16.476 $\pm$ 5.414 & 0.145 $\pm$ 0.091 & 0.138 $\pm$ 0.084 & 0.048 $\pm$ 0.004 & -14.933 $\pm$ 0.701 & 0.033 $\pm$ 0.005 & 0.217 $\pm$ 0.080 & 0.095 $\pm$ 0.035 & 0.094 $\pm$ 0.034 & 0.999 $\pm$ 0.000 \\
FactorVAE ($\gamma$=10.0) & 1.581  $\pm$  0.076 & -0.000  $\pm$  0.006 & 0.032  $\pm$  0.005 & 0.140  $\pm$  0.059 & -15.919  $\pm$  0.219 & 0.026  $\pm$  0.001 & 0.183  $\pm$  0.157 & 0.079  $\pm$  0.069 & 0.076  $\pm$  0.065 & 1.000  $\pm$  0.000 \\
$\beta$-TCVAE ($\beta$=4.0) & 2.900  $\pm$  0.258 & -0.010  $\pm$  0.009 & 0.023  $\pm$  0.007 & 0.188  $\pm$  0.105 & -13.720  $\pm$  0.129 & 0.042  $\pm$  0.001 & 0.198  $\pm$  0.120 & 0.093  $\pm$  0.078 & 0.090  $\pm$  0.075 & 1.000  $\pm$  0.000 \\
$\beta$-VAE ($\beta$=4) & 1.861  $\pm$  0.119 & 0.003  $\pm$  0.010 & 0.027  $\pm$  0.009 & 0.266  $\pm$  0.149 & -13.333  $\pm$  0.089 & 0.046  $\pm$  0.001 & 0.278  $\pm$  0.123 & 0.143  $\pm$  0.107 & 0.138  $\pm$  0.100 & 1.000  $\pm$  0.000 \\
\midrule
Standard AE & 9.459  $\pm$  4.389 & 0.042  $\pm$  0.019 & 0.072  $\pm$  0.019 & 0.200  $\pm$  0.082 & -17.811  $\pm$  0.134 & 0.017  $\pm$  0.001 & 0.227  $\pm$  0.090 & 0.087  $\pm$  0.052 & 0.085  $\pm$  0.049 & 1.000  $\pm$  0.000 \\
Standard VAE & 1.606  $\pm$  0.125 & 0.016  $\pm$  0.012 & 0.048  $\pm$  0.008 & 0.176  $\pm$  0.082 & -16.043  $\pm$  0.147 & 0.025  $\pm$  0.001 & 0.197  $\pm$  0.056 & 0.086  $\pm$  0.043 & 0.085  $\pm$  0.043 & 0.999  $\pm$  0.000 \\
\bottomrule
\end{tabular}
}
\end{table*}

\begin{table*}[tp]
\caption{Summary for XYCS}
\label{tab:xycs}
\centering
\resizebox{\textwidth}{!}{
\begin{tabular}{@{}l*{10}{l}@{}}
\toprule
 & \multicolumn{10}{c}{L=4} \\
\cmidrule(lr){2-11}
Baseline & Covariance Ratio & LPS\_LGBM & LPS\_MLP & MIG\_mig\_score & NMSE\_dB & NMSE\_linear & SAP\_sap\_score & completeness & disentanglement & informativeness \\
\midrule
AE-MMD-Gaussian (Ours) & 42.117  $\pm$  2.445 & 0.111  $\pm$  0.011 & 0.106  $\pm$  0.020 & 0.092  $\pm$  0.036 & -16.757  $\pm$  0.143 & 0.021  $\pm$  0.001 & 0.110  $\pm$  0.032 & 0.048  $\pm$  0.016 & 0.048  $\pm$  0.016 & 0.997  $\pm$  0.000 \\
AE-MMD-hybrid-sampler (Ours) & 133.148  $\pm$  52.617 & 0.203  $\pm$  0.021 & 0.151  $\pm$  0.025 & 0.235  $\pm$  0.070 & -16.101  $\pm$  0.283 & 0.025  $\pm$  0.002 & 0.278  $\pm$  0.060 & 0.149  $\pm$  0.051 & 0.146  $\pm$  0.050 & 0.998  $\pm$  0.000 \\
DGAE & 7.087  $\pm$  3.365 & 0.586  $\pm$  0.235 & 0.583  $\pm$  0.233 & 0.275  $\pm$  0.159 & -15.023  $\pm$  0.701 & 0.032  $\pm$  0.005 & 0.000  $\pm$  0.000 & 0.190  $\pm$  0.076 & 0.187  $\pm$  0.074 & 0.997  $\pm$  0.001 \\
NASHAE & 8.577 $\pm$ 4.533 & 0.368 $\pm$ 0.026 & 0.383 $\pm$ 0.023 & 0.040 $\pm$ 0.009 & -13.152 $\pm$ 0.681 & 0.049 $\pm$ 0.007 & 0.168 $\pm$ 0.007 & 0.145 $\pm$ 0.043 & 0.143 $\pm$ 0.044 & 0.997 $\pm$ 0.001 \\
FactorVAE ($\gamma$=10.0) & 1.609  $\pm$  0.127 & 0.167  $\pm$  0.020 & 0.165  $\pm$  0.032 & 0.176  $\pm$  0.114 & -15.517  $\pm$  0.366 & 0.028  $\pm$  0.002 & 0.192  $\pm$  0.090 & 0.149  $\pm$  0.098 & 0.148  $\pm$  0.097 & 0.997  $\pm$  0.001 \\
$\beta$-TCVAE ($\beta$=4.0) & 1.770  $\pm$  0.411 & 0.171  $\pm$  0.024 & 0.148  $\pm$  0.028 & 0.277  $\pm$  0.115 & -12.402  $\pm$  0.257 & 0.058  $\pm$  0.003 & 0.227  $\pm$  0.109 & 0.246  $\pm$  0.108 & 0.245  $\pm$  0.107 & 0.999  $\pm$  0.000 \\
$\beta$-VAE ($\beta$=4) & 1.224  $\pm$  0.050 & 0.276  $\pm$  0.063 & 0.258  $\pm$  0.073 & 0.396  $\pm$  0.065 & -10.983  $\pm$  0.672 & 0.081  $\pm$  0.012 & 0.238  $\pm$  0.053 & 0.264  $\pm$  0.031 & 0.262  $\pm$  0.031 & 0.988  $\pm$  0.006 \\
\midrule
Standard AE & 12.609  $\pm$  4.281 & 0.265  $\pm$  0.028 & 0.284  $\pm$  0.031 & 0.099  $\pm$  0.052 & -18.422  $\pm$  0.279 & 0.014  $\pm$  0.001 & 0.130  $\pm$  0.081 & 0.069  $\pm$  0.029 & 0.068  $\pm$  0.028 & 0.998  $\pm$  0.000 \\
Standard VAE & 1.630  $\pm$  0.175 & 0.179  $\pm$  0.042 & 0.165  $\pm$  0.051 & 0.163  $\pm$  0.077 & -15.531  $\pm$  0.432 & 0.028  $\pm$  0.003 & 0.184  $\pm$  0.079 & 0.106  $\pm$  0.062 & 0.105  $\pm$  0.061 & 0.996  $\pm$  0.001 \\
\bottomrule
\end{tabular}
}
\end{table*}

\clearpage
\section{Metric Analysis}
\label{app:Metric Analysis}

\noindent This section provides a deeper analysis of the properties of common disentanglement metrics, justifying the introduction and use of our Latent Predictability Score (LPS). While alignment-based metrics like DCI and MIG are popular, our experiments reveal that they can exhibit paradoxical behavior and high instability, potentially leading to misleading conclusions about the quality of a learned representation.

\subsection{The Alignment-Independence Paradox}
\label{app:paradox}

\begin{figure*}[tp]
    \centering
    \includegraphics[width=1\columnwidth]{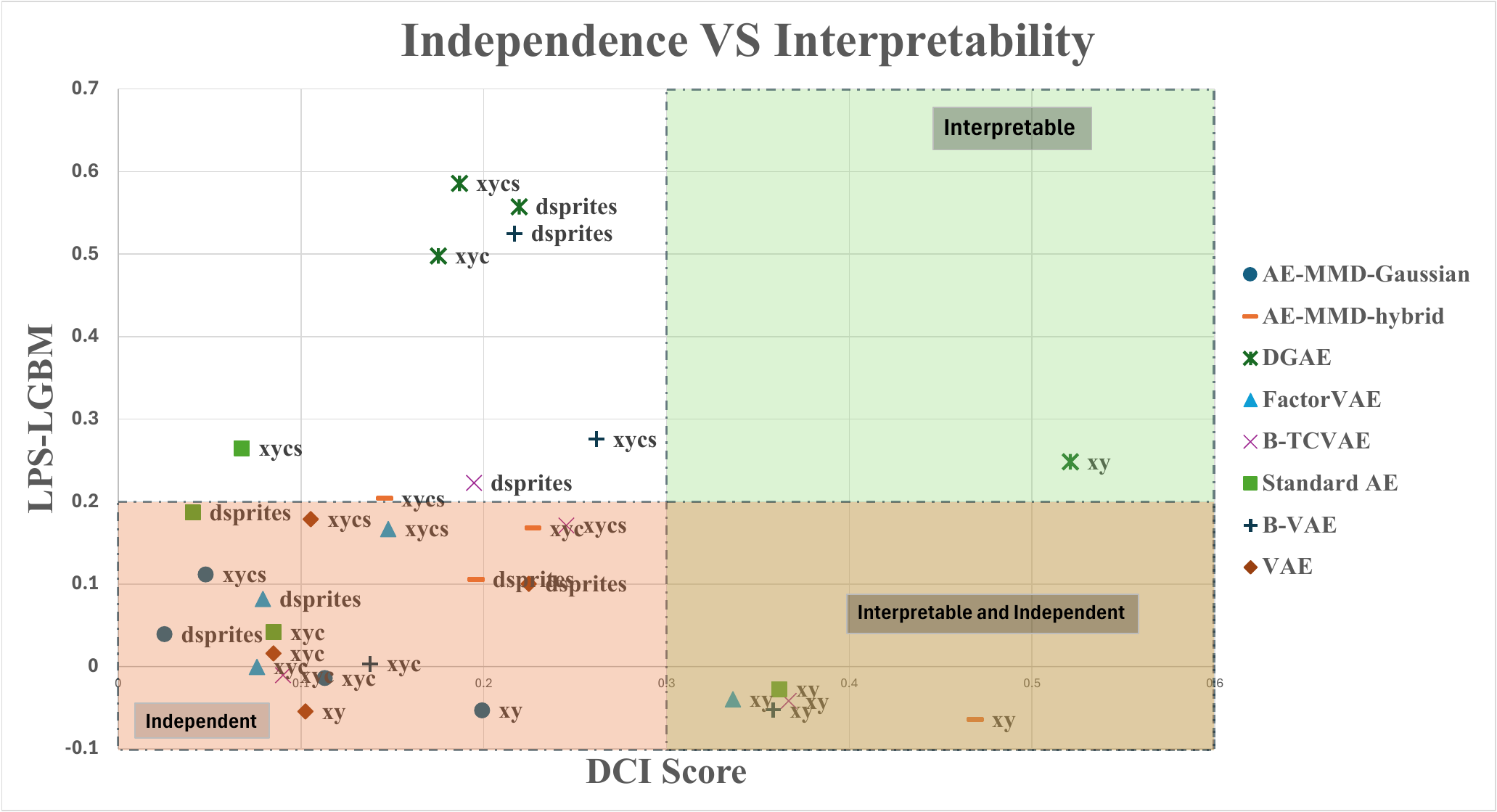}
    \caption{\textbf{The Paradox of Independence vs. Alignment.} Each point represents a model. The ideal model would be in the bottom-right (high alignment, high independence). We observe that models like DGAE occupy the unfavorable top-left quadrant, achieving a high DCI score only by sacrificing independence. Our method (with a Gaussian prior) occupies the bottom-left, prioritizing true independence over potentially flawed alignment metrics.}
    \label{fig:paradox_plot}
\end{figure*}
A core assumption is that a higher alignment score (like DCI) corresponds to a better, more disentangled representation. Our findings challenge this notion, revealing a troubling paradox: optimizing for alignment can actively work against the goal of learning statistically independent factors.

This effect is visualized in the scatter plot in Figure~\ref{fig:paradox_plot}. The ideal model would occupy the bottom-right quadrant, signifying both high alignment (high DCI) and high independence (low LPS). However, we observe that baseline models frequently land in the bottom-left or top-left "paradox zone." A clear example is DGAE on the XY dataset, which achieves the highest DCI score (0.520) but does so with a highly entangled representation (LPS of 0.25) and poor reconstruction. This suggests that DCI may be rewarding the model for learning spurious correlations, an effect conjectured in past work which noted that many benchmark datasets contain significant correlations between their ground-truth factors \citep{locatello2019challenging}. For instance, in dSprites and XYCS, an object's size and position are co-dependent (a large shape cannot be centered near an edge). A model that correctly learns this co-dependence may score well on DCI but has fundamentally failed to learn truly independent factors. This highlights a critical flaw in relying solely on alignment-based metrics as a proxy for disentanglement.

\begin{table*}[tp]
\centering
\caption{Comparison of Disentanglement (DCI) and Latent Predictability Score (LPS) across models and synthetic datasets. For Disentanglement, higher is better ($\uparrow$), and for LPS, lower is better ($\downarrow$).}
\label{tab:disentanglement_vs_lps}
\resizebox{\textwidth}{!}{
\begin{tabular}{@{}l*{8}{c}@{}}
\toprule
& \multicolumn{2}{c}{\textbf{dsprites}} & \multicolumn{2}{c}{\textbf{xy}} & \multicolumn{2}{c}{\textbf{xyc}} & \multicolumn{2}{c}{\textbf{xycs}} \\
\cmidrule(lr){2-3} \cmidrule(lr){4-5} \cmidrule(lr){6-7} \cmidrule(lr){8-9}
\textbf{Baseline} & Disent. ($\uparrow$) & LPS ($\downarrow$) & Disent. ($\uparrow$) & LPS ($\downarrow$) & Disent. ($\uparrow$) & LPS ($\downarrow$) & Disent. ($\uparrow$) & LPS ($\downarrow$) \\
\midrule
\textbf{AE-MMD-Gaussian} & 0.025 & \textbf{0.039} & 0.199 & -0.053 & 0.113 & \textbf{-0.014} & 0.048 & \textbf{0.111} \\
\textbf{AE-MMD-hybrid} & 0.196 & 0.105 & 0.469 & \textbf{-0.065} & \textbf{0.227} & 0.168 & 0.146 & 0.203 \\
DGAE & \textbf{0.219} & 0.557 & \textbf{0.521} & 0.248 & 0.175 & 0.498 & 0.187 & 0.596 \\
FactorVAE & 0.079 & 0.082 & 0.336 & -0.040 & 0.076 & -0.000 & 0.148 & 0.187 \\
$\beta$-TCVAE & 0.195 & 0.222 & 0.368 & -0.042 & 0.090 & -0.010 & 0.245 & 0.171 \\
$\beta$-VAE & 0.217 & 0.525 & 0.358 & -0.053 & 0.138 & 0.003 & \textbf{0.262} & 0.276 \\
\midrule
Standard AE & 0.041 & 0.187 & 0.362 & -0.028 & 0.085 & 0.042 & 0.068 & 0.265 \\
Standard VAE & 0.225 & 0.100 & 0.102 & -0.055 & 0.085 & 0.016 & 0.105 & 0.180 \\
\bottomrule
\end{tabular}
}
\end{table*}

\subsection{Metric Instability Across Runs}
\label{app:instability}
\begin{figure*}[tp]
    \centering
    \begin{subfigure}[b]{0.8\textwidth}
        \centering
        \includegraphics[width=\textwidth]{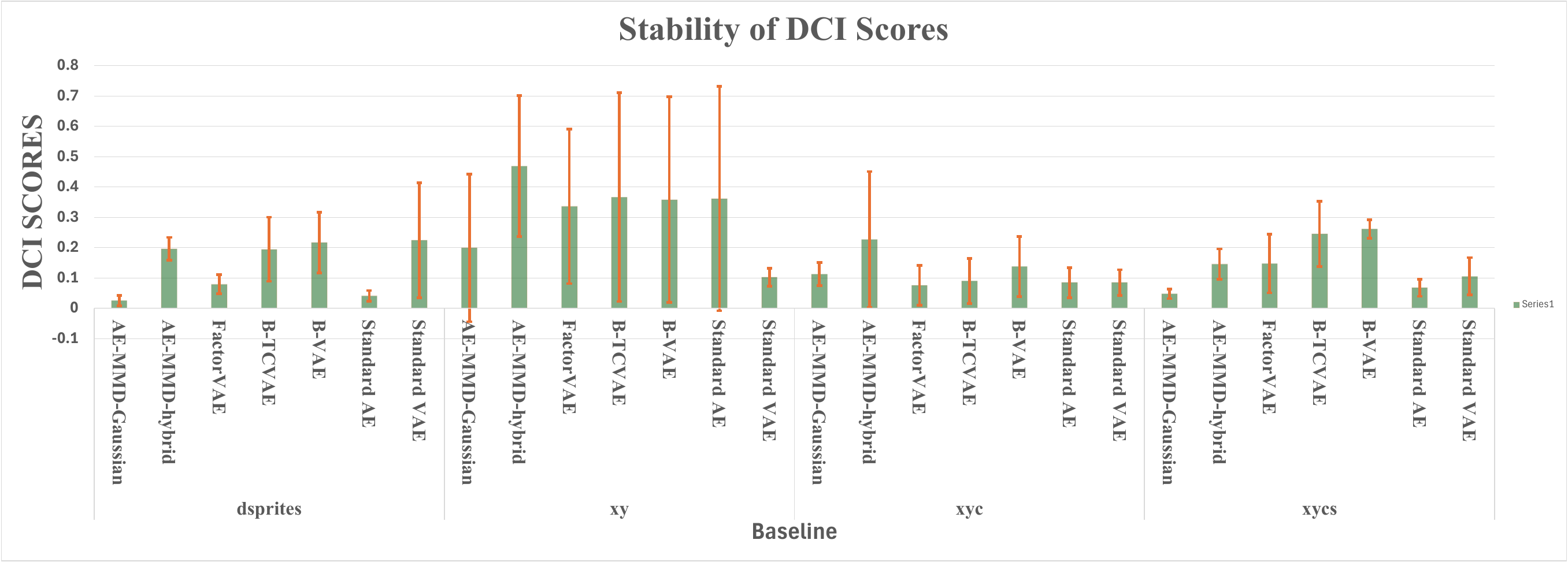}
        \caption{DCI Score Instability}
        \label{fig:dci_variance}
    \end{subfigure}
    \hfill 
    \begin{subfigure}[b]{0.8\textwidth}
        \centering
        \includegraphics[width=\textwidth]{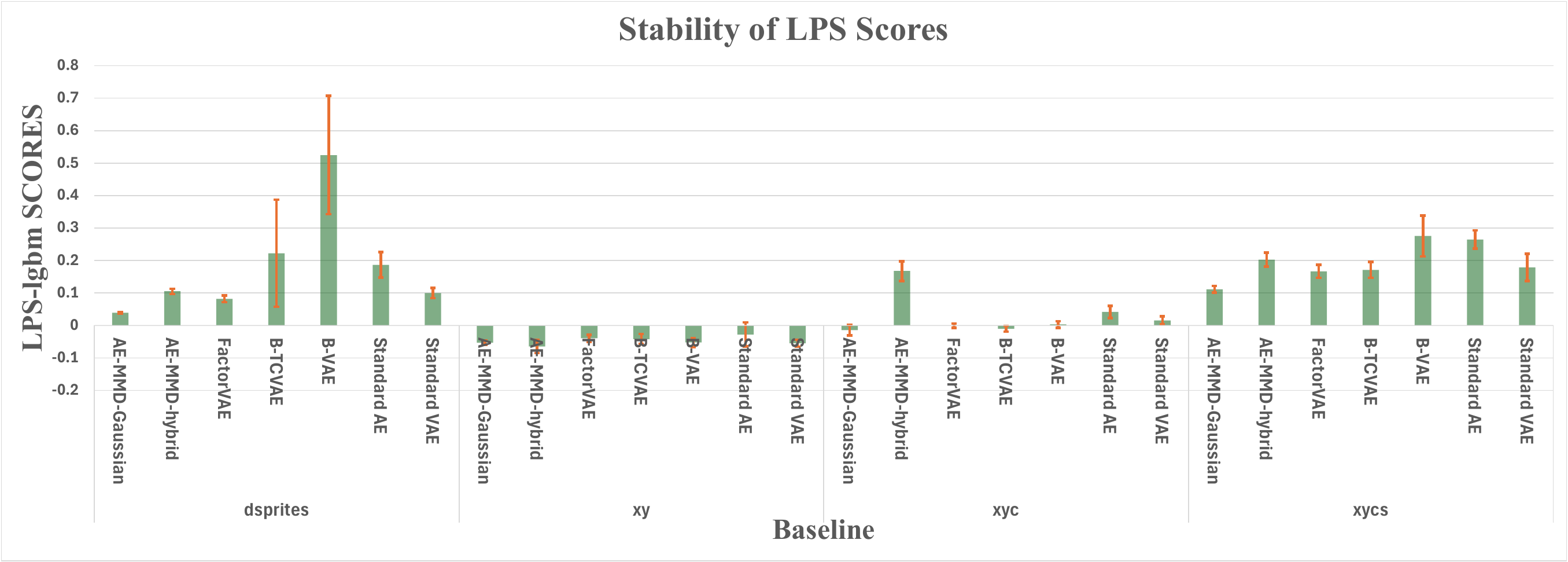}
        \caption{LPS Score Stability}
        \label{fig:lps_variance}
    \end{subfigure}
    \caption{
        \textbf{Comparison of metric stability across 5 random seeds.} 
        The error bars represent one standard deviation. 
        \textbf{(a)} The DCI scores exhibit extremely high variance, with the standard deviation often being as large as the mean itself, highlighting the metric's unreliability. 
        \textbf{(b)} In stark contrast, our proposed LPS metric shows minimal variance, demonstrating its stability and reliability as an intrinsic measure of representation quality.
    }
    \label{fig:variance_comparison}
\end{figure*}

Beyond the paradox, we find that standard alignment metrics are highly unstable, exhibiting extreme variance across different random seeds. This is quantified and visualized in Figure~\ref{fig:variance_comparison}. As shown in Figure~\ref{fig:variance_comparison}(a), the standard deviation of the DCI score for baseline methods is often as large as, or even larger than, the mean itself. This instability, a known symptom of the NICA identifiability problem \citep{locatello2019challenging}, makes scores from single runs untrustworthy and complicates reliable model comparison.

In stark contrast, Figure~\ref{fig:variance_comparison}(b) shows that our LPS metric is highly stable, with consistently small error bars across all models and datasets. This stability stems from the fact that LPS measures an intrinsic, unsupervised property of the representation itself (mutual independence) rather than its alignment to a specific, and potentially arbitrary, set of ground-truth labels.

\subsection{LPS as a Stable, Intrinsic Measure}
\label{app:lps_stability}
The paradoxical behavior and instability of alignment-based metrics underscore the need for a more fundamental and reliable measure of representation quality. We argue that the LPS fills this critical gap. By focusing on statistical independence, LPS provides a stable, intrinsic, and fully unsupervised signal. It directly answers the question, "Are the learned features statistically independent?" without being confounded by the potential issues of the ground-truth labels or the inherent instability of the alignment problem. While alignment is a desirable downstream goal, achieving a verifiably independent representation is a more foundational first step, and LPS provides the tool to measure it robustly.

\clearpage
\section{Scalibility, Stability, and Robustness of The MMD-Based Sculpting Regularizer}
\label{app:MMD_Robustness}

In the main text, we demonstrated that the Programmable Prior Framework achieves state-of-the-art disentanglement. In this section, we provide a comprehensive ablation study to verify the stability of the MMD Sculpting regularizer. Since MMD is a sample-based, non-parametric estimator, it is crucial to examine its behavior with respect to three key implementation factors:
\begin{enumerate}
    \item \textbf{Model Capacity:} Does the latent space become entangled as the dimension increases?
    \item \textbf{Batch Size:} Does the MMD estimator require prohibitively large batches to converge?
    \item \textbf{Kernel Choice:} Is the method sensitive to the specific choice of kernel function or bandwidth hyperparameters?
\end{enumerate}

The following experiments were conducted on the CIFAR-10 dataset using the LPS-lgbm metric (unless otherwise stated) to quantify mutual independence, alongside the reconstruction NMSE (dB) to verify that any gains in disentanglement do not come at the cost of model performance.

\subsection*{Robustness to Latent Space Dimensionality}
We first investigate the "capacity problem" common in VAEs, where increasing the latent dimension often leads to posterior collapse or increased entanglement. We trained models with latent dimensions ranging from $d=32$ to $d=96$.

\begin{figure*}[tp]
    \centering
    \begin{subfigure}[b]{0.28\textwidth}
        \centering
        \includegraphics[width=\linewidth]{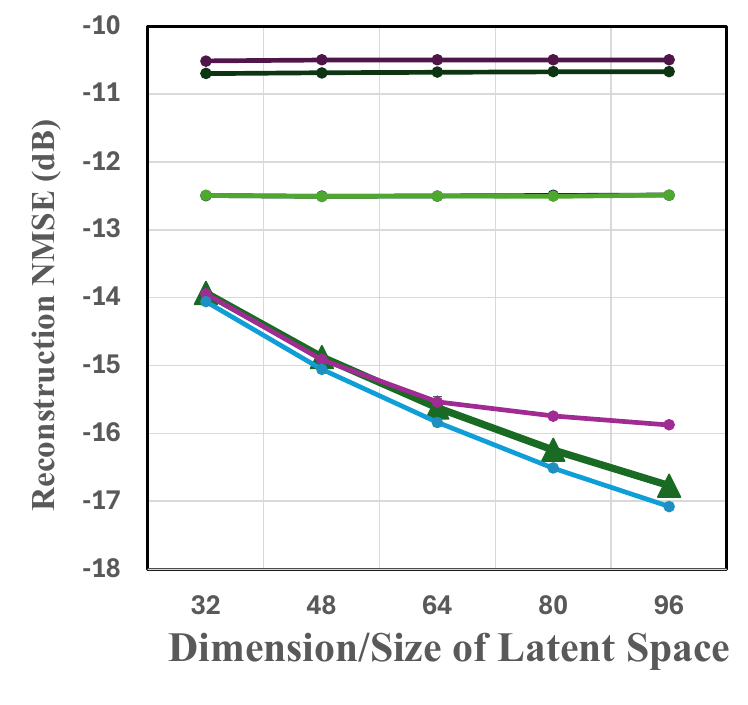}
    \end{subfigure}
    \begin{subfigure}[b]{0.28\textwidth}
        \centering
        \includegraphics[width=\linewidth]{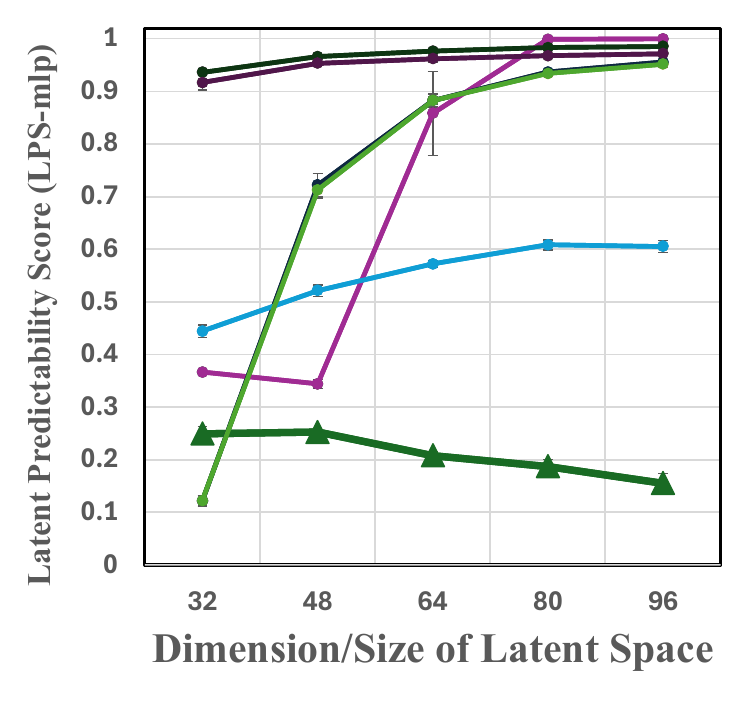}
    \end{subfigure}
    \hfill
    \begin{subfigure}[b]{0.41\textwidth}
        \centering
        \includegraphics[width=\linewidth]{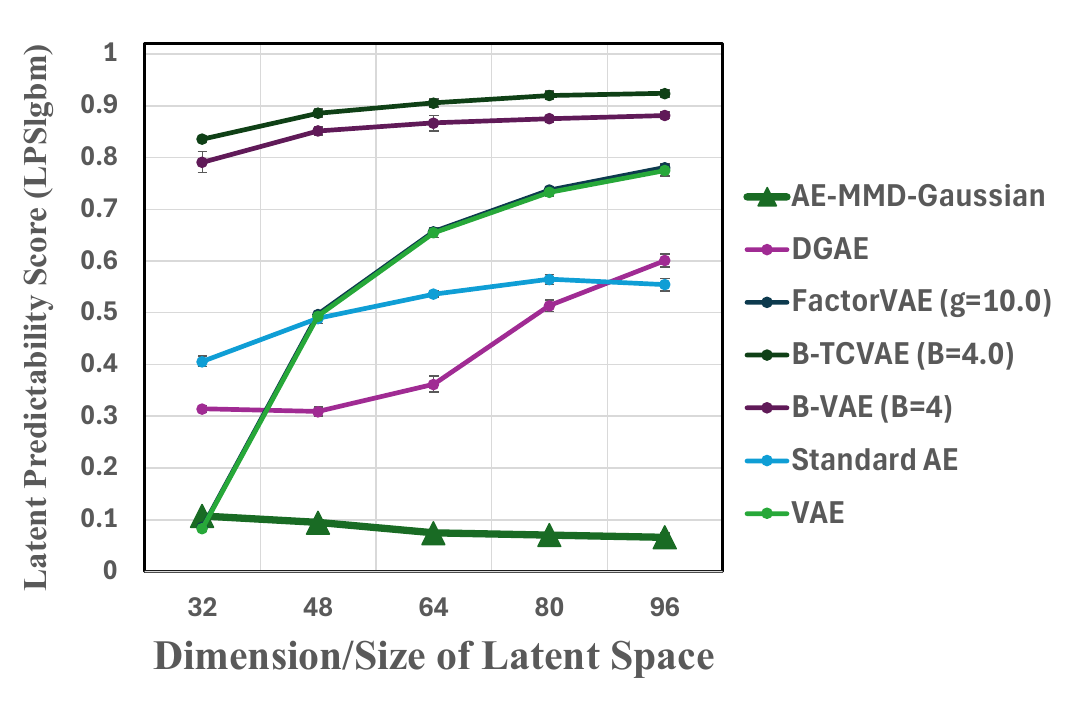}
    \end{subfigure}
    \caption{
        \textbf{Latent Independence vs. Latent Space Dimension on CIFAR-10.} The plots show the reconstruction NMSE in decibels (left), and the Latent Predictability Score (LPS) using MLP (middle) and LGBM (right) regression as the size of the latent bottleneck increases. Lower scores are better for all metrics. Our method (AE-MMD-Gauss, green line) maintains a high degree of independence across all dimensions, while most baselines learn increasingly entangled representations. Moreover, we achieve this stability at no significant cost to reconstruction error, which is surpassed only by the unregularized standard AE model.
    }
    \label{fig:LPS_vs_latent_dim}
\end{figure*}

As shown in Figure~\ref{fig:LPS_vs_latent_dim}, our MMD-based framework (green line with triangles) exhibits unique robustness:
\begin{itemize}
    \item \textbf{Independence (Middle/Right Plots):} The LPS scores for our model remain consistently low ($<0.25$) across all dimensions. In stark contrast, baseline methods like $\beta$-TCVAE and FactorVAE show clear degradation, with LPS scores rising towards $1.0$ (indicating complete predictability/entanglement) as the latent space grows.
    \item \textbf{Reconstruction (Left Plot):} Crucially, our method does not sacrifice reconstruction for this independence. The NMSE continues to improve (becoming more negative) as the dimension increases, tracking the performance of an unregularized Autoencoder (blue line). This contrasts with the standard VAE (light green circle), which plateaus at a higher error rate regardless of capacity.
\end{itemize}

\subsection*{Sensitivity to Batch Size and Regularization Strength}
The MMD is a sample-based statistic, meaning its variance depends on the batch size $N$. We evaluated the method's performance across batch sizes ranging from 64 to 2048 and regularization coefficients from 0.1 to 5.

\begin{figure*}[tp]
    \centering
    \begin{subfigure}[b]{0.44\textwidth}
        \centering
        \includegraphics[width=\linewidth]{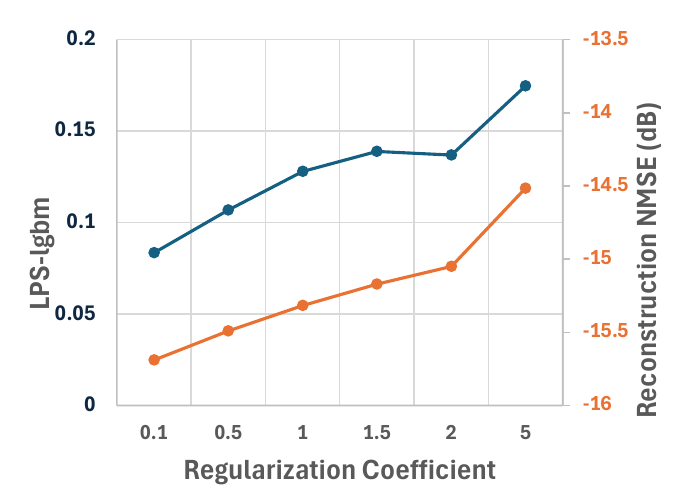}
    \end{subfigure}
    \begin{subfigure}[b]{0.44\textwidth}
        \centering
        \includegraphics[width=\linewidth]{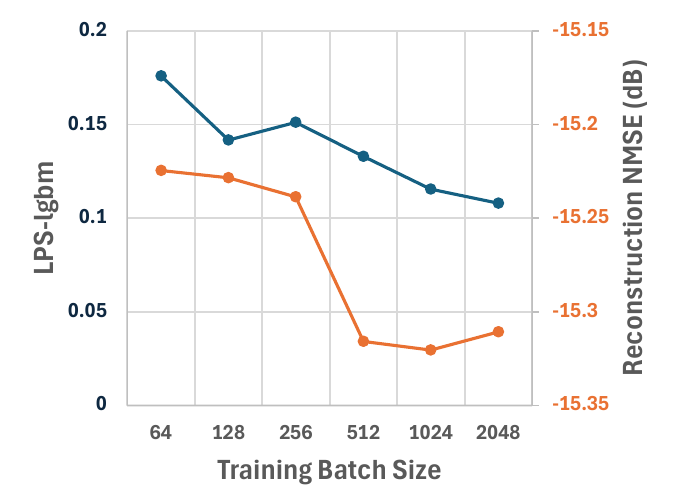}
    \end{subfigure}
    \caption{
        \textbf{Ablation Study on CIFAR-10: Regularization Coefficient and Batch Size.} We report the LPS-lgbm score and the test reconstruction NMSE (dB) against regularization coefficients ranging from $0.1$ to $5$ (left), and training batch sizes ranging from $64$ to $2048$ (right).
    }
    \label{fig:MMDAblation_regcoef_batchsize}
\end{figure*}

Figure~\ref{fig:MMDAblation_regcoef_batchsize} reveals two key insights:
\begin{itemize}
    \item \textbf{Independence (Blue Line):} There is a slight improvement in disentanglement (LPS drops from $\approx 0.17$ to $\approx 0.11$) as the batch size increases from 64 to 2048, or as the regularization coefficient decreases from 5 down to 0.1. This indicates some sensitivity, which is to be expected, but primarily highlights the robustness of the MMD regularizer to sample complexity per training iteration.
    \item \textbf{Reconstruction (Orange Line):} The reconstruction quality is remarkably stable, varying by less than $0.1$ dB across the entire range of batch sizes and by just over $1$ dB across the range of regularization coefficients. This suggests that even with smaller batches, the regularization does not overpower the reconstruction objective, indicating that the MMD objective to sculpt the latent space is not misaligned with the goal of reconstruction.
\end{itemize}

\subsection*{Robustness to Kernel Choice}
A common criticism of MMD is the need to tune kernel bandwidths. To address this, we utilize a mixture of kernels with the median heuristic (as detailed in Appendix A.3). Figure~\ref{fig:MMDAblation_KernelTypes} presents a scatter plot of performance for various kernel configurations.

\begin{figure*}[tp]
    \centering
    \includegraphics[width=0.6\linewidth]{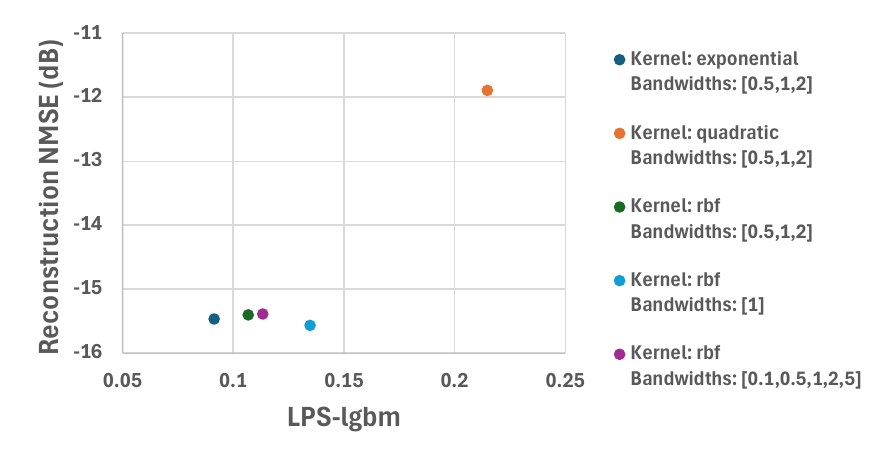}
    \caption{
        \textbf{Ablation Study of Kernel Type on CIFAR-10.} We report a scatter plot of the LPS-lgbm score versus the test reconstruction NMSE (dB) for exponential, quadratic, and Gaussian (RBF) kernel types. We also compare configurations with different numbers of bandwidths for kernel averaging according to the median heuristic.
    }
    \label{fig:MMDAblation_KernelTypes}
\end{figure*}

The results demonstrate high stability:
\begin{itemize}
    \item \textbf{Optimal Cluster:} The majority of configurations—including the Quadratic kernel and RBF kernels with widely varying bandwidth sets (e.g., single bandwidth vs. mixture of 5)—cluster tightly in the optimal bottom-left region (Low LPS $\approx 0.1$, Low NMSE $\approx -15.5$ dB).
    \item \textbf{Insensitivity to Bandwidth:} There is a small but statistically significant performance difference between using a single bandwidth (teal dot) and a mixture of 3 or even 5 bandwidths (purple dot). However, the difference between using 3 bandwidths versus 5 is negligible, validating the robustness of the median heuristic.
    \item \textbf{Outliers:} The Quadratic kernel (orange dot) performs noticeably worse in both reconstruction and independence at the same batch size and regularization coefficient, confirming that MMD is sensitive to the choice of kernel type (RBF/Exponential vs. Quadratic). However, we note that the quadratic kernel is rarely used for MMD applications compared to exponential or RBF kernels.
\end{itemize}

\clearpage
\section{The Power of Programmable Priors: Extended Results}
\label{app:programmable_priors}

This section provides a deeper dive into the concept of "programmable priors," showcasing (1) the flexibility of our MMD framework and (2) how just the choice of the target prior can significantly improve the unsupervised alignment of learned representations with desirable semantic factors. We present results where the latent space was sculpted to match a custom set of per-dimension priors, reflecting a better inductive bias for each dataset. The following visual and quantitative data will demonstrate that this prior engineering significantly improves ground-truth alignment metrics over the baseline \texttt{AE-MMD-Gaussian} model.

\begin{figure*}[tp]
    \centering
    \begin{subfigure}[b]{0.49\textwidth}
        \centering
        \includegraphics[width=\linewidth]{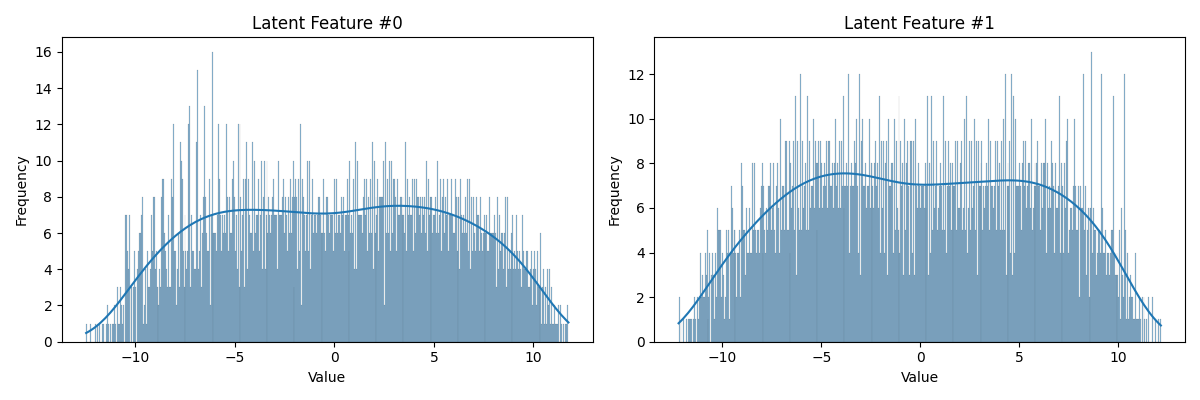}\\
        \vspace{1mm} 
        \includegraphics[width=\linewidth]{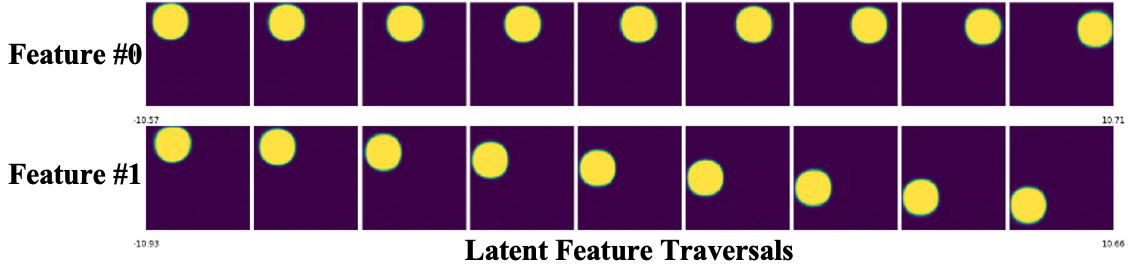}
        
        \caption{Our Model (AE with MMD-Uniform)}
        \label{fig:qual_col_mmd}
    \end{subfigure}
    \hfill 
    \begin{subfigure}[b]{0.49\textwidth}
        \centering
        \includegraphics[width=\linewidth]{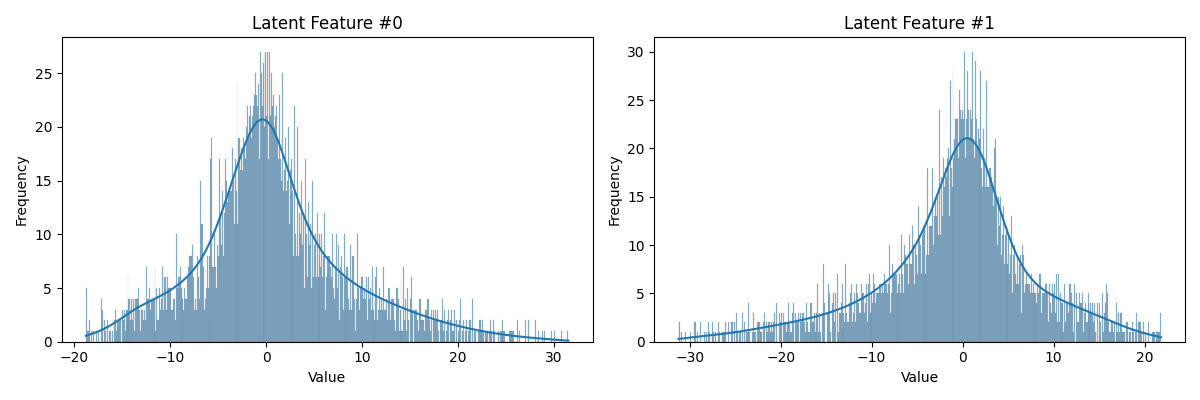}\\
        \vspace{1mm} 
        \includegraphics[width=\linewidth]{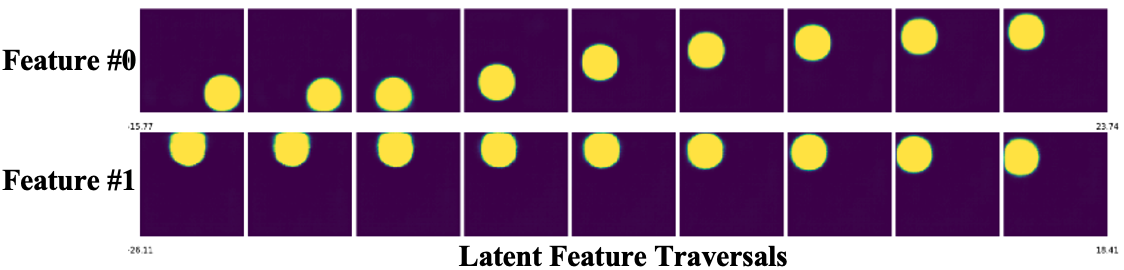}
        
        \caption{Standard AE}
        \label{fig:qual_col_ae}
    \end{subfigure}

    \caption{
        \textbf{Qualitative Comparison on the XY Dataset.} 
        Each column compares the learned latent marginals (top) with the corresponding latent traversals (bottom). 
        \textbf{(a)} Our MMD-trained model successfully learns the true uniform distribution, which results in clean, disentangled traversals. 
        \textbf{(b)} In contrast, the standard AE defaults to a uni-modal distribution, leading to entangled traversals.
    }
    \label{fig:qualitative_xy_columns}
\end{figure*}

Figure~\ref{fig:qualitative_xy_columns} provides a clear demonstration on the XY dataset (we provide additional results for the other synthetic datasets in the later subsections), where the ground-truth factor (position) follows a uniform distribution. When we program our prior to be uniform, our model successfully sculpts the latent marginals to match this target geometry. This is in stark contrast to a standard autoencoder, which defaults to a uni-modal representation that fails to capture the data structure with the vertical and horizontal x and y axis. The aligned latent space is evident in the latent traversals: our model's traversals are clean and disentangled, while the standard AE's are entangled and less interpretable.

\begin{table*}[tp]
\centering
\caption{
    Results for our MMD-based Autoencoder on the XY dataset with different engineered prior distributions. We find selecting a Uniform target prior significantly improve alignment metrics.
}
\label{tab:results_prior_engineering}
\begin{tabular}{lccccc}
\toprule
\textbf{Prior} & \textbf{NMSE (dB)} & \textbf{LPS-lgbm} & \textbf{LPS-mlp} & \textbf{DCI-RF} & \textbf{MIG} \\
\midrule
AE-Gaussian & -15.2          & -0.053          & \textbf{-0.004} & 0.199          & 0.751          \\
AE-Uniform  & \textbf{-15.8} & \textbf{-0.064} & -0.003          & \textbf{0.469} & \textbf{1.047} \\
AE-GMM (2 component)      & -14.9          & -0.051          & 0.000             & 0.251          & 0.863          \\
\bottomrule
\end{tabular}
\end{table*}

The quantitative results in Table~\ref{tab:results_prior_engineering} confirm that this improved structure directly translates to better alignment. On the XY dataset, using a standard factorized Gaussian prior achieves excellent statistical independence (LPS of -0.05) but yields a modest alignment score (DCI of 0.199). However, by simply switching to a Uniform prior that mirrors the true generative process, we dramatically improve the DCI and MIG scores, achieving near-perfect alignment without sacrificing independence. This result underscores a key contribution of our work: the prior is a critical tool for representation engineering, and our framework is unique in its ability to program this bias to achieve interpretable alignment.

\subsection{dSprites: Engineering a Mixed-Distribution Prior}
\label{app:dSprites_hybrid}

\begin{figure*}[tp]
    \centering
    \includegraphics[width=0.65\textwidth]{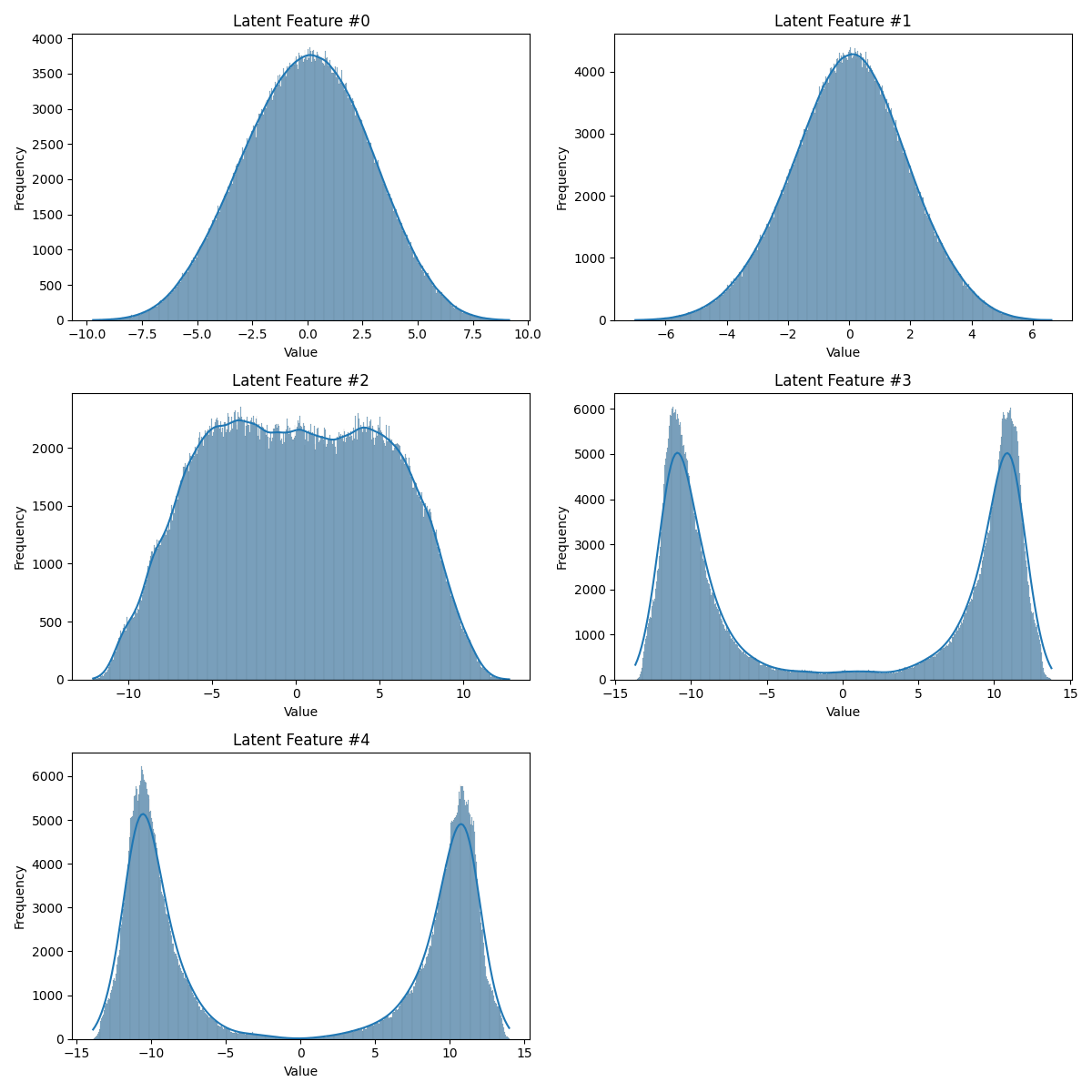}
    \caption{Learned marginal distributions for the \texttt{AE-MMD-hybrid-sampler} on dSprites. The histograms provide direct visual proof that the model successfully learned to match the specified target distributions for each latent dimension.}
    \label{fig:dsprite_hybrid_features}
\end{figure*}

For the 5-dimensional latent space of dSprites, we designed a hybrid prior to test the model's ability to learn a mix of simple and multi-modal distributions. The visual results in Figure~\ref{fig:dsprite_hybrid_features} confirm that the model's learned marginals consistently lead to improved recovery of desired latent features.

\noindent\textbf{Analysis.} The figure shows a direct correspondence between our specified priors and the learned representations:
\begin{itemize}
    \item \textbf{Dimension 1 (Uniform):} We specified a \texttt{Uniform(-5, 5)} prior. The model chose to ignore this instruction during training and instead adopted a Gaussian distribution with the same mean and variance.
    \item \textbf{Dimension 2 (Gaussian):} The target was a wide \texttt{Gaussian(0, 2)}, and the model learned the corresponding bell curve.
    \item \textbf{Dimensions 3 (Bi-modal GMMs):} For this dimensions, we used \texttt{auto-gmm} priors with two modes centered at -5 and 5, and with standard deviation equal to 5. The histograms clearly show that the model learned to partition the data into a more uniform distribution. We have found that specifying such a GMM structure would generally help the model adhere to a more uniform representation. while specifying the uniform distribution directly as in Dimension 1 would sometimes be 'ignored'.
    \item \textbf{Dimensions 4-5 (Bi-modal GMMs):} For the remaining dimensions, we used \texttt{auto-gmm} priors with two modes located at -10 and 10 respectively with standard deviation of 1 (using a small standard deviation of 1 is used to ensure the model adopts a bi-modal representation and not a uniform distribution as observed in Dimension 3) . The histograms clearly show that the model learned to partition the data into these distinct modes for each respective dimension.
\end{itemize}

\noindent\textbf{Quantitative Results.} As shown in Table~\ref{tab:dsprites}, this engineered prior leads to a dramatic improvement in alignment metrics. The \texttt{AE-MMD-hybrid-sampler} achieves a MIG score of 0.247, an over \textbf{4x improvement} compared to the Gaussian prior's score of 0.058. Similar substantial gains are seen in the SAP score (0.246 vs. 0.000) and the DCI disentanglement score (0.196 vs. 0.025), demonstrating that providing a better inductive bias directly translates to a more interpretable and aligned representation.
\begin{table*}[tp]
\caption{Summary for DSPRITES}
\label{tab:dsprites}
\centering
\resizebox{\textwidth}{!}{
\begin{tabular}{@{}l*{10}{l}@{}}
\toprule
 & \multicolumn{10}{c}{L=5} \\
\cmidrule(lr){2-11}
Baseline & Covariance Ratio & LPS\_LGBM & LPS\_MLP & MIG\_mig\_score & NMSE\_dB & NMSE\_linear & SAP\_sap\_score & completeness & disentanglement & informativeness \\
\midrule
AE-MMD-Gaussian & 102.265  $\pm$  21.202 & 0.039  $\pm$  0.002 & 0.033  $\pm$  0.006 & 0.058  $\pm$  0.024 & -17.888  $\pm$  0.281 & 0.016  $\pm$  0.001 & 0.000  $\pm$  0.000 & 0.026  $\pm$  0.018 & 0.025  $\pm$  0.017 & 0.950  $\pm$  0.002 \\
AE-MMD-hybrid-sampler & 263.218  $\pm$  87.577 & 0.105  $\pm$  0.008 & 0.087  $\pm$  0.012 & 0.247  $\pm$  0.030 & -16.311  $\pm$  0.261 & 0.023  $\pm$  0.001 & 0.246  $\pm$  0.027 & 0.200  $\pm$  0.038 & 0.196  $\pm$  0.038 & 0.957  $\pm$  0.001 \\
\bottomrule
\end{tabular}
}
\end{table*}

\subsection{XY: Engineering a Uniform Prior}
\label{app:XY_hybrid}

We provide the full table of results for the XY experiment.

\begin{table*}[tp]
\caption{Summary for XY}
\label{tab:xy}
\centering
\resizebox{\textwidth}{!}{
\begin{tabular}{@{}l*{10}{l}@{}}
\toprule
 & \multicolumn{10}{c}{L=2} \\
\cmidrule(lr){2-11}
Baseline & Covariance Ratio & LPS\_LGBM & LPS\_MLP & MIG\_mig\_score & NMSE\_dB & NMSE\_linear & SAP\_sap\_score & completeness & disentanglement & informativeness \\
\midrule
AE-MMD-Gaussian & 240.970  $\pm$  243.891 & -0.053  $\pm$  0.007 & -0.004  $\pm$  0.003 & 0.751  $\pm$  0.554 & -15.222  $\pm$  0.403 & 0.030  $\pm$  0.003 & 0.423  $\pm$  0.316 & 0.199  $\pm$  0.244 & 0.199  $\pm$  0.243 & 1.000  $\pm$  0.000 \\
AE-MMD-hybrid-sampler & 380.827  $\pm$  408.352 & -0.065  $\pm$  0.020 & -0.003  $\pm$  0.005 & 1.047  $\pm$  0.414 & -15.825  $\pm$  0.154 & 0.026  $\pm$  0.001 & 0.713  $\pm$  0.139 & 0.471  $\pm$  0.230 & 0.469  $\pm$  0.233 & 1.000  $\pm$  0.000 \\
\bottomrule
\end{tabular}
}
\end{table*}

\subsection{XYC: Engineering a GMM-Heavy Prior}
\label{app:XYC_hybrid}

\begin{figure*}[tp]
    \centering
    \includegraphics[width=0.65\textwidth]{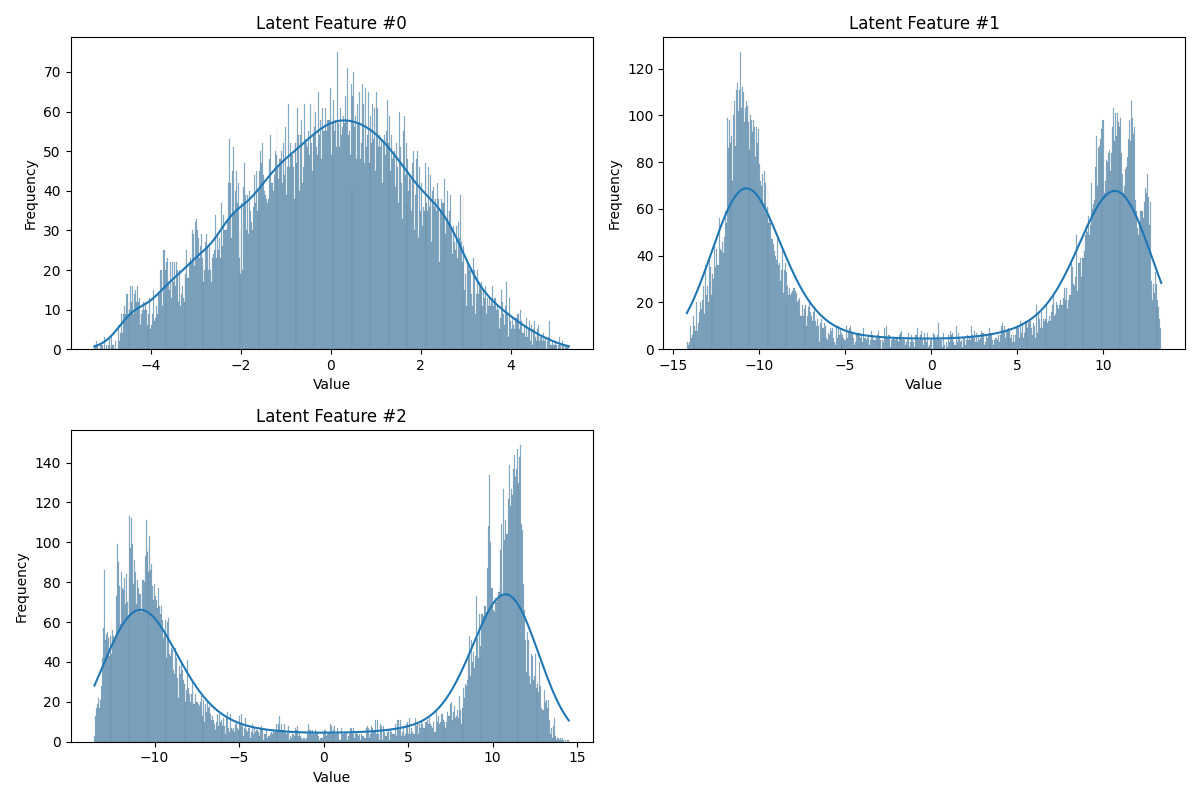}
    \caption{Learned marginal distributions for the \texttt{AE-MMD-hybrid-sampler} on XYC.}
    \label{fig:xyc_hybrid_features}
\end{figure*}

On the XYC dataset, we used a prior dominated by bi-modal distributions to encourage the model to discover clustered structures. Figure~\ref{fig:xyc_hybrid_features} visualizes the successful enforcement of this prior.

\noindent\textbf{Analysis.} The learned marginals again match the specified inductive bias:
\begin{itemize}
    \item \textbf{Dimension 1 (Gaussian):} The model correctly learns a \texttt{Gaussian(0, 2)} distribution.
    \item \textbf{Dimensions 2 \& 3 (Bi-modal GMMs):} We specified two identical \texttt{auto-gmm} priors with modes at -10 and 10. The resulting histograms clearly show two distinct clusters centered at these values, demonstrating the model's ability to learn multi-modal representations.
\end{itemize}

\noindent\textbf{Quantitative Results.} The quantitative results in Table~\ref{tab:xyc} confirm the benefit of this custom prior. The hybrid-sampler model nearly doubles the MIG score (0.465 vs. 0.256) and the DCI disentanglement score (0.227 vs. 0.113) compared to the model trained with a simple Gaussian prior.
\begin{table*}[tp]
\caption{Summary for XYC}
\label{tab:xyc}
\centering
\resizebox{\textwidth}{!}{
\begin{tabular}{@{}l*{10}{l}@{}}
\toprule
 & \multicolumn{10}{c}{L=3} \\
\cmidrule(lr){2-11}
Baseline & Covariance Ratio & LPS\_LGBM & LPS\_MLP & MIG\_mig\_score & NMSE\_dB & NMSE\_linear & SAP\_sap\_score & completeness & disentanglement & informativeness \\
\midrule
AE-MMD-Gaussian & 49.341  $\pm$  12.168 & -0.014  $\pm$  0.016 & 0.016  $\pm$  0.007 & 0.256  $\pm$  0.118 & -16.311  $\pm$  0.245 & 0.023  $\pm$  0.001 & 0.237  $\pm$  0.077 & 0.115  $\pm$  0.039 & 0.113  $\pm$  0.038 & 0.999  $\pm$  0.000 \\
AE-MMD-hybrid-sampler & 324.644  $\pm$  274.142 & 0.168  $\pm$  0.030 & 0.117  $\pm$  0.035 & 0.465  $\pm$  0.336 & -15.962  $\pm$  0.205 & 0.025  $\pm$  0.001 & 0.266  $\pm$  0.240 & 0.228  $\pm$  0.224 & 0.227  $\pm$  0.224 & 1.000  $\pm$  0.000 \\
\bottomrule
\end{tabular}
}
\end{table*}

\subsection{XYCS: Engineering a Mixed-Component Prior}
\label{app:XYCS_hybrid}

\begin{figure*}[tp]
    \centering
    \includegraphics[width=0.65\textwidth]{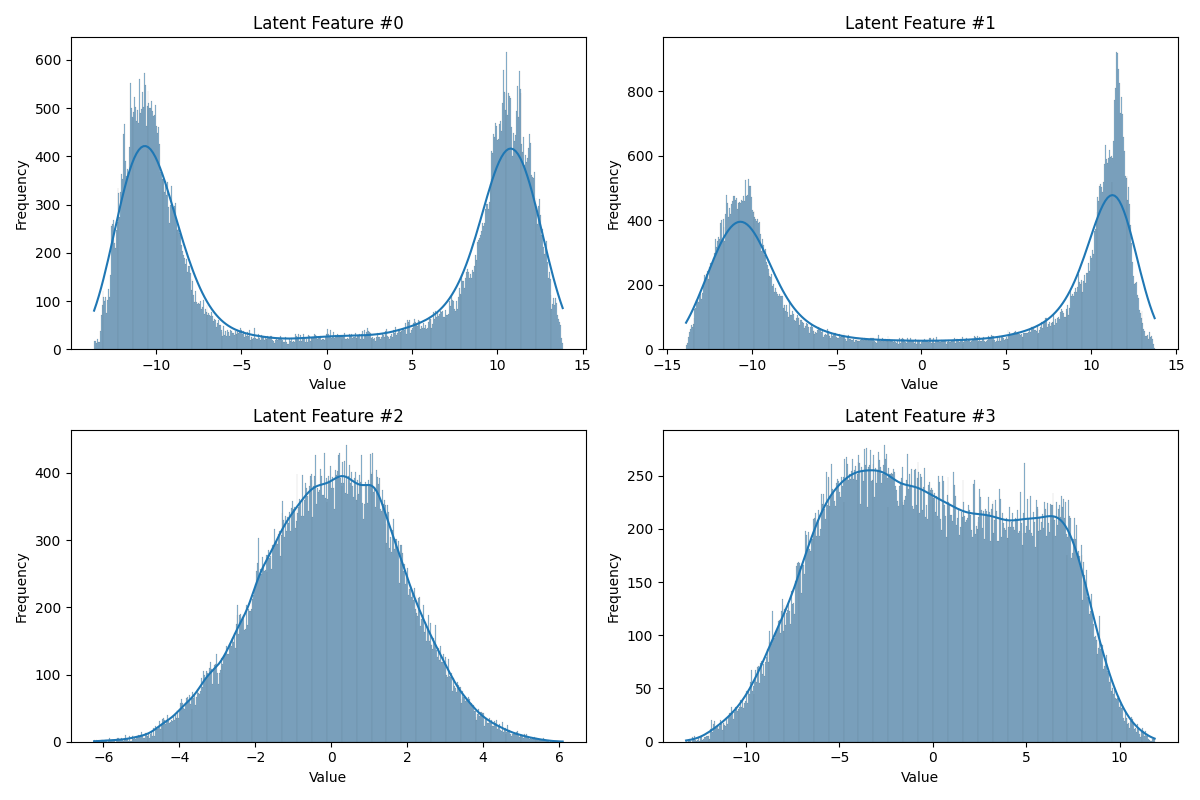}
    \caption{Learned marginal distributions for the \texttt{AE-MMD-hybrid-sampler} on XYCS.}
    \label{fig:xycs_hybrid_features}
\end{figure*}

For the 4-dimensional XYCS dataset, we again specified a mixed set of priors. The results in Figure~\ref{fig:xycs_hybrid_features} show that the model faithfully learns this complex target geometry.

\noindent\textbf{Analysis.} The learned distributions follow the specified configuration: two bi-modal GMMs with modes at $\pm10$, a standard Gaussian, and a narrower bi-modal GMM with modes at $\pm5$. Each histogram in the figure visually confirms that the intended structure was successfully imposed on the corresponding latent dimension.

\noindent\textbf{Quantitative Results.} As shown in Table~\ref{tab:xycs}, injecting this more informed inductive bias again leads to superior alignment. The hybrid-sampler model achieves a MIG score of 0.235, more than doubling the 0.092 score from the Gaussian model. Likewise, the SAP and DCI scores show significant improvements of over 2.5x and 3x, respectively.

\begin{table*}[tp]
\caption{Summary for XYCS}
\label{tab:xycs}
\centering
\resizebox{\textwidth}{!}{
\begin{tabular}{@{}l*{10}{l}@{}}
\toprule
 & \multicolumn{10}{c}{L=4} \\
\cmidrule(lr){2-11}
Baseline & Covariance Ratio & LPS\_LGBM & LPS\_MLP & MIG\_mig\_score & NMSE\_dB & NMSE\_linear & SAP\_sap\_score & completeness & disentanglement & informativeness \\
\midrule
AE-MMD-Gaussian & 42.117  $\pm$  2.445 & 0.111  $\pm$  0.011 & 0.106  $\pm$  0.020 & 0.092  $\pm$  0.036 & -16.757  $\pm$  0.143 & 0.021  $\pm$  0.001 & 0.110  $\pm$  0.032 & 0.048  $\pm$  0.016 & 0.048  $\pm$  0.016 & 0.997  $\pm$  0.000 \\
AE-MMD-hybrid-sampler & 133.148  $\pm$  52.617 & 0.203  $\pm$  0.021 & 0.151  $\pm$  0.025 & 0.235  $\pm$  0.070 & -16.101  $\pm$  0.283 & 0.025  $\pm$  0.002 & 0.278  $\pm$  0.060 & 0.149  $\pm$  0.051 & 0.146  $\pm$  0.050 & 0.998  $\pm$  0.000 \\
\bottomrule
\end{tabular}
}
\end{table*}

\subsection*{Summary}

The results in this section highlight a key takeaway: while a factorized Gaussian prior is highly effective at achieving statistical independence, it is not always the optimal choice for interpretability and alignment with ground-truth factors. The true power of the MMD framework is its flexibility, which allows practitioners to inject domain knowledge by engineering specific latent geometries. This capability leads to representations that are not only independent (or have a desired dependence structure) but are also more meaningfully aligned with the true factors of variation in the data.

However, our experiments with programmable priors also revealed a nuance in the MMD framework's behavior. While the regularizer consistently learned unimodal, Gaussian-like distributions with high fidelity, enforcing more specific geometries, such as uniform or multi-modal distributions, could sometimes present challenges. In these cases, the model occasionally defaulted to a unimodal distribution that correctly matched the target's mean and variance but not its overall shape.

We hypothesize this is primarily due to the inductive bias of the RBF kernel. As the RBF kernel is itself a Gaussian function, it is naturally more adept at matching distributions defined by their low-order moments \citep{tolstikhin2017wasserstein}. A secondary factor may be architectural constraints; the purely convolutional nature of the encoder might limit its capacity to shape the latent space into more complex geometries. We believe that future work could explore alternative kernels or incorporate fully connected layers to enhance the encoder's ability to model a wider variety of prior distributions.

\newpage
\section{Additional Experiments}
\label{app:Additional_Experiments}

\subsection{The Programmable Prior for Representational Knowledge Transfer}
\label{app:KnowledgeTransfer}

In the main text, we discussed our framework's ability to "copy" the latent distribution of a target model. Here, we provide this experiment's results on matching the aggregate prior. 

\begin{figure*}[tp]
    \centering
    \begin{subfigure}[b]{0.45\textwidth}
        \centering
        \includegraphics[width=0.55\linewidth]{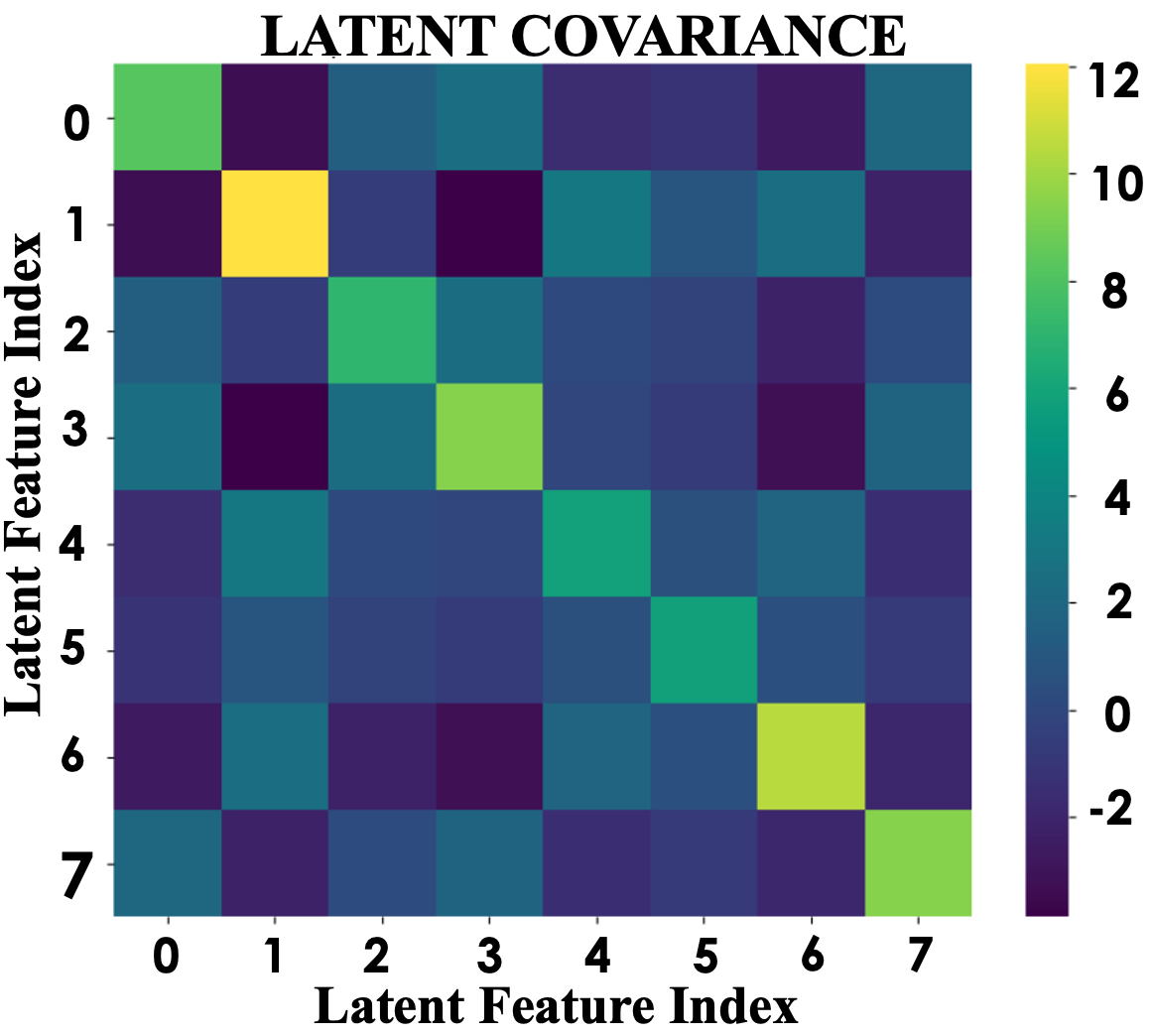}
        \includegraphics[width=0.43\linewidth]{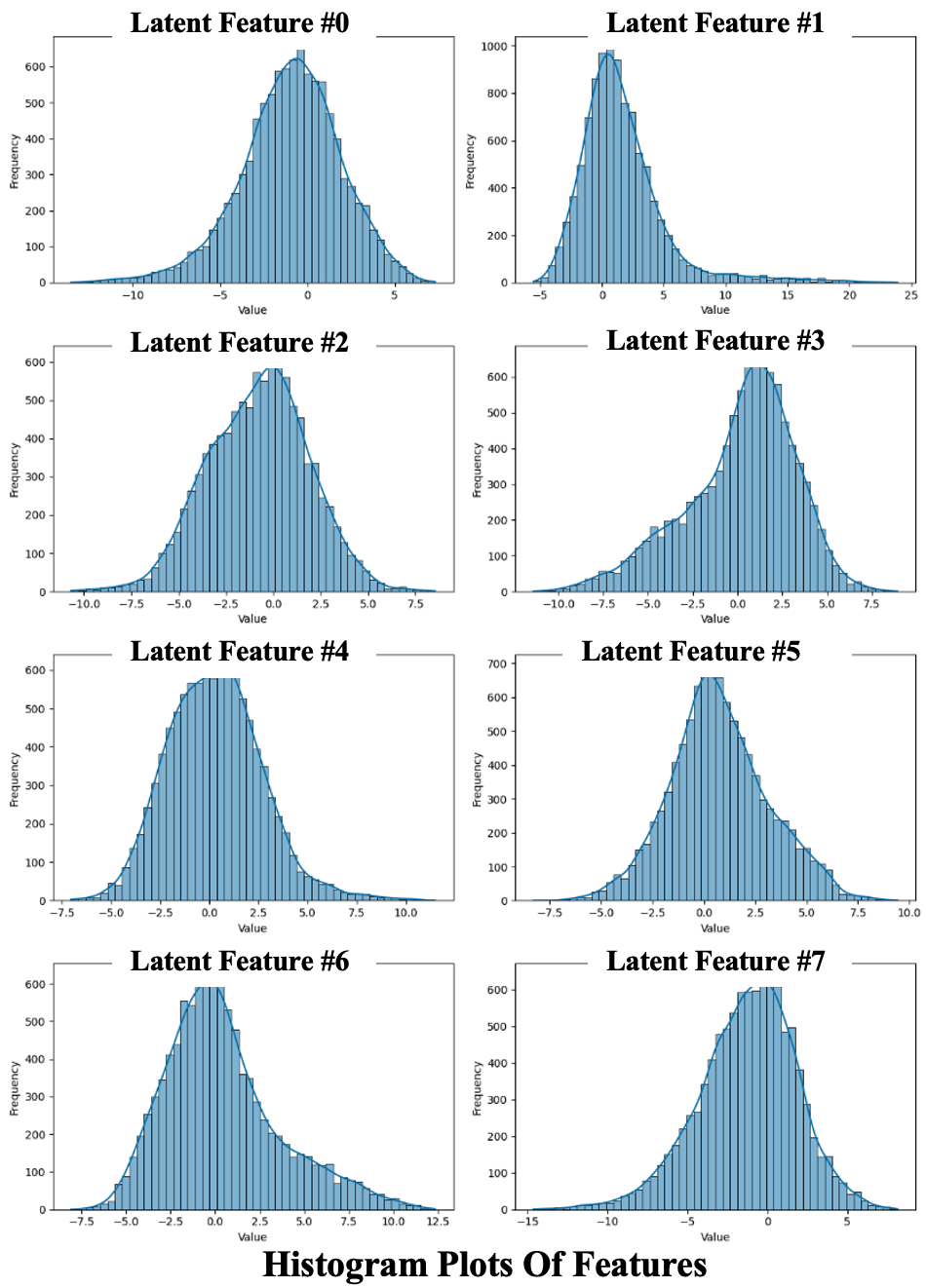}
        
        \caption{Our model (use MMD to copy (b))}
        \label{fig:mmd_copy_results}
    \end{subfigure}
    \begin{subfigure}[b]{0.45\textwidth}
        \centering
        \includegraphics[width=0.55\linewidth]{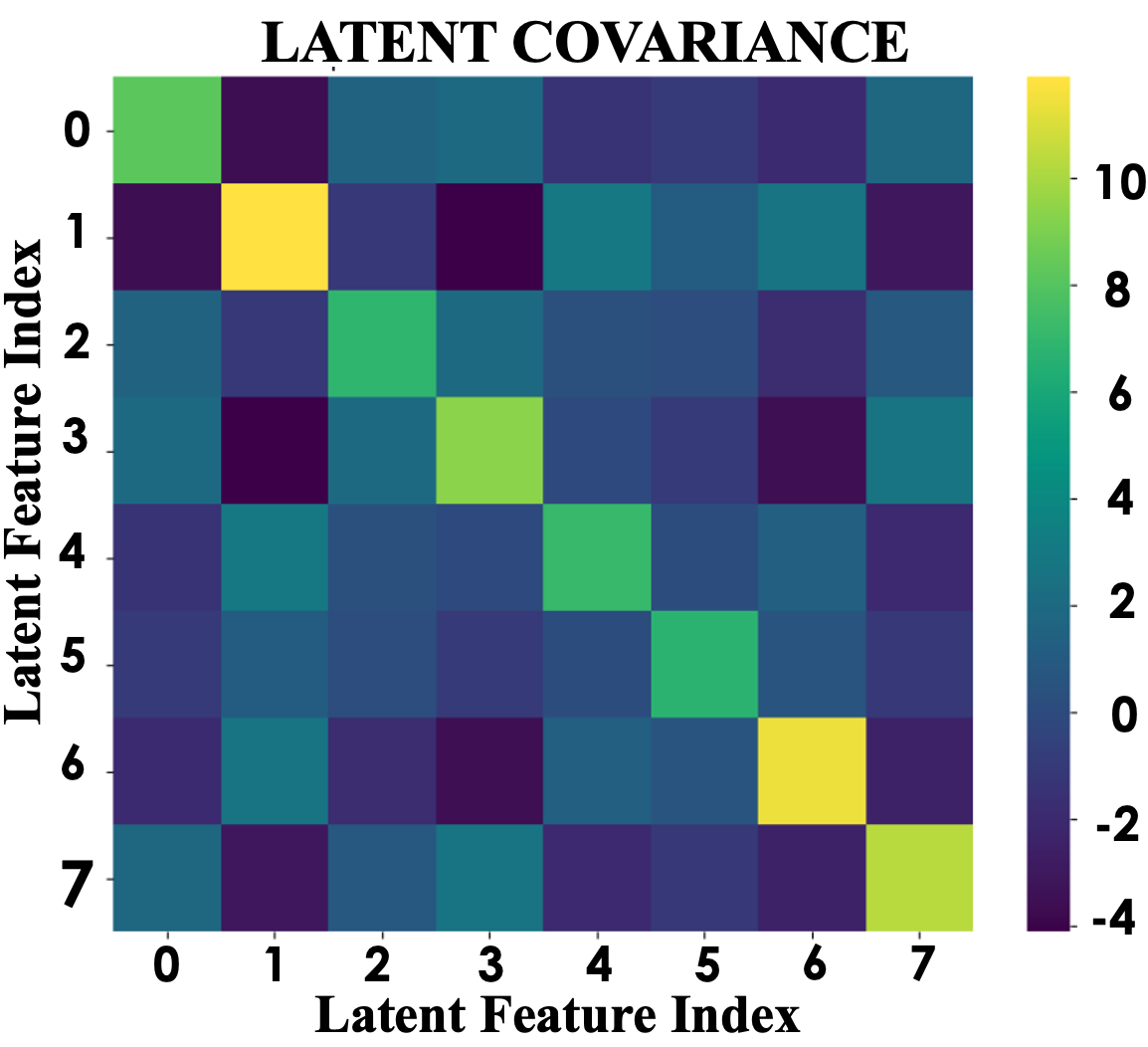}
        \includegraphics[width=0.43\linewidth]{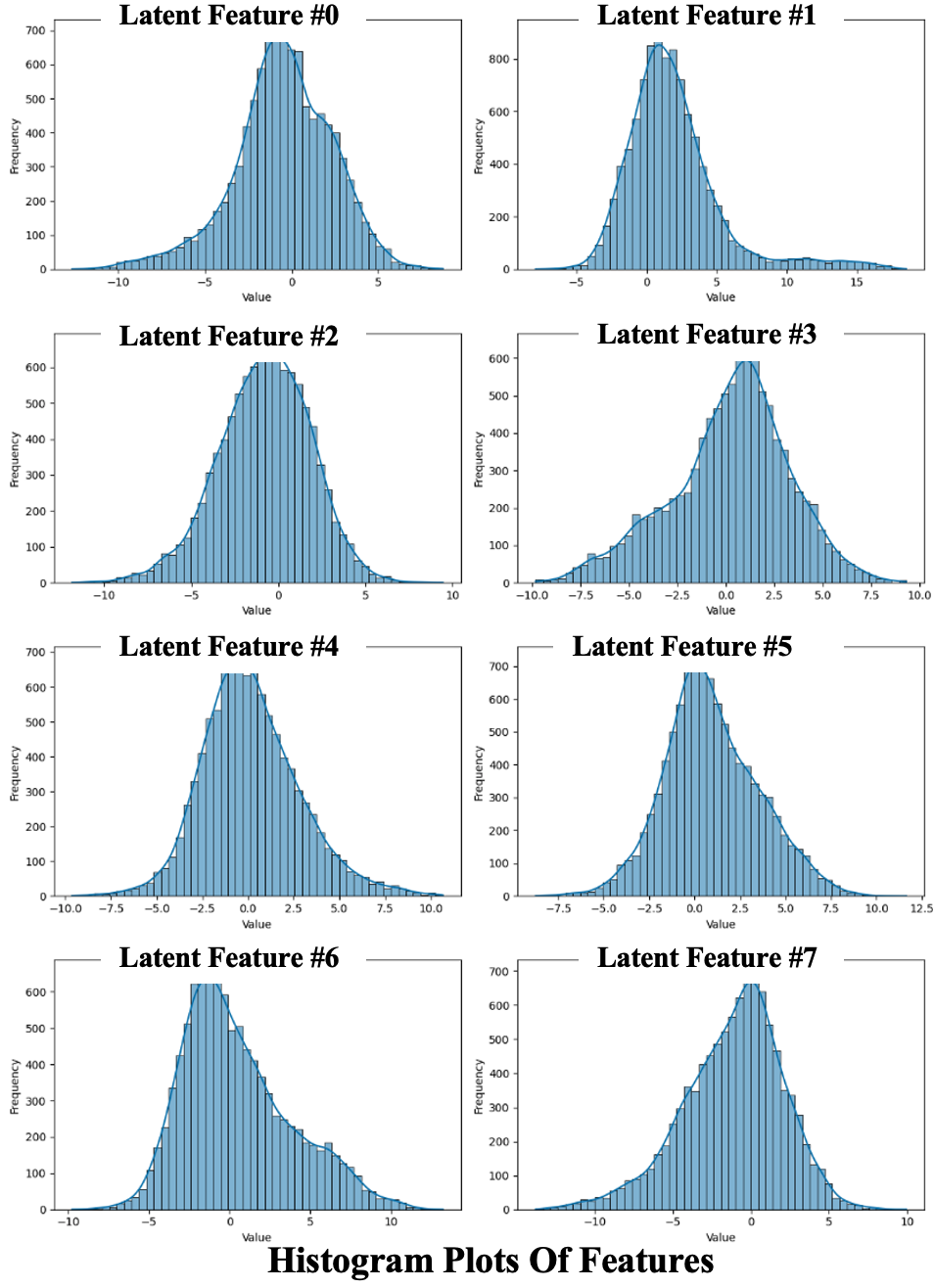}
        
        \caption{Prior Sampled from Target AE}
        \label{fig:ae_target_visuals}
    \end{subfigure}

    \caption{\textbf{Visualizing the Latent Space Copying experiment on MNIST.} (b) First, a standard Autoencoder was trained, and its complex, entangled latent distribution was saved to serve as an empirical prior. (a) Then, a new "student" model was trained using our MMD-based sculpting regularizer, tasked with replicating this non-analytic prior. The figure displays the covariance matrix between the latent space features over the whole dataset and the histogram plots of the marginal distribution of the latent space features over the whole dataset. The near-identical covariance matrices and marginal distributions visually confirm our method's ability to enforce a complex, co-dependent latent geometry that would be impossible to specify analytically.}
    \label{fig:latent_space_copy_combined}
\end{figure*}

We see that the model was able to match the target prior's covariance structure exactly while also closely matching the marginal histogram plots indicating that the Programmable Prior Framework is a powerful for being able to capture complex internal dependence structures and non analytic target distributions.

Additionally, we investigate this experiment from the perspective of student-teacher knowledge transfer to probe a deeper question: is matching the aggregate posterior sufficient to transfer a learned representation? We note that \citet{huang2017like} has also previously utilized MMD regularization for representational knowledge distillation. 

\begin{figure*}[tp]
    \centering
    \includegraphics[width=0.9\linewidth]{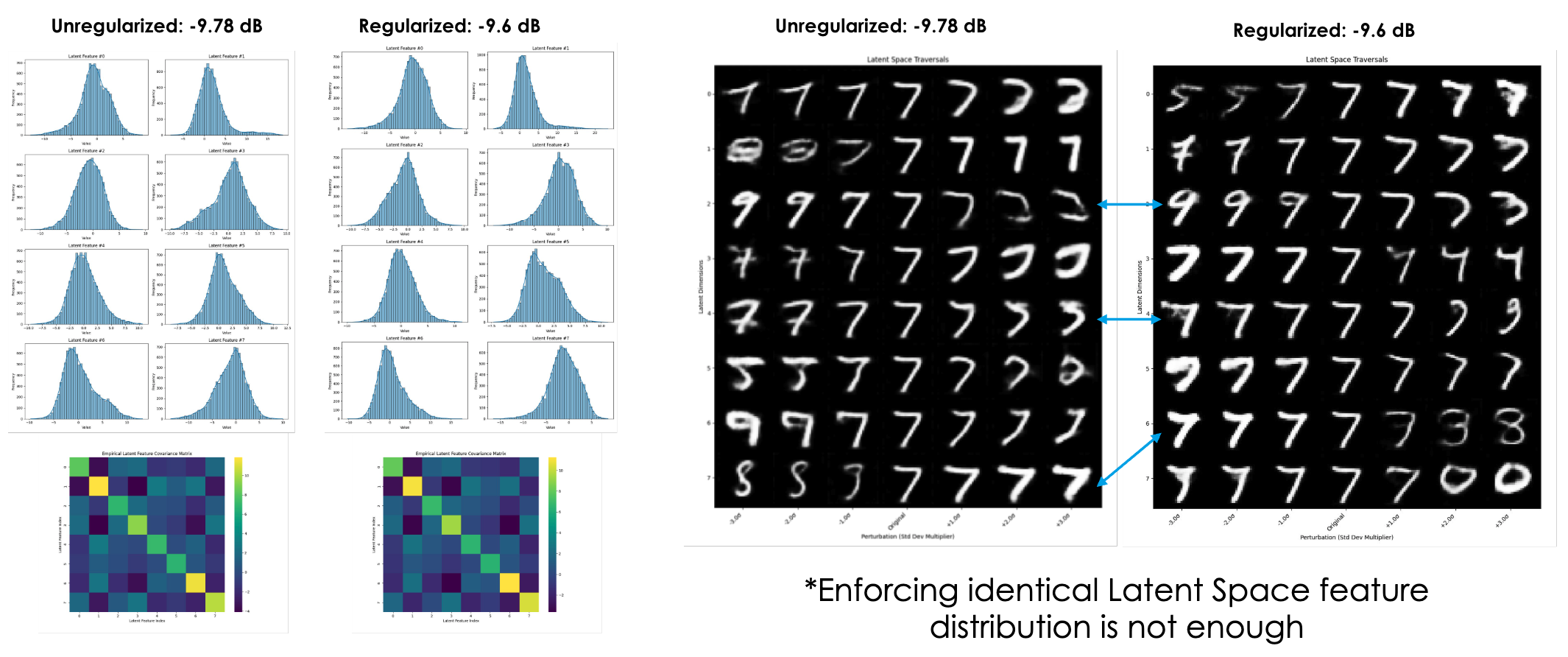}
    \caption{
        \textbf{Failure of Knowledge Transfer via Aggregate Posterior Matching.} (Left) Our MMD regularizer successfully forces a "student" model to replicate the latent distribution of a "teacher" model. (Right) Despite this, latent traversals show that the semantic meaning of the latent dimensions is completely different. For the same inputs (connected by lines), traversing a latent feature produces different visual changes, demonstrating that the underlying representations are not the same.
    }
    \label{fig:KnowledgeTransfer}
\end{figure*}

In this experiment, a "student" model was trained with our MMD regularizer to replicate the entire aggregate posterior distribution of a pre-trained, unregularized "teacher" autoencoder. As shown in the left panel of Figure~\ref{fig:KnowledgeTransfer}, the student was highly successful—the learned marginals and covariance matrix are nearly identical to those of the teacher. However, the right panel reveals that this statistical alignment does not translate to functional equivalence. Latent traversals, which reveal the semantic meaning encoded in each dimension, produce wildly different outputs for the same input image. While a traversal in the teacher model might alter an attribute like the slant of a digit, the corresponding traversal in the student model results in a completely different transformation.

This finding is crucial: \textbf{matching the aggregate posterior distribution is not enough to transfer a model's learned representation}. Even with identical latent distributions, the two models have learned fundamentally different mappings from the input data to the latent space. This suggests that for true representational knowledge transfer, one must enforce a more complex constraint, such as matching the \textit{joint} distribution of the latent representation with observable variables from the input data.